\documentclass[10pt,twocolumn,letterpaper]{article}

\usepackage{cvpr}              

\definecolor{gray}{HTML}{adbad0}
\definecolor{green}{HTML}{00b050}
\definecolor{red}{HTML}{e04f4f}
\definecolor{proposed}{HTML}{afc8d6}
\definecolor{accepted}{HTML}{c8e3d6}
\definecolor{rejected}{HTML}{d6b8c7}
\definecolor{empty}{HTML}{f4f8fb}

\usepackage{amsmath}
\usepackage[ruled,vlined,linesnumbered]{algorithm2e}

\usepackage{graphicx}

\newcommand{\emojiturtle}{\includegraphics[height=1.2em]{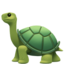}}
\newcommand{\emojirabbit}{\includegraphics[height=1.2em]{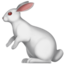}}
\newcommand{\emojirepeat}{\includegraphics[height=1.2em]{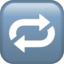}}

\newcommand{\methodname}{\textsc{MuLo-SD}}

\newcommand{\ttimes}{$\times$}

\usepackage{multirow}
\usepackage[table, dvipsnames]{xcolor}

\definecolor{myred}{HTML}{f67088}
\definecolor{myolive}{HTML}{96a331}
\definecolor{mygreen}{HTML}{35aca4}
\definecolor{mypurple}{HTML}{a38cf4}

\usepackage{xcolor}
\usepackage{tikz}

\definecolor{stepred}{HTML}{E3B8B8}
\definecolor{stepedge}{HTML}{445A78}  

\newcommand{\step}[1]{%
  \tikz[baseline=(n.base)]{
    \node[
      inner sep=0.5pt,
      minimum size=1.9ex,         
      circle,
      fill=stepred,               
      draw=stepedge,              
      line width=0.35pt,          
      text=white,
      font=\bfseries\scriptsize,
    ] (n) {#1};
  }%
}

\usepackage[cache=false]{minted} 
\setminted{
  fontsize=\small,
  linenos,
  numbersep=6pt,
  breaklines,
  tabsize=2,
  autogobble,
  frame=lines,
  framesep=2mm,
  baselinestretch=1.0
}
\usepackage{caption}
\usepackage[export]{adjustbox}

\definecolor{cvprblue}{rgb}{0.21,0.49,0.74}

\usepackage[pagebackref,breaklinks,colorlinks,allcolors=cvprblue]{hyperref}

\def\paperID{3667} 
\def\confName{CVPR}
\def\confYear{2026}

\title{Multi-Scale Local Speculative Decoding for Image Generation}

\author{Elia Peruzzo \qquad Guillaume Sautière \qquad Amirhossein Habibian \\
\small \textsuperscript{\textdagger}Qualcomm AI Research \\
{\tt\small\{eperuzzo,\,gsautie,\,ahabibia\}@qti.qualcomm.com}
}

\begin{document}
\maketitle
Autoregressive (AR) models have achieved remarkable success in image synthesis, yet their sequential nature imposes significant latency constraints. Speculative Decoding offers a promising avenue for acceleration, but existing approaches are limited by token-level ambiguity and lack of spatial awareness. In this work, we introduce \textbf{Mu}lti-Scale \textbf{Lo}cal Speculative Decoding (\methodname), a novel framework that combines multi-resolution drafting with spatially informed verification to accelerate AR image generation. Our method leverages a low-resolution drafter paired with an up-sampling step to propose candidate image tokens, which are then verified in parallel by a high-resolution target model. Crucially, we incorporate a local rejection and resampling mechanism, enabling efficient correction of draft errors by focusing on spatial neighborhoods rather than raster-scan resampling after the first rejection. When integrated with parallel decoding resampling, \methodname{} achieves substantial speedups -- up to $\mathbf{5\times}$ -- outperforming both speculative decoding and parallel decoding baselines in terms of acceleration, while maintaining comparable semantic alignment and perceptual quality. These results are validated using GenEval, DPG-Bench, and FID/HPSv2 on the MS-COCO 5k validation split. Extensive ablations highlight the impact of up-sampling design, probability pooling, and local rejection and resampling with neighborhood expansion. Our approach sets a new state-of-the-art in speculative decoding for image synthesis, bridging the gap between efficiency and fidelity.
Project page is available at \url{https://qualcomm-ai-research.github.io/mulo-sd-webpage/}.

\let\thefootnote\relax\footnotetext{\noindent\textsuperscript{\textdagger}Qualcomm AI Research is an initiative of Qualcomm Technologies, Inc.}

\begin{figure}[!t]
    \centering
    \includegraphics[width=\columnwidth]{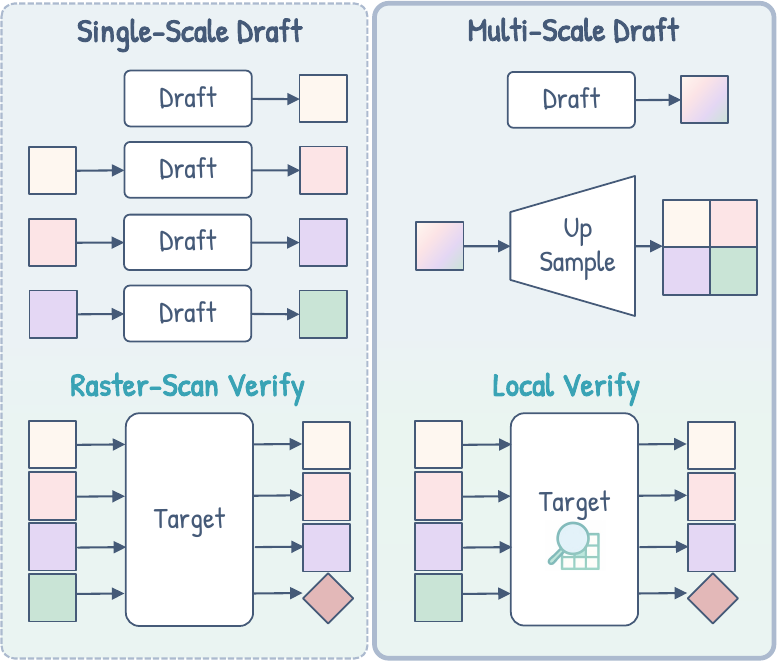}
    \caption{Multi-Scale Speculative Decoding extends speculative decoding by using a draft model working at a lower resolution than the target model, to enable acceleration through a coarse-to-fine approach. During verification, we exploit spatial locality in autoregressive models to resample only a neighborhood of rejected image tokens, improving efficiency without compromising quality.
    }
    \vspace{2em}
    \label{fig:teaser}
\end{figure}
    
\section{Introduction}
\label{sec:intro}

Recently unified multimodal large language models (MLLMs), merging the generation and understanding of language and vision in a unified autoregressive (AR) model, have seen a surge in popularity \cite{chameleonteam2025chameleon,liu2025lwm,liu2025luminamgpt,wang2024illume,wu2024janus,chen2025januspro,wang2024emu3,wu2025harmon,han2025tar}. Compared to diffusion models \cite{esser2024sd3,podell2023sdxl,chen2024pixartsigma,li2024playgroundv25,li2024hunyuandit,ramesh2021dalle,xie2025sana15efficientscaling}, unified MLLMs tend to perform better in text-to-image alignment tasks, and more generally in semantic understanding of complex prompts and knowledge-driven generation tasks~\cite{pan2025metaqueries}. 

Despite their success, a fundamental limitation persists: the sequential nature of AR decoding leads to high inference latency, especially for large-scale models and high-resolution outputs. Image and video synthesis with AR models is made harder due to the rapidly exploding sequence size, as the number of tokens grows quadratically with resolution and leads to thousands of tokens even for modest resolution like 1024p. 

By reformulating the objective from next-token prediction to next-scale prediction, autoregressive image generation can be significantly accelerated by sampling the image in a coarse-to-fine manner,~\ie starting from low-resolution samples and progressively refining them~\cite{tian2024var,voronov2025switti,han2025infinity,guo2025fastvar,ren2024mvar,li2025memoryefficientvar}. Despite these substantial efficiency gains, the next-scale prediction objective fundamentally differs from the next-token prediction used to train LLMs. This discrepancy hinders the adaptation of next-scale prediction AR models within unified MLLMs. Therefore, accelerating generation under the next-token prediction objective for AR models -- the focus of this paper -- remains an important and relatively underexplored problem.



Speculative Decoding (SD)~\cite{zhang2024selfspeculative}, originally developed for language models, introduces a draft-and-verify paradigm: a lightweight model (\emph{drafter}) proposes multiple tokens sampled sequentially and the full-size model (\emph{target)} verifies them in parallel. While this approach has shown impressive speedups in text generation, its application to image synthesis remains underexplored. Recent efforts such as LANTERN \cite{jang2025lantern,park2025lanternpp} have adapted speculative decoding to the visual domain by relaxing acceptance criteria to account for image token ambiguity in the latent space. However, these methods still operate at the token level and ignore the spatial structure and multi-scale nature of images. Additionally, locality-aware decoding strategies like ZipAR \cite{he2025zipar} and LPD~\cite{zhang2025lpd} demonstrate that exploiting spatial coherence can further reduce latency by enabling parallel generation across rows or patches.

We present \textbf{Mu}lti-Scale \textbf{Lo}cal \textbf{S}peculative \textbf{D}ecoding (\methodname{}, see \cref{fig:teaser}), a framework that exploits the structural properties of images to enhance speculative decoding. Our approach introduces three key innovations:

\begin{enumerate}
    \item \textbf{Multi-scale drafting}: We leverage the natural hierarchy of image resolutions by using a low-resolution drafter and an up-sampler to propose candidate image tokens, which are then verified by a high-resolution AR model.
    \item \textbf{Local verification}: Inspired by spatial coherence in images, we introduce a rejection and re-sampling mechanism that operates over local neighborhoods rather than full raster-scan sequences, improving both efficiency and acceptance rates.
    \item \textbf{Integration with Parallel Decoding}:  As an orthogonal component, we incorporate parallel decoding at the draft and resampling stages to reduce sequential steps without sacrificing quality of the generated tokens.
\end{enumerate}

We demonstrate that \methodname{} achieves substantial speedup -- up to $\mathbf{5\times}$ -- outperforming strong speculative decoding baselines such as EAGLE-2~\cite{li2024eagle2} and LANTERN~\cite{jang2025lantern} as well as parallel decoding methods like ZipAR \cite{he2025zipar}, while maintaining comparable semantic alignment and perceptual quality. These results are validated using GenEval~\cite{ghosh2023geneval}, DPG-Bench~\cite{hu2024dpgbench}, and FID~\cite{heusel2018fid}/HPSv2~\cite{wu2023hpsv2} on the MS-COCO 2017 5k validation split~\cite{lin2015coco}.

\section{Related art}
\label{sec:related-art}

\paragraph{Speculative decoding} methods aim to accelerate autoregressive generation by relaxing sequential dependencies. For text, Speculative Decoding~\cite{leviathan2023speculativedecoding} introduced a draft-and-verify scheme wherein a lightweight model proposes multiple tokens in sequence, and the target model verifies them in parallel --- achieving 2--3$\times$ speedups. Self-Speculative Decoding~\cite{zhang2024selfspeculative} reuses internal layers of the target model and hierarchical verification, reaching 3.5$\times$ acceleration without additional memory footprint. Medusa~\cite{cai2024medusa} employs multi-head decoding and tree attention for up to 3.6$\times$ speedup. EAGLE~\cite{li2025eagle1} drafts by making use of the target model's penultimate latent representations, while EAGLE-2~\cite{li2024eagle2} introduces dynamic draft trees based on token confidence, pushing speedups to 4.3$\times$. These methods are designed for text generation and do not generalize to the image domain.

LANTERN~\cite{jang2025lantern,park2025lanternpp} is the first to extend speculative decoding to image synthesis. It addresses token ambiguity in visual models by introducing a relaxed acceptance criterion based on latent token interchangeability. This improves acceptance rates while bounding total variation distance to preserve semantic fidelity, achieving 1.75--1.82$\times$ speedups over greedy decoding on LlamaGen~\cite{sun2024llamagen}.

\methodname{} is closely related to LANTERN in extending speculative decoding to images. Like LANTERN, it relaxes the verification objective to address token ambiguity in vision models. However, \methodname{} uniquely leverages the multi-scale prior to further improve decoding efficiency, making it the first speculative decoding method to do so.

\paragraph{Multi-scale} autoregressive models generate images in a coarse-to-fine manner, improving both efficiency and quality. VAR~\cite{tian2024var} introduced next-scale prediction, conditioning each resolution level on lower ones, and outperformed diffusion models in speed and fidelity. Follow-up works such as M-VAR~\cite{ren2024mvar}, Switti~\cite{voronov2025switti}, and others~\cite{guo2025fastvar,jiao2025flexvar,han2025infinity} extend this framework.
M-VAR decouples intra- and inter-scale modeling, combining bidirectional attention with linear-complexity mechanisms like Mamba~\cite{gu2024mamba}, achieving state-of-the-art FID with fewer parameters.
Switti~\cite{voronov2025switti} removes explicit cross-scale autoregression and classifier-free guidance at high resolutions, enabling up to 7$\times$ faster sampling with competitive quality.

These models align well with the hierarchical structure of visual data and demonstrate strong scalability. However, their bespoke sampling schedules hinders integration with next-token prediction frameworks and unified MLLMs, \eg, causing inefficient KV-cache usage and requiring ad-hoc designs~\cite{li2025memoryefficientvar,guo2025fastvar}.

\methodname{} shares the multi-scale design philosophy of these models but differs in its focus on decoding efficiency through speculative sampling. Unlike 
existing methods, which rely on custom sampling schedules, \methodname{} integrates well with next-token prediction MLLMs and inject the coarse-to-fine approach in its drafting strategy.

\paragraph{Locality-aware} autoregressive methods~\cite{he2025zipar,zhang2025lpd} leverage spatial coherence in images to improve generation efficiency. ZipAR~\cite{he2025zipar} is an inference-time trick which reduces the number of forward passes by up to 91\% with minimal quality degradation, outperforming prior parallel decoding methods like speculative Jacobi decoding~\cite{teng2025jacobidecoding}. ZipAR enables inter-row parallel decoding by exploiting spatial adjacency, allowing tokens in the next row to be decoded once sufficient context is available. 
Differently, LPD~\cite{zhang2025lpd} decouples the two roles tokens typically play, providing context and enabling generation. With separate query and context tokens they allow parallel and arbitrary order sampling of images, albeit requiring full re-training of the AR model.

\methodname{} exploits the locality of AR models exposed by ZipAR~\cite{he2025zipar} and LPD~\cite{zhang2025lpd} by performing re-sampling within local neighborhoods rather than in raster-scan order. In addition, Parallel Decoding is orthogonal to the draft-and-verify paradigm; in practice, it can be combined with \methodname{} — during low-resolution drafting and high-resolution resampling — to couple fast locality-driven sampling with the verification from the target model.

\section{Method}
\label{sec:method}

\begin{figure*}[!t]
    \centering
    \includegraphics[width=\textwidth]{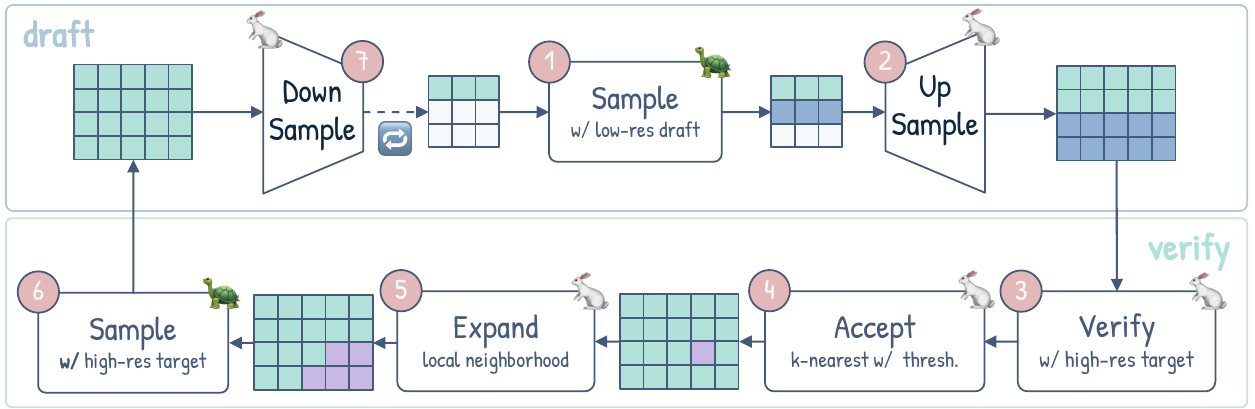}
    \caption{Overview of our proposed method Multi-Scale Local Speculative Decoding (\methodname{}). \colorbox{proposed!30}{Blue} indicates draft tokens, \colorbox{accepted!30}{green} accepted  tokens, \colorbox{rejected!30}{purple} rejected tokens, \colorbox{empty!30}{blank} placeholder tokens. \emojiturtle~indicates sequential operations, \emojirabbit~parallel operations, \emojirepeat~a drawing discontinuity due to looping.
    }
    \label{fig:main-method}
\end{figure*}

\subsection{Preliminaries}
\label{subsec:preliminaries}
Let $M_p$ denote the target autoregressive model, which defines a conditional probability distribution $p(x_t|x_{<t})$ over the next token $x_t$ given a prefix $x_{<t}$. Let $M_q$ denote the draft model, a more efficient model, that defines a distribution $q(x_t|x_{<t})$ for the same task.

Speculative Decoding~\cite{zhang2024selfspeculative} accelerates sampling from $M_p$ by leveraging $M_q$ to propose a sequence of $n$ draft tokens $\tilde{x}_0, \ldots, \tilde{x}_{n-1}$ sampled autoregressively from $q$. These drafts are then verified in parallel by $M_p$. Each token $\tilde{x}_i$ is accepted with probability:

\begin{equation}
\min\left(1, \frac{p_i(\tilde{x}_i)}{q_i(\tilde{x}_i)}\right),
\label{eq:spec_accept}
\end{equation}

where $p_i$ and $q_i$ denote the distributions from $M_p$ and $M_q$ conditioned on the prefix extended by previously accepted tokens. If a token is rejected, it is resampled from an adjusted distribution:

\begin{equation}
p'_i(x) = \text{norm}\left(\max\left(0, p_i(x) - q_i(x)\right)\right),
\label{eq:spec_resample}
\end{equation}

\noindent ensuring that the overall sampling process is exact \ie the same as sampling from the target distribution $p$.

In visual AR models, next-token distributions are often ambiguous (flatter), significantly reducing acceptance rate.
LANTERN~\cite{jang2025lantern} addresses this by pooling probability mass over the $k$ nearest codebook neighbors $B_k(\tilde{x}_i)$ and relaxing acceptance to
\begin{equation}
\label{eq:lantern_accept}
\alpha_i^{(k)} \;=\; \min\!\left(1,\; \frac{\sum_{x \in B_k(\tilde{x}_i)} p_i(x)}{q_i(\tilde{x}_i)}\right),
\end{equation}
while constraining the deviation from $p_i$ via a total variation distance (TVD) bound:
\begin{equation}
\label{eq:tvd_bound}
\mathrm{TVD}\big(p_i^{(k,\delta)},\, p_i\big) \;<\; \delta,
\end{equation}
where $p_i^{(k,\delta)}$ redistributes mass within $A_{k,\delta}(\tilde{x}_i)\subseteq B_k(\tilde{x}_i)$ to satisfy the bound.
This relaxation improves acceptance in domains with higher token uncertainty while controlling distributional drift.

\subsection{Multi-Scale Drafting}
\label{subsec:ms-sd}

Multi-scale modeling is a strong inductive bias in image synthesis, powering UNet~\cite{ronneberger2015unet}, VQ-VAE-2~\cite{razavi2019vqvae2}, and VAR~\cite{tian2024var}. It follows a coarse-to-fine strategy—implicitly present even in diffusion~\cite{dieleman2024spectral}—which we leverage to inject a multi-scale bias into speculative decoding.

Recent AR models for image synthesis~\cite{sun2024llamagen,han2025tar,liu2025luminamgpt} are commonly released at multiple resolutions, with separate finetuning for each scale. Given a desired target resolution $s_p$, we employ a draft model $M_q$ at lower resolution $s_q$, with resolution ratio $r = s_p / s_q$. The drafter is paired with an up-sampler $U_r$ and a down-sampler $D_r$. The target model $M_p$ operates at higher resolution $s_p$. 
An overview of the method is shown in \Cref{fig:main-method}, and refer to the supplementary material for a detailed algorithm and schematic comparison to related approaches.


Following Fig.~\ref{fig:main-method}, the process begins by sequentially sampling draft tokens from the low-resolution model ($\tilde{y} \sim M_q$, Step \step{1}). These tokens are then upsampled ($\tilde{x} = U_r(\tilde{y})$) to expand the sequence length by $r^2$ (Step \step{2}). In Step \step{3}, the target model $M_p$ verifies $\tilde{x}$ in parallel. Steps \step{4}–\step{5}, discussed in the next \Cref{subsec:lo-sd}, apply an acceptance rule to determine which tokens to keep. Rejected tokens are resampled sequentially using $M_p$ (Step \step{6}). Finally, verified tokens are appended to the accepted prefix to form $x$, which is downsampled to $y = D_r(x)$ (Step \step{7}). These downsampled tokens serve as the prefix for the next low-res draft sampling. The cycle repeats until $|x| = N$, the target sequence length.

Our method has three key differences with standard speculative decoding: (i) unlike speculative decoding, where the draft model typically proposes the next $n$ tokens without regard to resolution or image boundaries, our draft model generates full rows to help the up-sampler produce coherent high-resolution patches; (ii) all rejected tokens are re-sampled by the target model, which simplifies the down-sampler’s role to only processing verified tokens; and (iii) the draft model has the same capacity as the target model, so speedup comes from reducing the number of function evaluations (NFE) and exploiting the quadratic gap in sequence size between low- and high-resolution representations. While this design simplifies down-sampling, it introduces a bottleneck during inference due to sequential sampling within the target model. Consequently, achieving speedups comparable to LANTERN or speculative decoding requires higher acceptance rates.

\subsection{Local Verification}
\label{subsec:lo-sd}
Our initial experiments used the LANTERN rule~\cite{jang2025lantern} as described in \cref{eq:lantern_accept}, which rejects all draft tokens after the first rejected token in raster-scan order. However, because our framework requires re-sampling every rejected token with the target model, this approach resulted in low acceptance rates and negligible speedup.

To address this, we adopt a relaxed criterion: accept a draft token if the pooled probability over its neighborhood exceeds a threshold $\tau$ (Step \step{4} in \cref{fig:main-method}):
\begin{equation}
\text{Accept if } \sum_{x \in B_k(\tilde{x}_i)} p_i(x) \ge \tau.
\label{eq:relaxed-acceptance}
\end{equation}
\noindent Higher $\tau$ values yield a closer approximation to the target model but slower inference, while lower values trade accuracy for speed.

\begin{figure}[!t]
    \centering
    \includegraphics[width=\columnwidth]{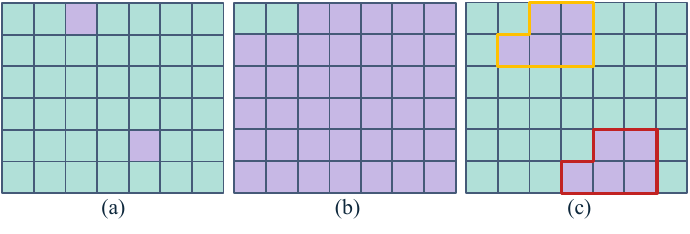}
    \caption{
    Representation of the local expansion rule. \colorbox{accepted!30}{Green} accepted and \colorbox{rejected!30}{purple} rejected tokens. (a) $R_t$, the set of rejected indices under the target mode as in \cref{eq:relaxed-acceptance}, (b) shows raster-scan rejection as in standard SD, (c) $R_X$, the newly introduced local expansion around rejected tokens $R_t$ with a radius $l=1$ as in \cref{eq:union-neighborhoods}. We outline with different colors the rejection islands that get sampled concurrently, see \Cref{subsec:parallel}.
    }
    \label{fig:locality-method}
\end{figure}

Visual AR models rely on localized attention, where token predictions are strongly influenced by nearby context and weakly by distant regions~\cite{zhang2025lpd,he2025zipar}. To exploit this, we introduce local expansion -- a strategy that re-samples only within a small neighborhood around rejected tokens. This targets areas with high local dependency while preserving distant accepted tokens, whose influence is minimal. It is illustrated in Step \step{5} of \cref{fig:main-method}, and compared to raster-scan rejection in \Cref{fig:locality-method}. Our ablation study (\cref{subsec:ablation}) confirms that omitting local expansion degrades performance, validating its necessity. This approach improves sampling efficiency without compromising perceptual quality.

Let $R_T=(t_0, \ldots, t_m)$ be the set of $m$ rejected indices under the target model following \cref{eq:relaxed-acceptance}. 
For any position $t \in R_T$, we define its local neighborhood of radius $l$ as:
\begin{equation}
\label{eq:neighborhood-definition}
N(t,l) = \Big\{ u \;\Big|\; |i_u - i_t| \le l,\; |j_u - j_t| \le l,\; u \ge t_0 \Big\},
\end{equation}
\noindent where $(i_u,j_u)$ and $(i_t,j_t)$ are the 2D coordinates of indices $u$ and $t$, and $t_0$ be the index of the first token rejected by the target model. The last condition ensures we do not revisit tokens before the first rejection.

We consider the set $R_X$ of all locally expanded rejected tokens:
\begin{equation}
\label{eq:union-neighborhoods}
R_{X} = \bigcup_{t \in R_T} N(t,l),
\end{equation}
and sequentially re-sample all positions in $R_X$ using the target model $M_p$ (illustrated in Step \step{6} of \cref{fig:main-method}),. 

\subsection{Integration with Parallel Decoding}
\label{subsec:parallel}
\methodname{} is dominated by two sequential stages: (i) low-resolution drafting (Step~\step{1}) and (ii) target-model resampling after verification (Step~\step{6}).
We mitigate both by incorporating \emph{parallel decoding} strategies, see \emph{Supp. Mat.}\,for a detailed latency analysis.

\paragraph{Parallel low-resolution drafting.} We draft the low-resolution image at once using ZipAR \cite{he2025zipar} instead of drafting it iteratively row-by-row. In turn, we fully exploit parallel decoding for the draft tokens.

\paragraph{Parallel resampling via rejection islands.}
After verification and local expansion, we obtain the set of target positions to resample, $R_X$, from \cref{eq:union-neighborhoods}. We group these positions into \emph{rejection islands}, defined as 8-connected components on the 2D token grid:

\begin{equation} 
\begin{split}
\{\mathcal{I}_m\}_{m=1}^{M} & \;=\; \mathrm{CC}_8(R_X) \\ \qquad
R_X & =\bigcup_{m=1}^{M}\mathcal{I}_m,\;\; \mathcal{I}_m\cap\mathcal{I}_{m'}=\emptyset~(m\neq m').
\label{eq:islands}
\end{split}
\end{equation}
where $\mathrm{CC}_8$ is a connected component algorithm \cite{grana2010optimized}. We illustrate these rejection islands in \Cref{fig:locality-method}.

We then resample \emph{across islands} in parallel (different $\mathcal{I}_m$ are processed concurrently). This design further leverages the locality of the AR visual model, as we exploit the accepted tokens as a valid and sufficient context to resample different islands concurrently. This is the key difference compared to naive ZipAR. Furthermore, \emph{within each island} we apply the ZipAR decoding strategy to reduce the number of sequential target-model steps required to fill in the missing tokens.

Overall, our pipeline combines parallel decoding for low-resolution drafting, batched verification, and island-wise parallel resampling, yielding significant end-to-end latency reductions in practice.
\section{Experiments}
\label{sec:experiments}

\subsection{Implementation Details}
\label{subsec:implementation}
Our primary evaluations use Tar-1.5B and Tar-7B~\cite{han2025tar}, multimodal LLMs equipped with the AR-DTok generative detokenizer for fully autoregressive text-to-image generation in a discrete VQ-VAE latent space. Tar provides resolution-specific AR-DTok checkpoints at 256p, 512p, and 1024p. To assess generality beyond MLLMs, we also evaluate LlamaGen (XL)~\cite{sun2024llamagen}, a representative autoregressive text-to-image generator with 256p and 512p checkpoints.

Unless otherwise noted, we use the 256p autoregressive (AR) checkpoint of the base model as the \emph{drafter}, and the AR model at the target resolution as the \emph{verifier}. The drafter is paired with upsamplers to enable $2\times$ (512p) and $4\times$ (1024p) generation. We consider two upsampler designs: (i) a learnable latent-space upsampler and (ii) a pre-trained pixel-space upsampler.
\paragraph{Latent-space upsampler}
The up- and downsamplers are lightweight convolutional networks built from residual blocks~\cite{he2016resnet} with pixel-shuffle resampling~\cite{shi2016pixelshuffle}. To preserve the autoregressive order, all convolutions are masked to be row-causal. These modules are inexpensive and support iterative speculative decoding, but they are model-specific (tied to the choice of VQ model) and require training.
\paragraph{Pixel-space upsampler}
We also explore a \emph{training-free} path using off-the-shelf pixel-space super-resolution. Concretely: (i) draft the full low-resolution image; (ii) decode tokens to pixels with the VQ decoder; (iii) upsample in pixel space~\cite{zhang2024hitsr}; and (iv) re-encode to tokens to obtain the high-resolution draft. This variant is model-agnostic and requires no additional training, at the cost of higher FLOPs and potential loss of token-space efficiency. In practice, this variant is competitive and model-agnostic. It pairs well with our parallel decoding integration (see \cref{subsec:parallel}) and is used as the default throughout, except where stated. We refer to the Supp. Mat. for additional details.

\begin{table*}[!ht]
\centering
\caption{Comparison of decoding speedup versus GenEval~\cite{ghosh2023geneval}, DPG-Bench~\cite{hu2024dpgbench}, and perceptual metrics (FID~\cite{heusel2018fid} and HPSv2~\cite{wu2023hpsv2}), computed on the MS-COCO 5k validation split~\cite{lin2015coco}. We sweep the acceptance threshold $\tau$ in \methodname{} and report the operating point that more closely matches LANTERN’s GenEval score.
}

\begin{tabular}{llcllll}
\toprule

 & \multirow{2}{0em}{\textbf{Method}} & \multicolumn{1}{c|}{\textbf{Efficiency}} & \multicolumn{2}{c|}{\textbf{Semantic Alignment}} & \multicolumn{2}{c}{\textbf{Perceptual Quality}} \\
 \cmidrule{3-7}
 & & \multicolumn{1}{c|}{Speedup ($\uparrow$)} & GenEval ($\uparrow$) & \multicolumn{1}{l|}{DPG-Bench ($\uparrow$)} & FID ($\downarrow$) & HPSv2 $(\uparrow)$\\
\midrule

\multirow{15}{0.5em}{\rotatebox{90}{512p}} & LlamaGen-XL  &  1.00\ttimes & 37.1 & 65.1  & 56.2 & 23.1 \\
& \quad + ZipAR-16 \cite{he2025zipar} & \textbf{1.88}\ttimes & 37.0 \tiny(\textcolor{gray}{-0.1}) & 64.9 \tiny(\textcolor{gray}{-0.2}) & 55.1 \tiny(\textcolor{gray}{-1.1})& 23.0 \tiny(\textcolor{gray}{-0.1}) \\
& \quad + EAGLE-2  \cite{li2024eagle2} & 0.96\ttimes & 37.1 \tiny(\textcolor{gray}{-0.0}) & 65.1 \tiny(\textcolor{gray}{-0.0}) & 56.2 \tiny(\textcolor{gray}{-0.0}) &  23.1 \tiny(\textcolor{gray}{-0.0}) \\
& \quad + LANTERN \cite{jang2025lantern} & \underline{1.59}\ttimes & 36.3 \tiny(\textcolor{gray}{-0.8}) & 64.5 \tiny(\textcolor{gray}{-0.6}) & 55.4 \tiny(\textcolor{gray}{-0.8}) & 22.2  \tiny(\textcolor{gray}{-0.9}) \\
 & \cellcolor{mypurple!10}\quad + \textbf{\methodname{} (2\ttimes)}  & \cellcolor{mypurple!10}1.40\ttimes & \cellcolor{mypurple!10}36.1 \tiny(\textcolor{gray}{-1.0})& \cellcolor{mypurple!10}64.0 \tiny(\textcolor{gray}{-1.1}) & \cellcolor{mypurple!10}54.3 \tiny(\textcolor{gray}{-1.6}) & \cellcolor{mypurple!10}23.1 \tiny(\textcolor{gray}{-0.0}) \\

\cmidrule{2-7}

 & Tar-1.5B    & 1.00\ttimes  &  77.7  &  82.8   &  33.0  &  28.6   \\
& \quad + ZipAR-16 \cite{he2025zipar}          & \underline{1.88}\ttimes  &  76.6 \tiny(\textcolor{gray}{-1.1})  &  82.8 \tiny(\textcolor{gray}{-0.0})  &  33.0 \tiny(\textcolor{gray}{-0.0}) &   28.5  \tiny(\textcolor{gray}{-0.1})   \\
& \quad + EAGLE-2  \cite{li2024eagle2}         & 0.72\ttimes  &  77.7 \tiny(\textcolor{gray}{-0.0}) &  82.8 \tiny(\textcolor{gray}{-0.0})   &  33.0 \tiny(\textcolor{gray}{-0.0})  &  28.6 \tiny(\textcolor{gray}{-0.0})  \\
& \quad + LANTERN \cite{jang2025lantern}           & 1.08\ttimes  &  75.9 \tiny(\textcolor{gray}{-1.8})  &  82.1 \tiny(\textcolor{gray}{-0.9})  &  32.7 \tiny(\textcolor{gray}{-0.3})  &  27.7  \tiny(\textcolor{gray}{-0.8})  \\
  & \cellcolor{mypurple!10}\quad + \textbf{\methodname{} (2\ttimes)} & \cellcolor{mypurple!10}\textbf{1.94}\ttimes  &  \cellcolor{mypurple!10}76.4 \tiny(\textcolor{gray}{-1.3})  & \cellcolor{mypurple!10}82.6 \tiny(\textcolor{gray}{-0.2}) & \cellcolor{mypurple!10}33.4 \tiny(\textcolor{gray}{+0.4}) & \cellcolor{mypurple!10}28.3 \tiny(\textcolor{gray}{-0.3}) \\
 
\cmidrule{2-7}
 & Tar-7B  & 1.00\ttimes & 85.1 & 81.3 & 38.6 & 29.8 \tiny(\textcolor{gray}{-0.0}) \\
& \quad + ZipAR-16 \cite{he2025zipar}  & \underline{1.88}\ttimes &  85.3 \tiny(\textcolor{gray}{+0.2}) & 81.0 \tiny(\textcolor{gray}{-0.3}) & 38.7 \tiny(\textcolor{gray}{+0.1}) & 29.8 \tiny(\textcolor{gray}{-0.0}) \\
& \quad + EAGLE-2  \cite{li2024eagle2} & 0.76\ttimes & 85.1 \tiny(\textcolor{gray}{-0.0})& 81.3 \tiny(\textcolor{gray}{-0.0})& 38.6 \tiny(\textcolor{gray}{-0.0}) & 29.8 \tiny(\textcolor{gray}{-0.0}) \\
& \quad + LANTERN \cite{jang2025lantern} & 1.20\ttimes & 84.9 \tiny(\textcolor{gray}{-0.2}) & 80.5 \tiny(\textcolor{gray}{-0.8}) & 36.9 \tiny(\textcolor{gray}{-1.8}) & 28.7 \tiny(\textcolor{gray}{-0.9})   \\
& \cellcolor{mypurple!10}\quad + \textbf{\methodname{} (2\ttimes)}  & \cellcolor{mypurple!10}\textbf{2.03}\ttimes & \cellcolor{mypurple!10}85.1 \tiny(\textcolor{gray}{-0.0}) & \cellcolor{mypurple!10}80.8 \tiny(\textcolor{gray}{-0.5}) & \cellcolor{mypurple!10}38.2 \tiny(\textcolor{gray}{-0.4}) & \cellcolor{mypurple!10}29.5 \tiny(\textcolor{gray}{-0.3}) \\

\midrule
\multirow{10}{0.5em}{\rotatebox{90}{1024p}} & Tar-1.5B     & 1.00\ttimes  &  77.1  &  82.3   &  32.4   & 29.5      \\
& \quad + ZipAR-16 \cite{he2025zipar} & \underline{3.65}\ttimes  &  76.6 \tiny(\textcolor{gray}{-0.5})  &   82.5 \tiny(\textcolor{gray}{+0.2})  &  32.4 \tiny(\textcolor{gray}{-0.0}) &   29.6 \tiny(\textcolor{gray}{+0.1})    \\
& \quad + EAGLE-2 \cite{li2024eagle2} & 0.78\ttimes  &  77.1 \tiny(\textcolor{gray}{-0.0})  &  82.3 \tiny(\textcolor{gray}{-0.0})  & 32.4 \tiny(\textcolor{gray}{-0.0})     & 29.5 \tiny(\textcolor{gray}{-0.0})   \\
& \quad + LANTERN \cite{jang2025lantern}  & 1.42\ttimes  &  75.4 \tiny(\textcolor{gray}{-1.7})  &   82.3 \tiny(\textcolor{gray}{-0.0})     &  31.1 \tiny(\textcolor{gray}{-1.3})  & 28.5 \tiny(\textcolor{gray}{-1.0})       \\
& \cellcolor{mypurple!10}\quad + \textbf{\methodname{} (4\ttimes)} & \cellcolor{mypurple!10}\textbf{3.90}\ttimes  & \cellcolor{mypurple!10}76.8 \tiny(\textcolor{gray}{-0.3}) & \cellcolor{mypurple!10}82.2 \tiny(\textcolor{gray}{-0.1})  & \cellcolor{mypurple!10}31.3 \tiny(\textcolor{gray}{+0.4})  & \cellcolor{mypurple!10}28.7 \tiny(\textcolor{gray}{-0.3})\\

\cmidrule{2-7}
 & Tar-7B   & 1.00\ttimes & 85.2  & 80.4 & 37.9 & 30.5 \\
& \quad + ZipAR-16 \cite{he2025zipar} & \underline{3.65}\ttimes & 85.2 \tiny(\textcolor{gray}{-0.0})  & 80.3 \tiny(\textcolor{gray}{-0.1})& 37.9 \tiny(\textcolor{gray}{-0.0}) & 30.5 \tiny(\textcolor{gray}{-0.0}) \\
& \quad + EAGLE-2  \cite{li2024eagle2} & 0.83\ttimes & 85.2 \tiny(\textcolor{gray}{-0.0}) & 80.4 \tiny(\textcolor{gray}{-0.0}) & 37.9 \tiny(\textcolor{gray}{-0.0}) & 30.5 \tiny(\textcolor{gray}{-0.0}) \\
& \quad + LANTERN \cite{jang2025lantern} & 1.45\ttimes & 82.9 \tiny(\textcolor{gray}{-2.3}) & 80.5 \tiny(\textcolor{gray}{+0.1})& 34.6 \tiny(\textcolor{gray}{-3.3}) & 29.4 \tiny(\textcolor{gray}{-0.9}) \\
& \cellcolor{mypurple!10}\quad + \textbf{\methodname{} (4\ttimes)}   & \cellcolor{mypurple!10}\textbf{5.33}\ttimes & \cellcolor{mypurple!10}85.4 \tiny(\textcolor{gray}{+0.2}) & \cellcolor{mypurple!10}80.8 \tiny(\textcolor{gray}{+0.4}) & \cellcolor{mypurple!10}34.8 \tiny(\textcolor{gray}{-3.1}) & \cellcolor{mypurple!10}29.5 \tiny(\textcolor{gray}{-0.8}) \\
\bottomrule
\vspace{0.5em}
\end{tabular}
\label{tab:sota}
\end{table*}

\subsection{Setting}
\paragraph{Baselines} 
We ported the official implementation of ZipAR~\cite{he2025zipar} into the Tar and Llamagen codebase and used it as a training-free parallel decoding method. Next, we adopted the official LANTERN~\cite{jang2025lantern} repository, which supports both LANTERN and EAGLE-2~\cite{li2024eagle2}. For Tar, we trained two drafter models -- one for 512p and one for 1024p -- using the provided scripts, adapting them to operate in the new latent space. We set the LANTERN hyperparameters to $k=1000$ (the codebook search space) and $\delta=0.4$ (TVD threshold), following the configuration reported in the original paper. For details, please refer to the  Supp. Mat.

\paragraph{Metrics} 
We evaluate all methods along three key dimensions: decoding efficiency, semantic alignment, and perceptual quality.

\noindent \textbullet~\emph{Decoding efficiency} is measured with the speedup \ie, the ratio between the latency of the baseline sequential decoding and that of the evaluated method. Values greater than 1 indicate acceleration, while values below 1 reflect a slowdown.
Latency is measured in seconds using PyTorch CUDA events on an NVIDIA A100 GPU, with a batch size of 1 and image resolutions of either 512p or 1024p, depending on the experiment. 

\noindent \textbullet~\emph{Semantic alignment} between text prompts and generated images is evaluated using GenEval~\cite{ghosh2023geneval} and DPG-Bench~\cite{hu2024dpgbench}, two recent benchmarks designed to measure multimodal consistency and grounding. 

\noindent \textbullet~\emph{Perceptual quality} is assessed using Fréchet Inception Distance (FID)~\cite{heusel2018fid}, which quantifies the distributional similarity between generated and real images, and Human Preference Score v2 (HPSv2)~\cite{wu2023hpsv2}, a learned metric that approximates human judgments of image quality.

\paragraph{Datasets} 
The latent-space upsampler modules of \methodname{}, as well as the drafter models used in LANTERN, are trained on the LAION-COCO-Aesthetic dataset~\cite{li2024laioncocoaesthetic}. 
For evaluation, we compute FID and HPSv2 on the MS-COCO 2017 validation split (5k images)~\cite{lin2015coco}.
\label{sec:results}
\subsection{Main Results}
\paragraph{Quantitative Evaluation} 
\begin{table*}[!ht]
    \centering
    \def\arraystretch{1.0}
    \resizebox{\linewidth}{!}{
    \setlength\tabcolsep{0pt}
    \footnotesize
    \renewcommand{\arraystretch}{0.5}
    \begin{tabular}{cccc@{\hskip 1mm}cccc}
         \scriptsize Tar & \scriptsize ZipAR-16 & \scriptsize LANTERN & \scriptsize \methodname & 
         \scriptsize Tar & \scriptsize ZipAR-16 & \scriptsize LANTERN & \scriptsize \methodname \\
         \includegraphics[trim=2 0 0 2,clip,width=0.2\columnwidth]{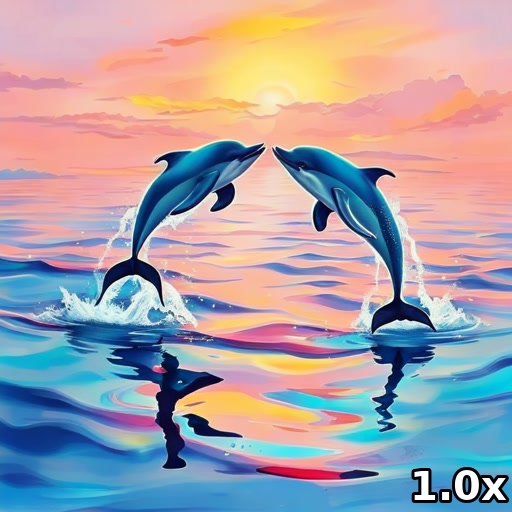} & 
         \includegraphics[trim=2 0 0 2,clip,width=0.2\columnwidth]{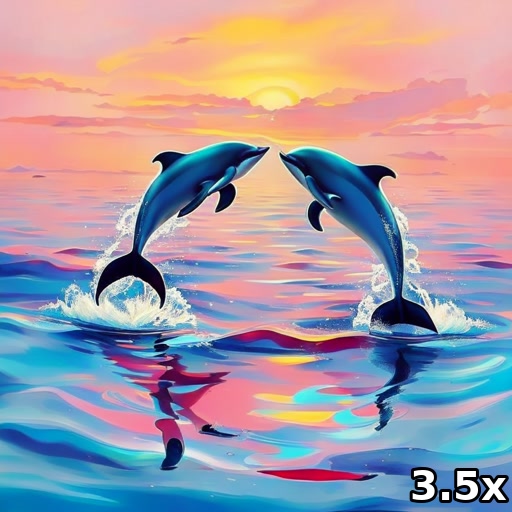} & 
         \includegraphics[trim=2 0 0 2,clip,width=0.2\columnwidth]{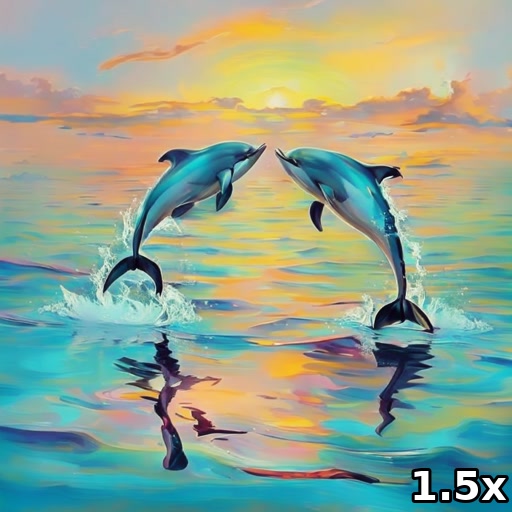} & 
         \includegraphics[trim=2 0 0 2,clip,width=0.2\columnwidth]{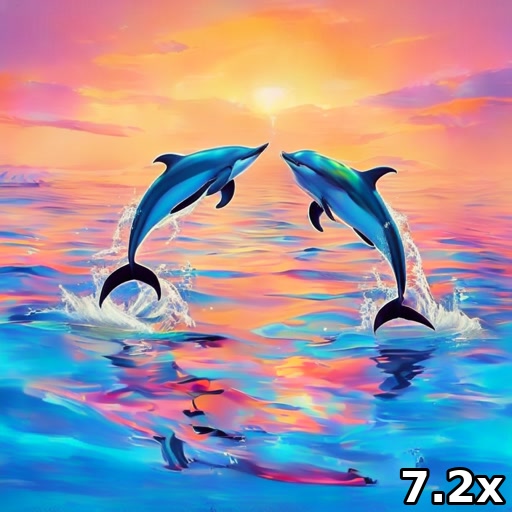} &
         
        \includegraphics[trim=2 0 0 2,clip,width=0.2\columnwidth]{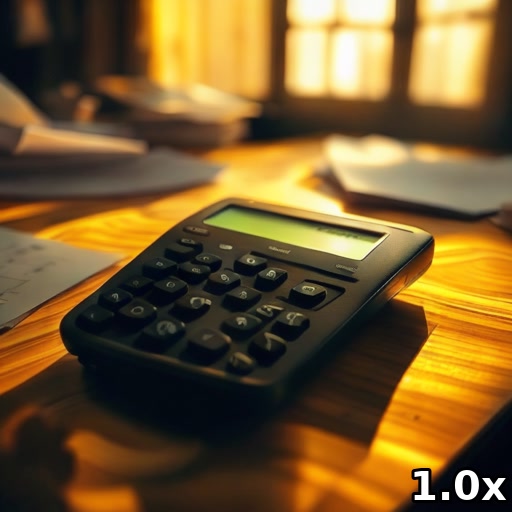} & 
         \includegraphics[trim=2 0 0 2,clip,width=0.2\columnwidth]{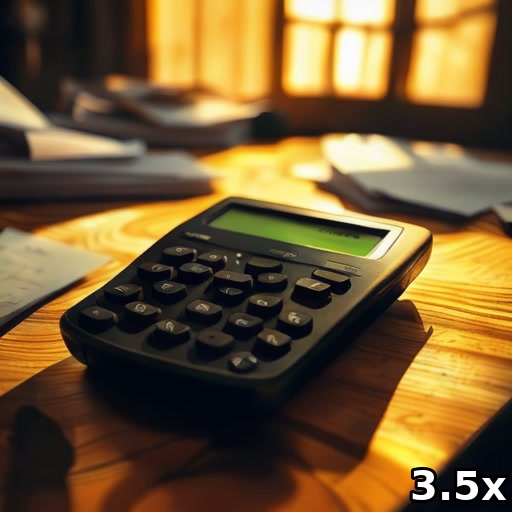} & 
         \includegraphics[trim=2 0 0 2,clip,width=0.2\columnwidth]{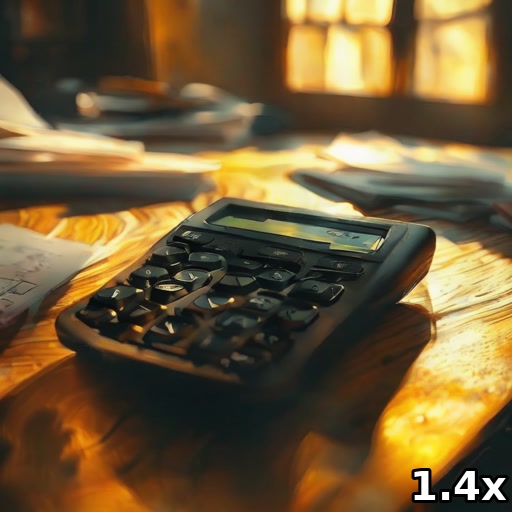} & 
         \includegraphics[trim=2 0 0 2,clip,width=0.2\columnwidth]{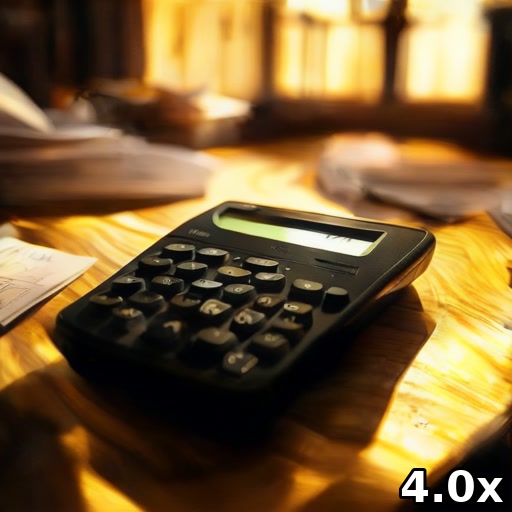} \\

        \includegraphics[trim=2 0 0 2,clip,width=0.2\columnwidth]{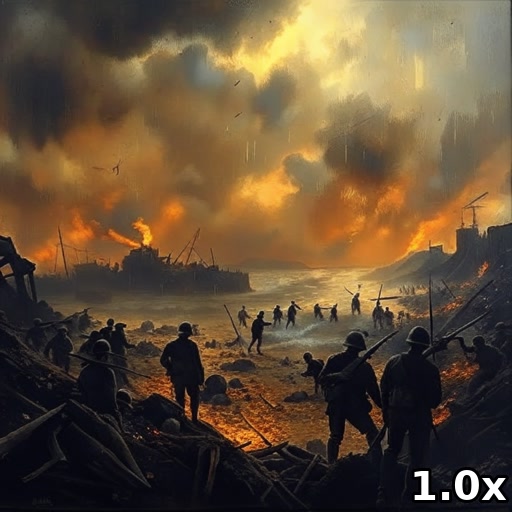} &
         \includegraphics[trim=2 0 0 2,clip,width=0.2\columnwidth]{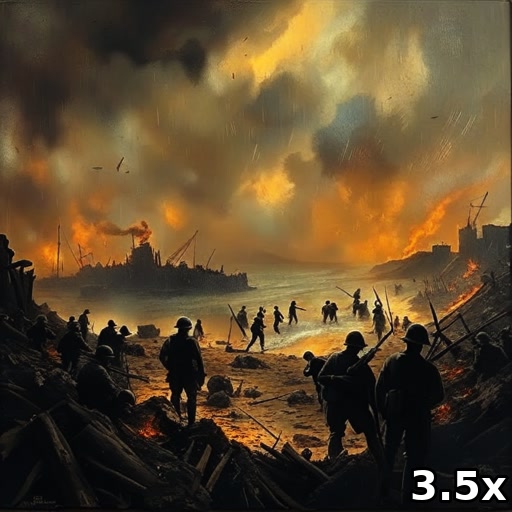} & 
         \includegraphics[trim=2 0 0 2,clip,width=0.2\columnwidth]{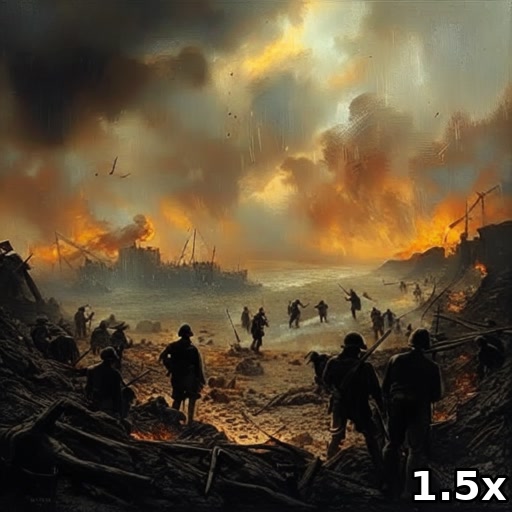} & 
         \includegraphics[trim=2 0 0 2,clip,width=0.2\columnwidth]{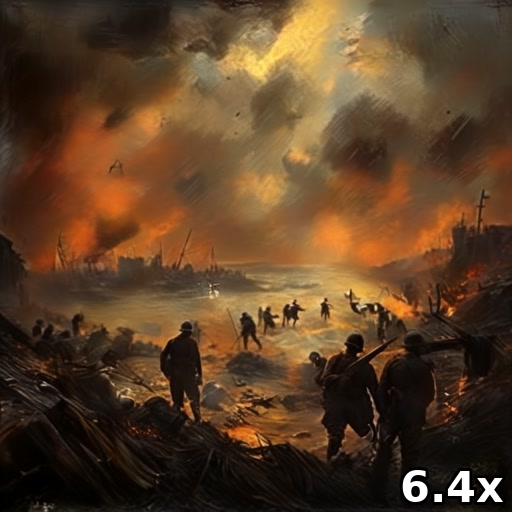} &
         
        \includegraphics[trim=2 0 0 2,clip,width=0.2\columnwidth]{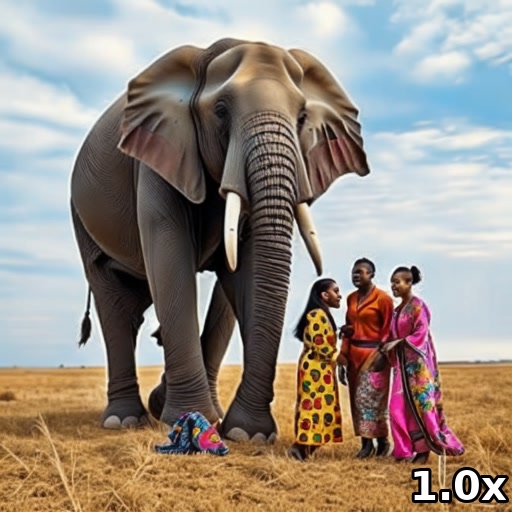} & 
         \includegraphics[trim=2 0 0 2,clip,width=0.2\columnwidth]{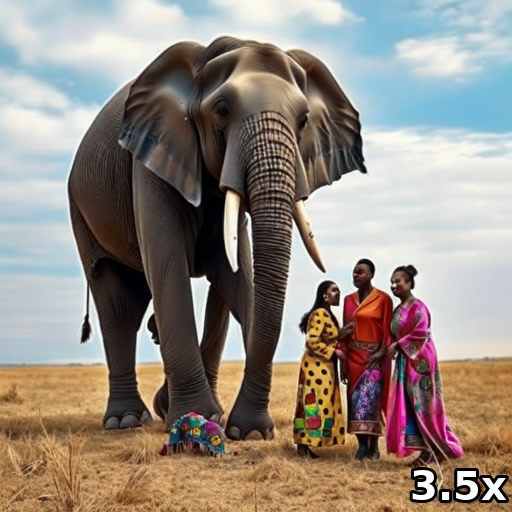} & 
         \includegraphics[trim=2 0 0 2,clip,width=0.2\columnwidth]{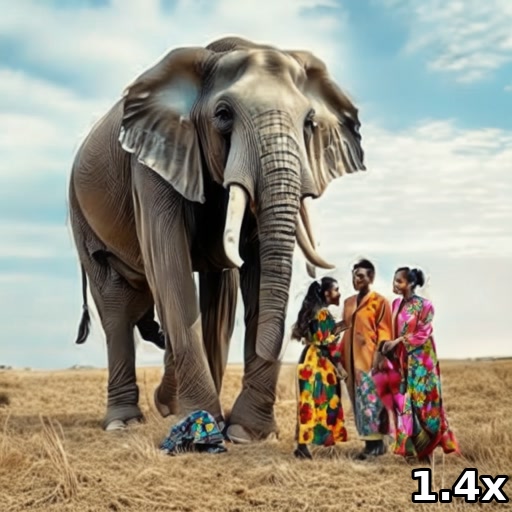} & 
         \includegraphics[trim=2 0 0 2,clip,width=0.2\columnwidth]{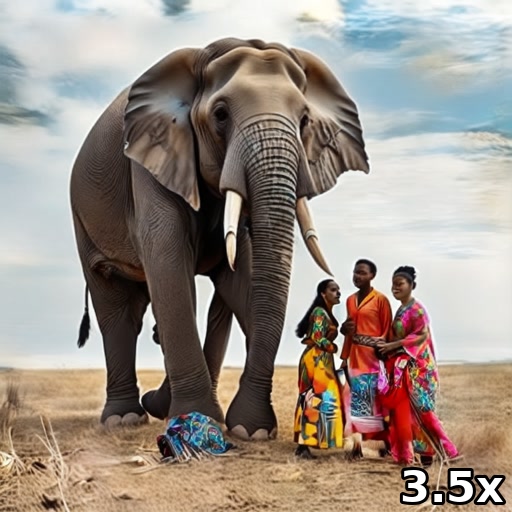} \\

        \includegraphics[trim=2 0 0 2,clip,width=0.2\columnwidth]{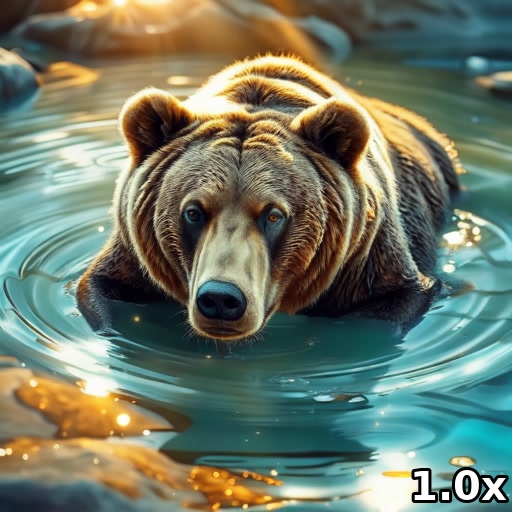} & 
         \includegraphics[trim=2 0 0 2,clip,width=0.2\columnwidth]{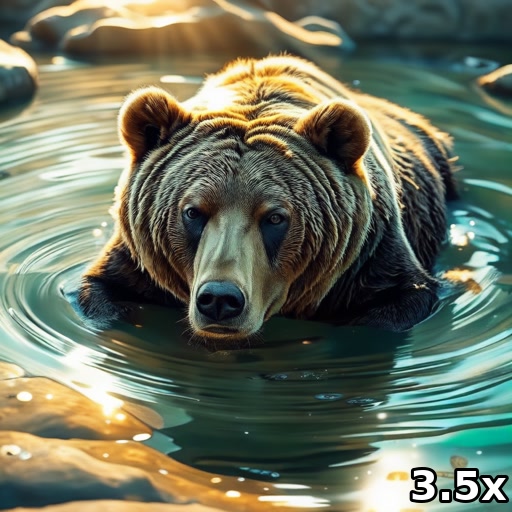} & 
         \includegraphics[trim=2 0 0 2,clip,width=0.2\columnwidth]{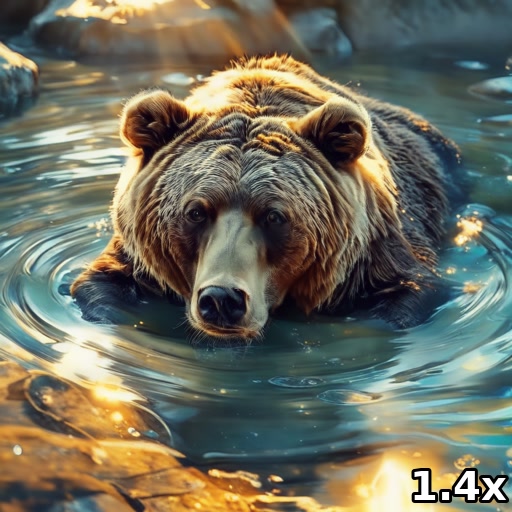} & 
         \includegraphics[trim=2 0 0 2,clip,width=0.2\columnwidth]{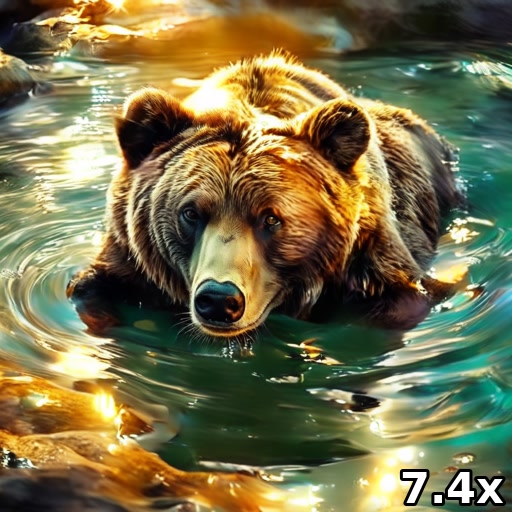} &
         
        \includegraphics[trim=2 0 0 2,clip,width=0.2\columnwidth]{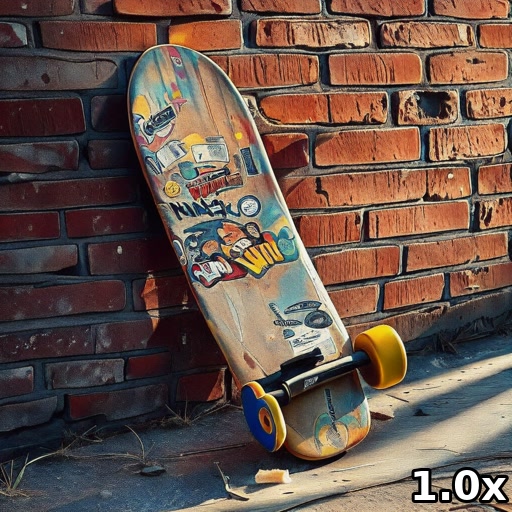} & 
         \includegraphics[trim=2 0 0 2,clip,width=0.2\columnwidth]{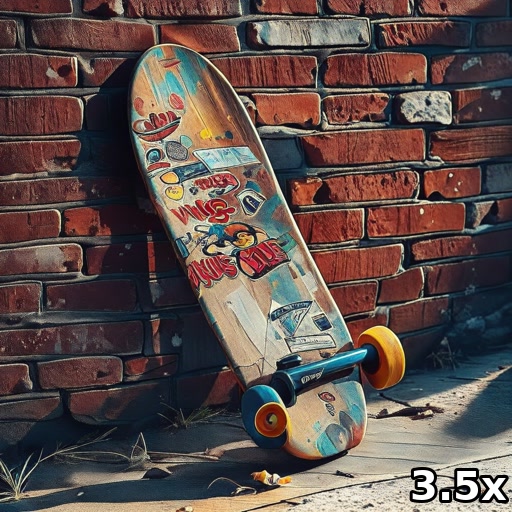} & 
         \includegraphics[trim=2 0 0 2,clip,width=0.2\columnwidth]{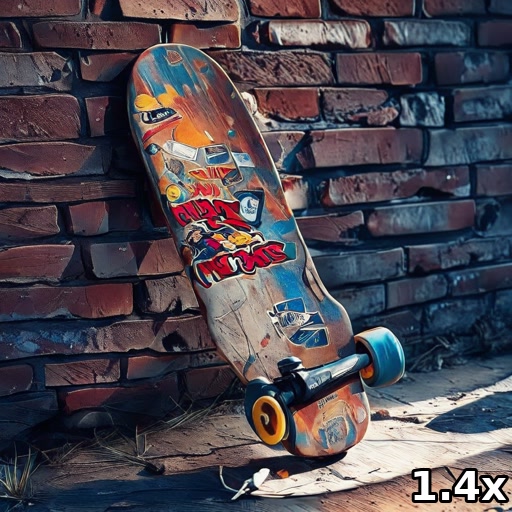} & 
         \includegraphics[trim=2 0 0 2,clip,width=0.2\columnwidth]{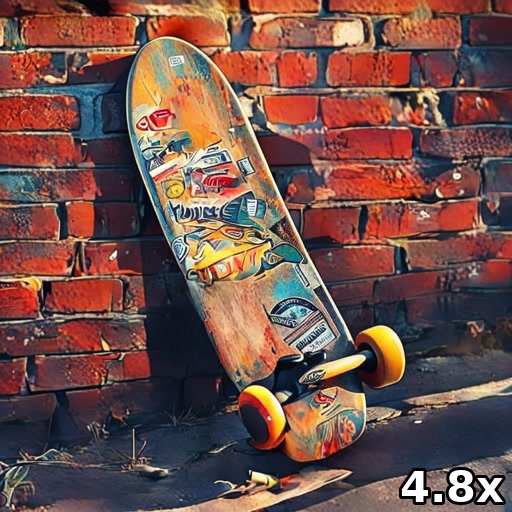} \\

              \includegraphics[trim=2 0 0 2,clip,width=0.2\columnwidth]{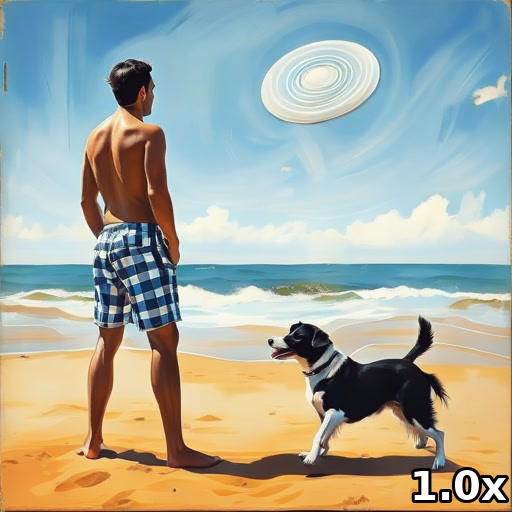} & 
         \includegraphics[trim=2 0 0 2,clip,width=0.2\columnwidth]{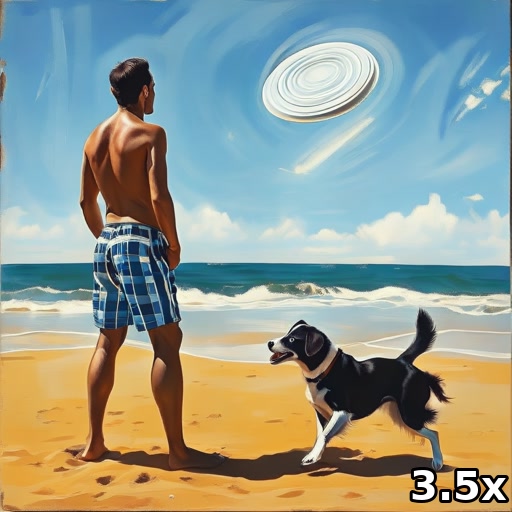} & 
         \includegraphics[trim=2 0 0 2,clip,width=0.2\columnwidth]{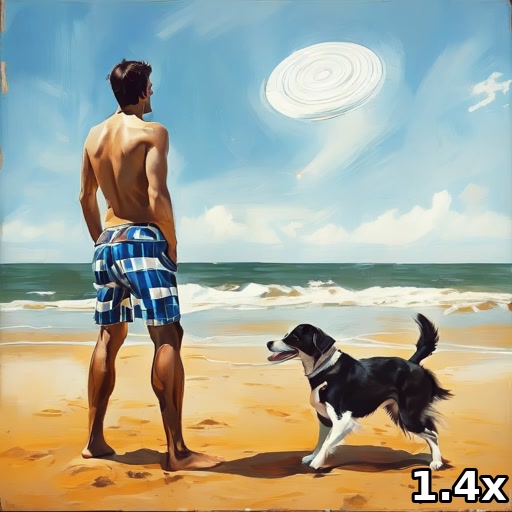} & 
         \includegraphics[trim=2 0 0 2,clip,width=0.2\columnwidth]{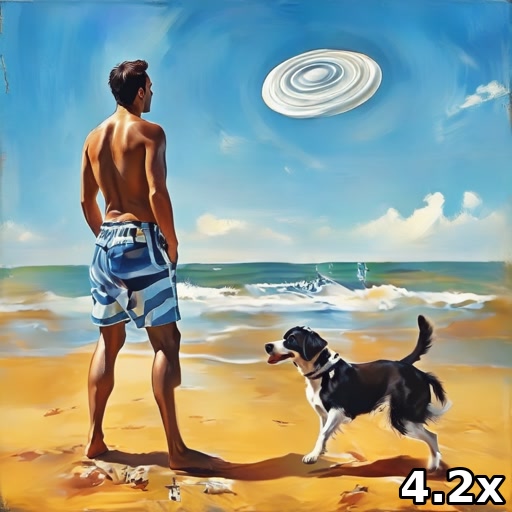} &
         
        \includegraphics[trim=2 0 0 2,clip,width=0.2\columnwidth]{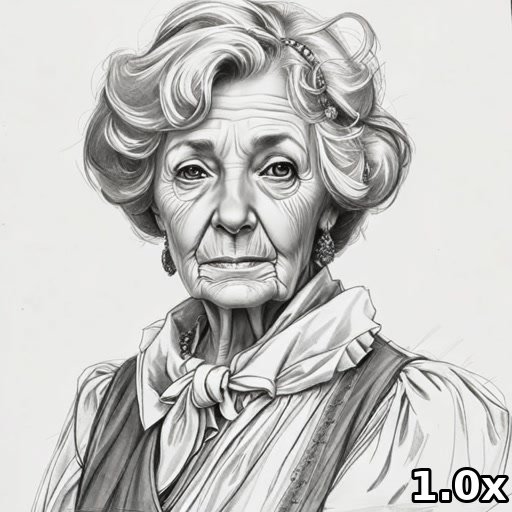} & 
         \includegraphics[trim=2 0 0 2,clip,width=0.2\columnwidth]{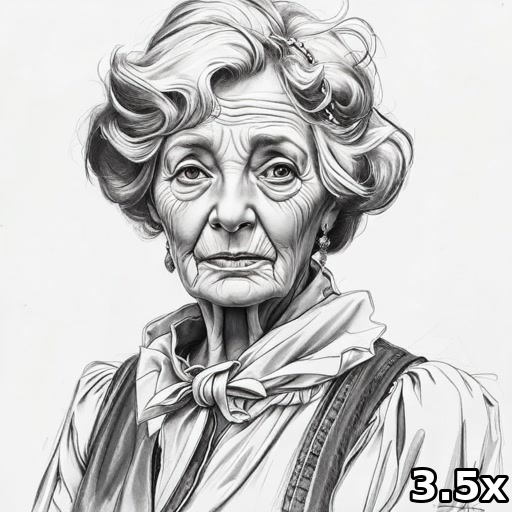} & 
         \includegraphics[trim=2 0 0 2,clip,width=0.2\columnwidth]{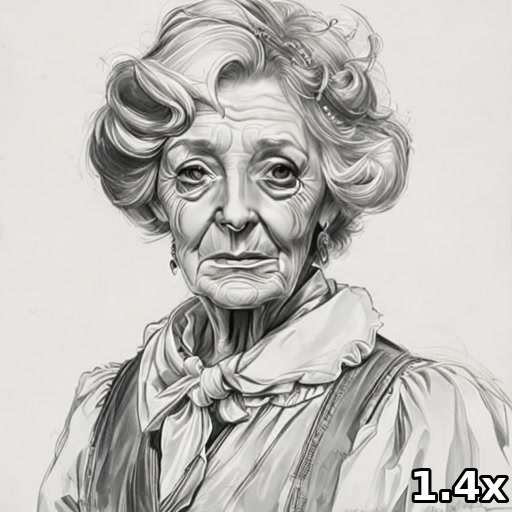} & 
         \includegraphics[trim=2 0 0 2,clip,width=0.2\columnwidth]{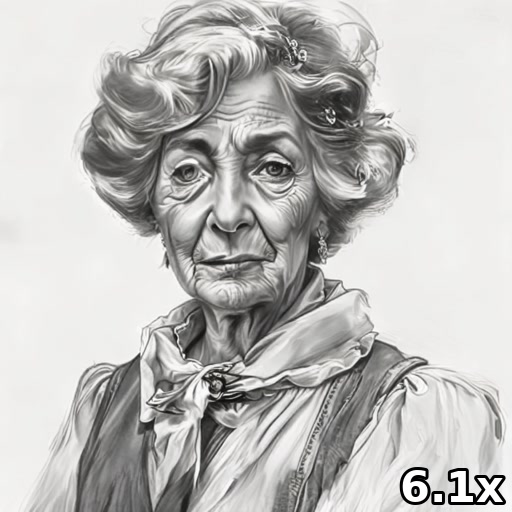} \\

    \end{tabular}
    }
    \captionof{figure}{Visual comparison of 1024p image generations. Each example shows its speedup over the base Tar-1.5B model (bottom-right). Outputs from EAGLE-2 are omitted since, as an exact decoding method, they match the base model.} 
    \label{fig:qualitative_figure}
\end{table*}

In \cref{tab:sota}, we compare \methodname{} against several baselines: ZipAR~\cite{he2025zipar}, a parallel decoding method for image generation; EAGLE-2~\cite{li2024eagle2}, a standard speculative decoding method; and LANTERN~\cite{jang2025lantern}, a speculative approach tailored for images. 

We structure our comparison across two resolutions: 512p and 1024p. Standard Speculative Decoding methods such as EAGLE-2 perform poorly on image data, often resulting in \emph{negative} speedups due to low acceptance rates caused by token ambiguity. LANTERN relaxes the acceptance criterion and achieves latency improvements, at the cost of small degradation in the metrics. We observe lower speedups when applying LANTERN to Tar compared to those reported with LlamaGen, likely because Tar is a significantly stronger model (\eg, GenEval 77.7\% vs. 37.1\% \cite{sun2024llamagen}), making its distribution harder to approximate. We discuss this in more detail in the Supp. Mat..

We sweep the acceptance threshold $\tau$ in \methodname{} to match the GenEval scores achieved by LANTERN. Under similar or better scores, \methodname{} consistently delivers greater speedups, ranking as the best method overall. At 1024p, our method incurs a slight drop in metrics but achieves between \textbf{4-5\ttimes{} faster end-to-end generation} compared to standard sampling.

\paragraph{Qualitative Evaluation} We provide a qualitative assessment in \Cref{fig:qualitative_figure}, using the same operating points as those reported in \Cref{tab:sota}. The visual comparisons highlight the perceptual quality of outputs generated by \methodname{}, LANTERN, and ZipAR under similar GenEval scores. 
Overall, \methodname{} achieves image quality comparable to LANTERN while consistently delivering higher speedups. This is particularly evident in complex scenes, such as the calculator in the first row, where our multi-scale formulation proves more effective at maintaining structural coherence. 
We also observe that our method performs robustly across a range of visual patterns, including textures, object boundaries, and semantic layouts and different styles from photorealistic to cartoonish (see Supp. Mat. for high-resolution samples and extended comparison). 

\subsection{Ablation Studies}
\label{subsec:ablation}

\begin{figure*}[htbp]
    \centering
    \begin{subfigure}[b]{0.24\textwidth}
    \centering
    \includegraphics[width=\textwidth]{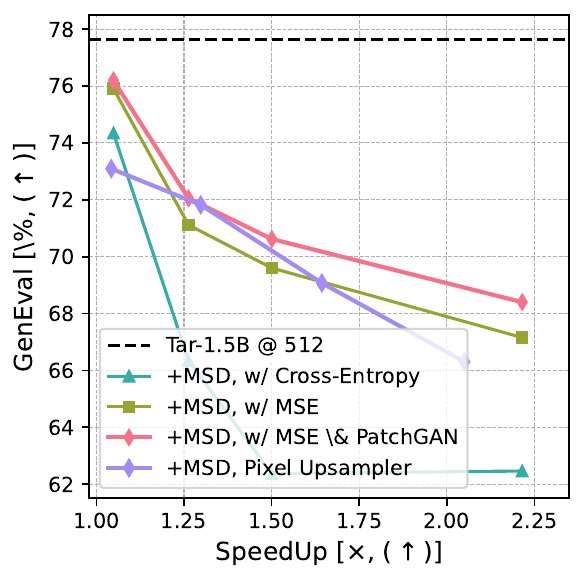}
    \caption{Up- Down- sampling Losses}
    \label{fig:ablation_up_downsamplers}
\end{subfigure}
    \begin{subfigure}[b]{0.24\textwidth}
        \centering
        \includegraphics[width=\textwidth]{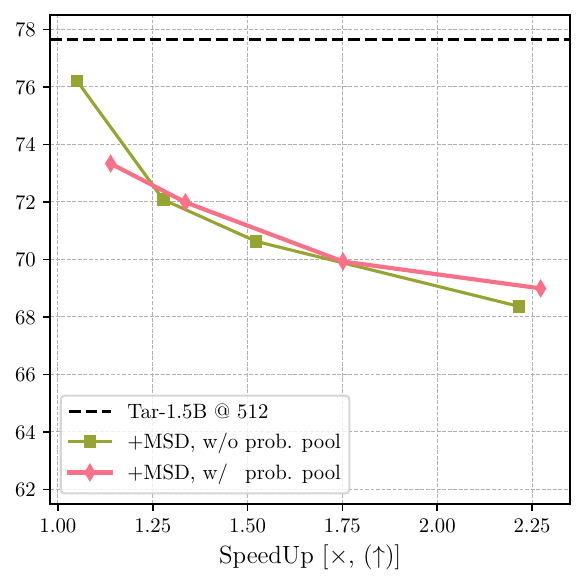}
        \caption{Probability Pooling}
        \label{fig:ablation_prob_pooling}
    \end{subfigure}
    \begin{subfigure}[b]{0.24\textwidth}
        \centering
        \includegraphics[width=\textwidth]{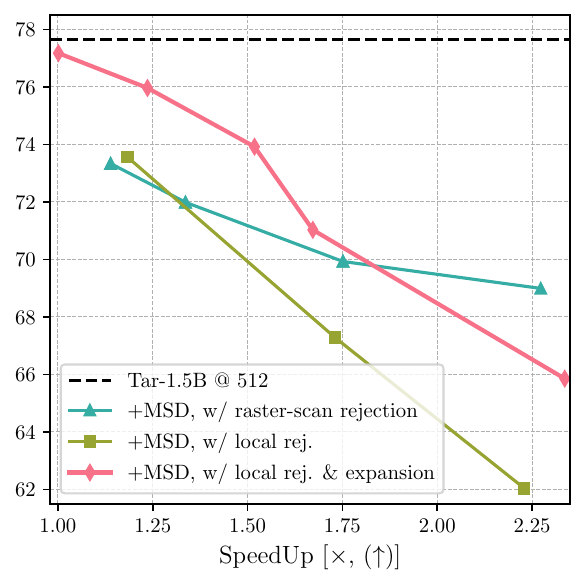}
        \caption{Local Verification and Expansion}
        \label{fig:ablation_locality_expansion}
    \end{subfigure}
    \begin{subfigure}[b]{0.24\textwidth}
        \centering
        \includegraphics[width=\textwidth]{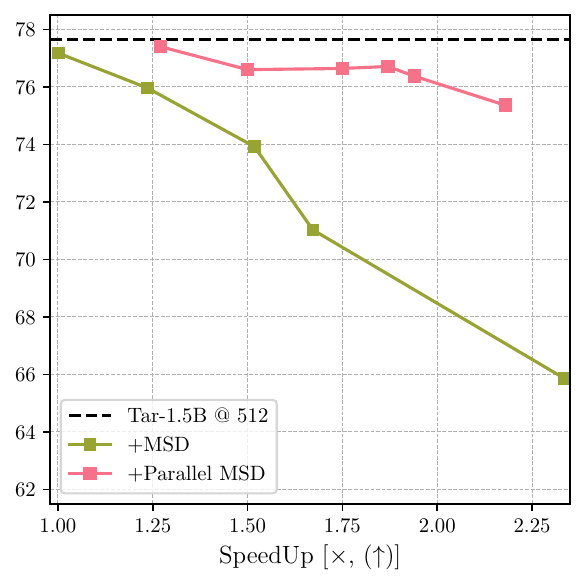}
        \caption{Parallel Decoding}
        \label{fig:ablation_parallel}
    \end{subfigure}

    \caption{We ablate different components of our method: (a) the contribution of loss functions in the up- down- samplers training, (b) the role of probability pooling during the verification process, and (c) comparison between standard rater-scan rejection and our proposed local rejection and expansion, and (d) comparison between sequential resampling and parallel decoding. MSD shortened version for multiscale speculative decoding.}
    \label{fig:ablations}
\end{figure*}


\paragraph{Upsampler Designs} are shown in \cref{fig:ablation_up_downsamplers}. Since Tar operates in the latent space of a discrete VQ-VAE, our initial approach employed a simple token-level classification loss (\colorbox{mygreen!20}{teal}). While this setup provided a functional starting point, the resulting images exhibited poor visual fidelity. 

Next, we removed latent-space supervision and instead applied reconstruction losses directly in pixel space rendering images with the VQ-Decoder. Specifically, we adopted a combination of MSE and LPIPS~\cite{zhang2018lpips} losses, which significantly improved perceptual quality (\colorbox{myolive!20}{green}). To further refine high-frequency details, we incorporate an adversarial component 
through a lightweight PatchGAN discriminator~\cite{isola2018patchgan} (\colorbox{myred!20}{pink}). 
Lastly, we evaluate an off-the-shelf, pixel-level upsampler~\cite{zhang2024hitsr} (\colorbox{mypurple!20}{purple}; see \cref{subsec:implementation}). Although this configuration underperforms the learned latent variant, it is training-free and integrates cleanly with our parallel-decoding resampling. We therefore adopt it as our default.

\paragraph{Probability Pooling} as introduced by LANTERN~\cite{jang2025lantern} is explored in \cref{fig:ablation_prob_pooling}. We compare two settings: considering only the drafted token probability  (\colorbox{myolive!20}{green}), compared to pooling the probability of the $k$ nearest neighbors in the VQ codebook space (\colorbox{myred!20}{pink}). Incorporating codebook-level proximity information improves acceptance rates and stabilizes performance, particularly beyond the $1.2\times$ speedup regime. However, the gains remain modest compared to the baseline without pooling. This is expected, as the pooling parameter behaves similarly to the acceptance threshold $\tau$, which serves as our primary relaxation mechanism.

\paragraph{Local Verification and Expansion} is shown in \Cref{fig:ablation_locality_expansion}. We compare three configurations: (i) standard raster-scan rejection from speculative decoding (\colorbox{mygreen!20}{teal}), which compromise image quality due to the low acceptance thresholds $\tau$ required to achieve high acceptance rates;  
(ii) naive local verification (\colorbox{myolive!20}{green}), which resamples only the rejected tokens without modifying their local context, resulting in even poorer performance;  
(iii) local verification with expansion (\colorbox{myred!20}{pink}), our proposed method, which resamples tokens within a radius $l$ around each rejected position.

Local verification leverages the strong spatial locality inherent in visual AR models, resulting in higher speedups for the same acceptance threshold $\tau$ (\colorbox{mygreen!20}{teal} vs \colorbox{myred!20}{pink}). At the same time, our proposed expansion mechanism plays a crucial role in enabling the verifier to correct not only the rejected tokens but also their surrounding context (\colorbox{myolive!20}{green} vs \colorbox{myred!20}{pink}). Additional ablations in the Supp. Mat..

\paragraph{Parallel Decoding}
We compare \methodname{} with and without parallel decoding enabled (\colorbox{myolive!20}{green}); see \cref{fig:ablation_parallel}.
Parallel decoding (\colorbox{myred!20}{pink})targets the two remaining sequential components: low-resolution drafting and resampling of rejected tokens at high resolution.
Empirically, Parallel Decoding yields consistent end-to-end latency gains at matched $\tau$ 
and with negligible changes in the metrics.

\subsection{Discussion and Limitations}
Our framework currently relies on two distinct checkpoints—a low‑resolution \emph{drafter} and a high‑resolution \emph{verifier}. While multi‑resolution checkpoints are increasingly common in modern AR systems, this assumption carries practical constraints. 
First, inference must load two sets of weights and maintain two KV‑caches, which can be prohibitive on memory‑constrained devices. 
Second, effectiveness depends on distributional alignment between the drafter and verifier: when they are trained independently or differ architecturally, acceptance rates drop and multi‑scale drafting yields smaller gains. We partially observe this effect on LlamaGen, where speedups are lower than on Tar.

A promising remedy is \emph{self‑speculative decoding}, in which a single checkpoint proposes drafts using internal layers and verifies with the full model. This removes duplicate weights/KV‑caches and improves alignment across resolutions. 

\section{Conclusion}
In this work we introduced \methodname{}, a multi-scale speculative decoding framework for accelerating autoregressive image generation. By combining low-resolution drafting with upsampling mechanism and a locality verification strategy, our method achieves substantial speedups while maintaining strong semantic alignment and perceptual quality. When paired with parallel decoding, it delivers the best speed–quality trade‑off among existing methods.

Through extensive experiments on Tar-1.5/7B across 512p and 1024p resolutions, we demonstrated that MuLo-SD consistently outperforms speculative decoding baselines such as EAGLE-2 and LANTERN, as well as parallel decoding methods like ZipAR. Ablation studies further validate the effectiveness of our multi-scale design, probability pooling, and local verification and expansion mechanisms.

MuLo-SD integrates seamlessly with next-token prediction objectives and unified MLLMs, making it a practical and scalable solution for high-resolution image synthesis. Future work includes extending our framework to video generation and other multimodal tasks.
\clearpage
{
    \small
    \bibliographystyle{ieeenat_fullname}
    \bibliography{main}
}

\clearpage
\setcounter{page}{1}
\maketitlesupplementary

\section{Primer on Tar}
We provide a concise description of the Tar architecture and the default parameters used to obtain the results reported in this paper. For additional details, we refer the reader to the original publication~\cite{han2025tar}.

\paragraph{Architecture}
For the purposes of this work, Tar consists of two main components:
\begin{enumerate}
    \item A Multimodal Large Language Model (MLLM) that processes the input prompt and generates a conditioning sequence,
    \item A generative detokenizer that maps the conditioning sequence to a VQ-VAE token sequence, which is then decoded to pixel space by the VQ-VAE decoder.
\end{enumerate}

The MLLM is fine-tuned from QwenVL~\cite{bai2025qwenvl} and extended to predict visual tokens. It is trained to output sequences of three different lengths: 81, 169, and 729 tokens. Each length corresponds to a progressively stronger conditioning signal for the detokenizer.

The autoregressive generative detokenizer (AR-DTok) is based on the LlamaGen~\cite{sun2024llamagen} model, fine-tuned to use the output of the MLLM as conditioning. Conditioning is implemented by pre-filling the sequence with one of the desired lengths (\eg, 81, 169, or 729). Importantly, the AR-DTok model operates in the latent space of a VQ-VAE~\cite{oord2018vqvae,esser2021vqgan}, which performs $16\times$ down-sampling along both spatial dimensions, resulting in sequence lengths of 256, 1024, and 4096 tokens for the resolutions 256p, 512p and 1024p respectively.

\paragraph{Sampling}
We now describe the sampling procedure. For the MLLM, we use the default configuration:
$\text{top}_k=1200$, $\text{top}_p=0.95$, the temperature $\tau_{\text{logits}}=1.0$ (different from the relaxed acceptance threshold $\tau$ defined in Equation 5 in the main paper) and set the sequence length to 729. The absolute latency for generating this conditioning sequence is approximately 17 seconds. Note that this value is not included in our latency analysis.
This conditioning sequence is then used to sample from the AR-DTok model. For AR-DTok, we set:
$\text{top}_k=0$, $\text{top}_p=1.0$ and the temperature $\tau_{\text{logits}}=1.0$ (\ie sampling from the full distribution of logits). Additionally, we apply classifier-free guidance with a scale of $4.0$, we use an empty sequence for the negative prompt.

Sampling from AR-Dtok takes on average 5s, 18s and 80s for each resolution respectively, see \Cref{tab:tar-latency} for an overview. Given that sampling the conditioning sequence takes an average of 17s, it reinforces that \methodname{}'s best setting is the 4\ttimes{} case i.e., going from 256p to 1024p. In this scenario, the total latency is largely dominated by the AR-DTok decoding time, and accelerating the visual token generation will lead to substantial speedups.


\begin{table}[t]
\centering
\caption{Summary of AR-Dtok configurations from Tar \cite{han2025tar}.}
\begin{tabular}{ccc}
\toprule
Resolution & Seq. Length & Latency \\ 
\midrule
256p & 256 & 5s \\
512p & 1024 & 18s \\
1024p & 4096 & 80s \\
\bottomrule
\end{tabular}
\label{tab:tar-latency}
\end{table}

\section{\methodname{}}

\begin{figure*}[t]
  \centering

  \begin{subfigure}[t]{0.48\textwidth}
    \centering
    \includegraphics[page=1,width=1.0\columnwidth,keepaspectratio]{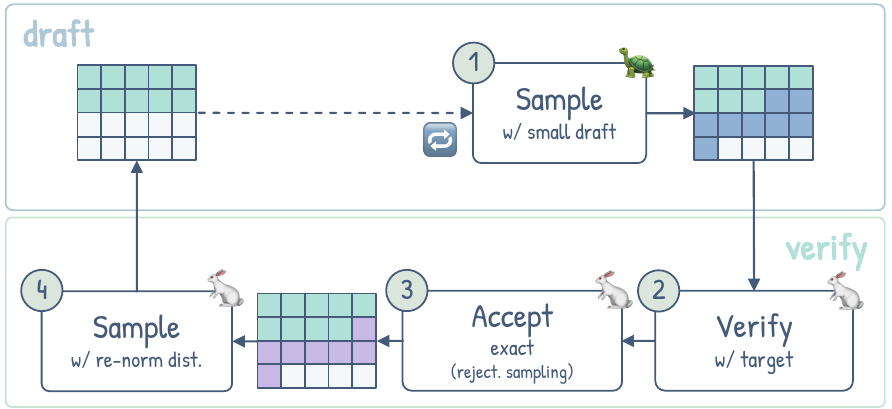}
    \caption{Speculative Decoding}
    \label{fig:sd}
  \end{subfigure}\hfill
  \begin{subfigure}[t]{0.48\textwidth}
    \centering
    \includegraphics[page=1,width=1.0\columnwidth,keepaspectratio]{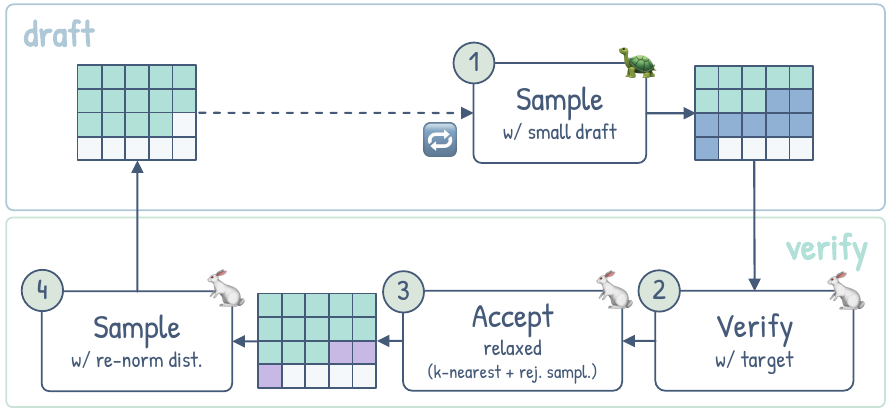}
    \caption{LANTERN}
    \label{fig:lantern}
  \end{subfigure}

  \caption{Overview of the standard speculative decoding~\cite{leviathan2023speculativedecoding} and LANTERN~\cite{jang2025lantern} methods. They are drawn in the same style as our main method figure for ease of comparison. \colorbox{proposed!30}{Blue} indicates draft tokens, \colorbox{accepted!30}{green} accepted  tokens, \colorbox{rejected!30}{purple} rejected tokens, \colorbox{empty!30}{blank} placeholder tokens. \emojiturtle indicates sequential operations, \emojirabbit parallel operations, \emojirepeat a drawing discontinuity due to looping.}
  \label{fig:sd-vs-lantern}
\end{figure*}

\paragraph{Method} We describe the algorithm of \methodname{} in \Cref{alg:mulo}, a full description can be found in Section 3.2 and Section 3.3 of the main paper. The step numbers \step{1} - \step{7} are a reference to the schematic representation in Figure 2 of the main paper. We present speculative decoding and LANTERN in the same style as our main method schema in~\Cref{fig:sd-vs-lantern}. For a detailed description of their algorithm, see Section 3.1 in the main paper.

\begin{algorithm*}[t]
\caption{Multi-Scale Local Speculative Decoding}
\label{alg:mulo}
\KwIn{The target model $M_p$ at scale $s_p$, the draft model $M_q$ at scale $s_q$, the up- and down-sampler $U_r$ and $D_r$ with a resampling factor of $r=s_p/s_q$, the initial sequence $x_0,\dots,x_t$, draft sequence length $L$, the target sequence length $T$, the cardinality of latent neighborhood $k$, the TVD threshold $\delta$, the probability mass threshold $\tau$ and $l$ the local neighborhood radius.}

\textbf{Initialize:} $n \leftarrow t$\;
\While{$n < T$}{
    \step{7} In parallel, down-sample the $n$ tokens to obtain prefix for draft model at scale $s_p$: $y_{1:n/r}=D_r(x_{1:n})$;
    
    \For{$t = 1,\dots,L/r$}{
        \step{1} In sequence, sample tokens from draft model $\tilde{y}_t \sim M_q(x \mid y_0,\dots,y_{n/r},\tilde{y}_1,\dots,\tilde{y}_{t-1})$\;
    }
    \step{2} In parallel, up-sample the $L/r$ tokens to obtain $L$ draft tokens at scale $s_q$: $\tilde{x}_{n:n+L}=U_r(\tilde{y}_{n/r:(n+L)/r})$\;
    \step{3} In parallel, compute $L$ sets of logits: $M_p(x \mid x_0,\dots,x_n), M_p(x \mid x_0,\dots,x_n,\tilde{x}_1), \dots, M_p(x \mid x_0,\dots,x_n,\tilde{x}_1,\dots,\tilde{x}_{L})$\;
    Initialize set of locally expanded rejected tokens $R_X \leftarrow \{\}$\;
    \For{$t = 1,\dots,L$}{
        Find the neighborhood $A_{k,\delta}(\tilde{x}_t)$\;
        \If{$\sum_{x \in A_{k,\delta}(\tilde{x}_t)} M_p(x \mid x_0,\dots,x_{n+t-1}) > \tau$}{
            \step{4} Accept: set $x_{n+t} \leftarrow \tilde{x}_t$;
        }
        \Else{
            \step{5} Reject: expand rejection to local neighborhood $N(t,l)$ around position $t$ with radius $l$, $R_X \leftarrow R_X \cup N(t,l)$
        }
    }
    Sort indices in $R_X$\;
    \For{$k \in R_X$}{
        \step{6} In sequence, sample rejected tokens from target model $x_{n+k} \sim M_p(x \mid x_0,\dots,x_{n+k-1})$;
    }
    Set $n \leftarrow n+L$
}
\KwOut{$x_{t+1},\dots,x_{T}$}
\end{algorithm*}

\paragraph{Implementation Details} The drafter model consists of three main components: an autoregressive model, an up-sampler, and optionally a down-sampler. The autoregressive model is set as AR-DTok @ 256p and remains fixed throughout all experiments. 

\paragraph{Learnable Up/Down-Samplers}
The up- and down- sampler are implemented as lightweight convolutional networks with residual blocks, and use pixel-shuffle to perform the correspondent resampling operation. We progressively train the up- and down- sampler for the 2\ttimes{} and the 4\ttimes{} settings.

In the $2\times$ setup, each module contains approximately 20M learnable parameters. These modules are trained on the LAION-COCO-Aesthetic \cite{li2024laioncocoaesthetic} dataset for 150k steps with a batch size of 32, using the AdamW optimizer with learning rate of $3\mathrm{e}\!-\!4$. We use a combination of losses for training: MSE, LPIPS, commitment loss, and discriminator loss. The overall objective is defined as:
\begin{equation}
\mathcal{L}_{\text{tot}} = \mathcal{L}_{\text{MSE}} + \mathcal{L}_{\text{LPIPS}} + \mathcal{L}_{\text{commit}} + \lambda_{\text{GAN}} \cdot \mathcal{L}_{\text{GAN}}.
\end{equation}
For the first 20k iterations, the up- and down- samplers are trained without the discriminator loss; this component is introduced afterward. The discriminator follows the standard PatchGAN design \cite{isola2018patchgan}, consisting of three convolutional layers, and is trained from scratch using AdamW with a learning rate of $5\mathrm{e}\!-\!4$ with $\lambda_{GAN} = 0.25$.

Next, we add an additional block of convolutions for the 4\ttimes{} case (resulting in approximately 30M parameters for each module). The up- and down- sampler are warm-started from the 2\ttimes{} checkpoints and trained for another 150k steps. We use the same configurations, except a smaller batch size of 8 to fit into memory.

\paragraph{Inference Hyperparameter}
During inference, \methodname{} introduces one primary hyperparameter: the acceptance threshold $\tau$ (see Equation 5). We perform a sweep over various values and ultimately fix $\tau = 1\mathrm{e}\!-\!4$, unless otherwise specified. Additionally, two other hyperparameters control the probability aggregation from neighboring elements (see Step 4 of Fig.~2 of the main paper). These are set to $k = 1000$ and $\delta = 0.1$, and remain constant across all experiments. As discussed in the main paper, $(k, \delta)$ and $\tau$ play a similar role in relaxing the acceptance criterion; therefore, we primarily experiment with $\tau$ while keeping the others fixed.

\paragraph{Latency Analysis} As discussed in the main paper, one of the key characteristics of \methodname{} is the computational cost associated with the drafter model, which shares the same architecture as the target model. This allows us to estimate the theoretical speedup under different acceptance rates by considering the reduction in the number of function evaluations (NFE) throughout the model. The theoretical speedup $S_T$ can be computed as follows. Using the notation from the main paper, let $M_p$ denote the target model and $M_q$ the drafter model, and define $T_p$ and $T_q$ as the sequence lengths for the target and drafter respectively, and let $a$ denote the acceptance rate. Then: 

\begin{figure*}[!t]
    \centering
    \includegraphics[width=\textwidth]{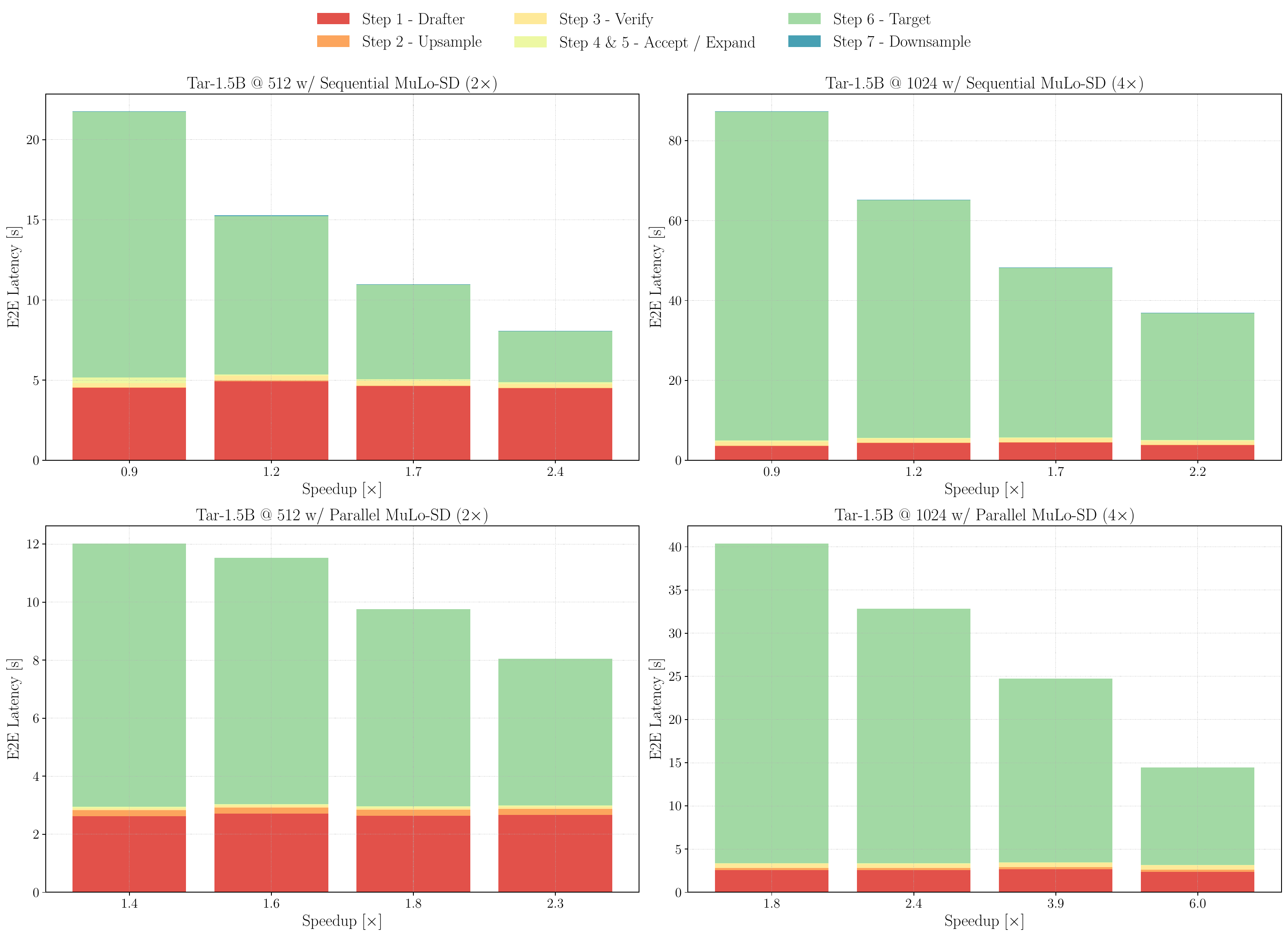}
    \caption{Breakdown of latency analysis. The figure illustrates the proportion of time spent in each step of our algorithm relative to the total latency. The step number in the legend refers to Fig.~2 in the main paper. The top row shows results \emph{without} applying parallel decoding to Steps 1 and 6; the bottom row incorporates it. Note the different x- and y-axis scales across rows.}
    \label{fig:latency_breakdown}
\end{figure*}

\begin{equation}
    S_{T} = \dfrac{T_p}{(1-a) \cdot T_p + T_q}.
\end{equation}

We compute the empirical speedup by measuring the time required to generate 500 prompts from MS-COCO 2017 Validation Set \cite{lin2015coco} on a single NVIDIA A100 GPU with a batch size of 1. We break down the individual cost of each component in \Cref{fig:latency_breakdown} (top row).
First, we observe that the cost of the drafter is fixed, regardless of the acceptance rate, since we always sample the same number of tokens from it. This eventually becomes the bottleneck in the 512p case, reducing the overall utility of our method. Conversely, at higher resolutions, the number of tokens generated by the target model is so large that the drafter’s cost becomes negligible. This further reinforces the suitability of the 4\ttimes{} setting (256p $\rightarrow$ 1024p) for our model.
As shown visually, almost all of the latency budget is spent sequentially sampling from either the target model or the drafter. This leads to two important considerations:
(i) our proposed multi-scale speculative decoding introduces only a negligible overhead—about 5\% and 3\% for the 512p and 1024p settings, respectively; and
(ii) there is still room for improvement by reducing the cost of the drafter and the verifier. 

By introducing parallel decoding, we substantially reduce end-to-end latency (\Cref{fig:latency_breakdown}, bottom row). As detailed in \Cref{subsec:parallel}, we perform drafter and target sampling using a parallel decoding technique. Although these stages still dominate the latency budget, their impact is markedly reduced, yielding a larger overall speedup. Note the different scales on the x- and y-axes when comparing the top and bottom panels of the breakdown.


\section{Experiments}
\paragraph{Quantitative Evaluation}

\begin{figure*}[!ht]
    \centering
    \includegraphics[width=\textwidth]{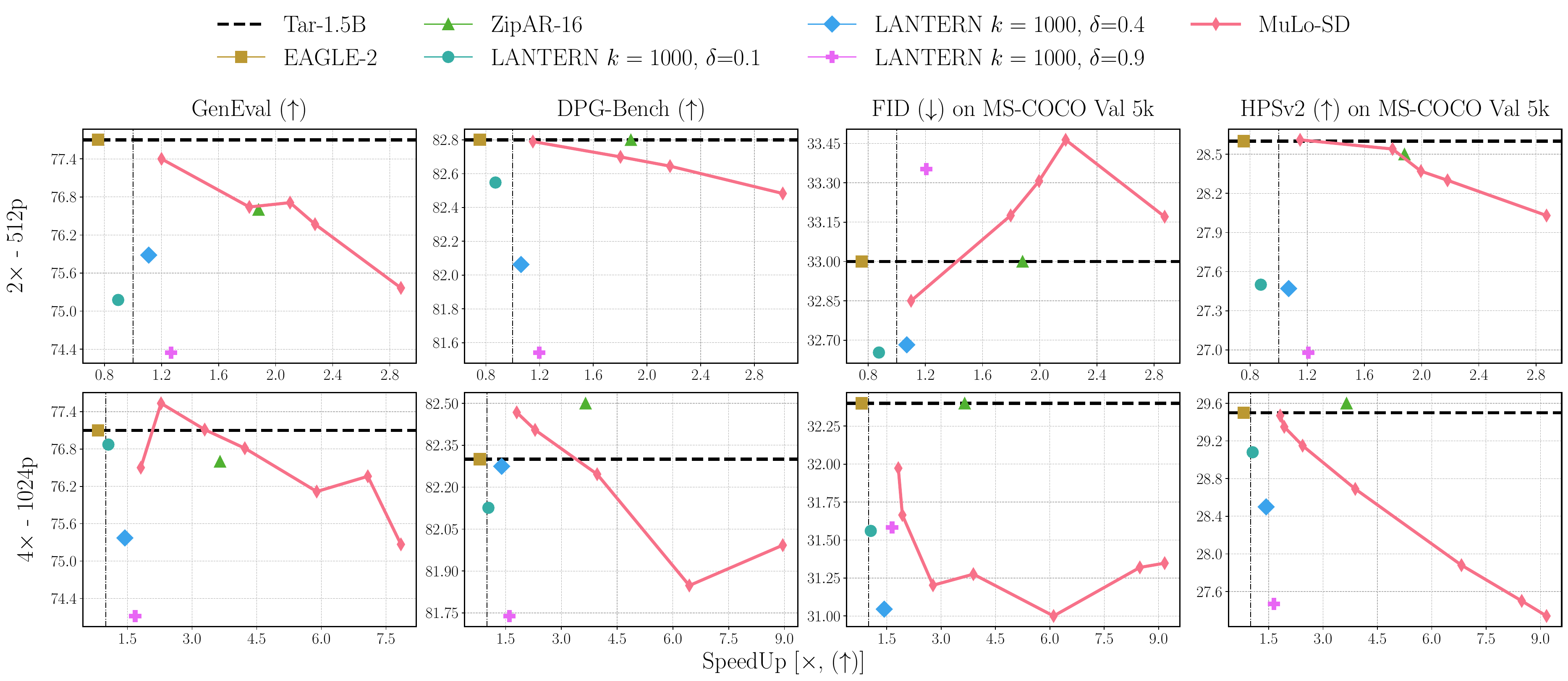}
    \caption{Quantitative evaluation of ZipAR~\cite{he2025zipar}, EAGLE-2~\cite{li2024eagle2}, LANTERN~\cite{jang2025lantern} and our method \methodname{}. We report the GenEval~\cite{ghosh2023geneval} and DPG-Bench~\cite{hu2024dpgbench} semantic alignment metrics, along with the FID~\cite{heusel2018fid} and HPSv2~\cite{wu2023hpsv2} perceptual quality metrics as computed on MS-COCO~\cite{lin2015coco} 2017 Val 5k. To obtain a curve for \methodname{}, we sweep the acceptance relaxation parameter $\tau$, as described in Section 3.3 and Equation (5) in the main paper.}
    \label{fig:pareto512}
\end{figure*}

We extend the quantitative results from the main paper by providing a graphical visualization of Table 1 in the main paper in ~\Cref{fig:pareto512}. It shows the pareto front of \methodname{} and contextualizes its performance with competing methods such as ZipAR~\cite{he2025zipar}, EAGLE-2~\cite{li2024eagle2} and LANTERN~\cite{jang2025lantern}. To create a pareto front, we vary the acceptance rate by sweeping different values for the relaxed acceptance threshold $\tau$ as defined in Equation 5 in the main paper. We can see that ZipAR dominates all other methods, with mostly unchanged perceptual quality compared to the reference, and only slight degradation to GenEval. Next comes our method \methodname{}, which across the semantic alignment metrics dominates EAGLE-2 and LANTERN. When it comes to perceptual quality metrics, FID tends to suffer for \methodname{} compared to other methods, and HPSv2 is sligthly better for \methodname{}.

\paragraph{Qualitative Evaluation} We supplement the qualitative results with additional visual comparisons. In~\Cref{fig:supp-qualitative-512} we show samples from Tar-1.5B 512p and \methodname{} for the 2\ttimes{} case (256p $\rightarrow$ 512p). In~\Cref{fig:supp-qualitative-1024-1} and~\Cref{fig:supp-qualitative-1024-2} we show additional results from Tar-1.5B 1024p and \methodname{} for the 4\ttimes{} case (256p $\rightarrow$ 1024p). In both the 512p and 1024p cases, the acceleration comes at a slight cost in perceptual quality, howevere the semantic alignment is mostly unaltered. Finally, in~\Cref{fig:qualitative-tau-sweep-1024} we showcase the effect of sweeping the relaxed acceptance threshold $\tau$ on the output image quality, where the different $\tau$ used correspond to the points in~\Cref{fig:pareto512}. The third column $\tau=1e-4$ corresponds the setting reported in Table 1 in the main paper. Note that it seems like the best tradeoff between speedup and perceptual quality, where the rightmost column ($\tau=1e-5$) shows the greatest speedup but largest degradation in quality, and the leftmost column ($\tau=1e-3$) is the closest to the original but ends up slower for more complex prompts.


Note that for all qualitative figures, both in the main text and the supplementary, we use prompts sourced from the DPG-Bench~\cite{hu2024dpgbench} benchmark dataset. We report the IDs of the prompts used in Fig.~4 of the main paper (order top-bottom, left-right) and refer to the official code for the actual text\footnote{\url{https://github.com/TencentQQGYLab/ELLA/tree/main/dpg_bench/prompts}.}:
\texttt{78.txt, midjourney32.txt, COCOval2014000000580698.txt, stanford34.txt, 5.txt, COCOval2014000000183648.txt, 62.txt, diffusiondb10.txt}.

\begin{figure}[!ht]
    \centering
    \includegraphics[width=0.5\textwidth]{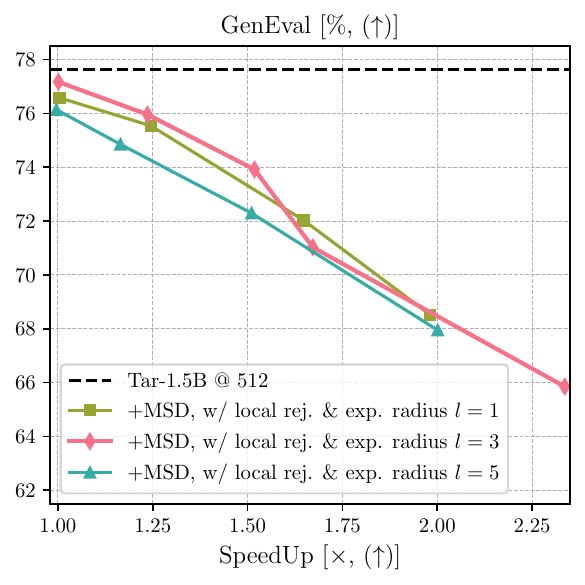}
    \caption{We study the effect of the local expansion radius $l$ in \methodname{} on GenEval~\cite{ghosh2023geneval} for the 2\ttimes case (256p $\rightarrow$ 512p). This expands the ablation in Figure 5 (c) in the main paper.}
    \label{fig:ablation_expansion_radious}
\end{figure}

\paragraph{Additional Ablation}
We extend the ablation presented in Figure 5 (c) in the main paper. We show the effect of the local expansion radius $l$ in \Cref{fig:ablation_expansion_radious}, showcasing $l=1$ and $l=5$ in addition to our default value of $l=3$ shown in the main paper. Similar to the other ablations in the main paper, the study is performed in the 2\ttimes{} case (256p $\rightarrow$ 512p). We can see that $l=3$ provides the best boost in GenEval performance across the 1 - 1.5\ttimes{} speedup range of interest. It is closely followed by $l=1$, with $l=5$ lagging behind. We expect the optimal value for $l$ to depend heavily on the resolution, as large resolution will benefit from larger radii, and conversely smaller resolution will suffer from larger radii as it will lead to high rejection rate even for permissive relaxed acceptance thresholds $\tau$. We anyway use $l=3$ for the 1024p case based on the result of this ablation due to lack of computational resource and time to ablate the parameter on the higher resolution. 

\paragraph{Discussion on LANTERN} As discussed in the main paper, porting LANTERN~\cite{jang2025lantern} (and EAGLE-2~\cite{li2024eagle2}) to Tar~\cite{han2025tar} proved significantly more challenging—yielding worse performance—than what was originally reported for LlamaGen~\cite{sun2024llamagen}. In this section, we detail our training procedure and provide additional justifications for the observed results.
We follow the original training script from the LANTERN codebase. The drafter consists of a single transformer layer and is trained using activations from the last transformer block of the target model (\ie before the final softmax fully connected layer).

The training objective combines two losses:
(i) standard cross-entropy loss for next-token prediction, and
(ii) an L1 loss to regress the hidden state of the teacher (\ie the target model).
The overall loss is weighted using the configuration reported in LANTERN, with $\lambda_{L1} = 0.1$ for the regression term. We train the drafter on a subset of the LAION-COCO-Aesthetic dataset~\cite{lin2015coco}, using 100k samples for training and reserving 1k samples for evaluation. Since LANTERN does not specify the dataset used to train the drafter, a one-to-one comparison is not possible. Nevertheless, in our setting, we measure Top-1 and Top-3 accuracy on the held-out test set as proxies for drafter quality. Higher accuracy correlates with greater inference-time speedups, as more tokens are accepted by the target model. We select the drafter achieving the highest test accuracy as our final model.
Our results are as follows:

\begin{itemize}
    \item \textbf{512p}: Top-1 = 0.12, Top-3 = 0.19
    \item \textbf{1024p}: Top-1 = 0.22, Top-3 = 0.33 
\end{itemize}

When compared to LlamaGen results reported in the LANTERN paper (see Fig. 2(b) in~\cite{jang2025lantern}), our Top-1 accuracy is substantially lower (0.12 vs. 0.38). We attribute this discrepancy to Tar being a much stronger model than LLamaGen, making it harder to approximate due to its closer alignment with the true data distribution. For instance, Tar achieves significantly higher scores on benchmarks such as GenEval, where LlamaGen reportedly~\cite{wang2024emu3} scores 32\% compared to 78\% for Tar. Furthermore, the original paper notes that the drafter performs worse on slightly stronger models like Anole~\cite{chern2024anole} compared to LlamaGen, reinforcing our hypothesis. Finally, we emphasize that the test sets differ, so direct comparisons are not strictly valid, although they provide context for interpreting our results.

\section{Acknowledgments}

We thank the authors of Tar~\cite{han2025tar}, LANTERN~\cite{jang2025lantern} and ZipAR~\cite{he2025zipar} for sharing their models and implementations.

\begin{table*}[!ht]
    \centering
    \def\arraystretch{1.0}
    \resizebox{\linewidth}{!}{
    \setlength\tabcolsep{0pt}
    \footnotesize
    \renewcommand{\arraystretch}{0.5}
    \begin{tabular}{cc@{\hskip 0.5mm}cc}
         {\tiny Tar-1.5B @ 512} & {\tiny \methodname{} (2\ttimes)} & {\tiny Tar-1.5B @ 512} & {\tiny \methodname{} (2\ttimes)} \\
         \includegraphics[trim=2 0 0 2,clip,width=0.3\columnwidth]{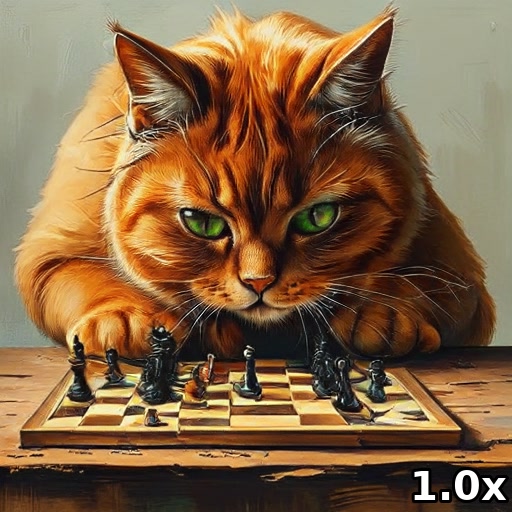} & 
         \includegraphics[trim=2 0 0 2,clip,width=0.3\columnwidth]{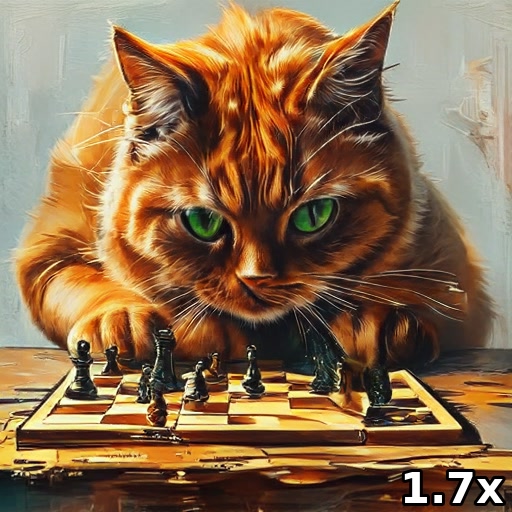} & 
         \includegraphics[trim=2 0 0 2,clip,width=0.3\columnwidth]{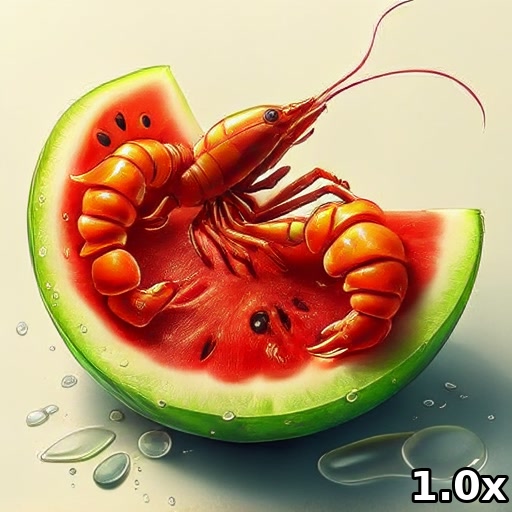} & 
         \includegraphics[trim=2 0 0 2,clip,width=0.3\columnwidth]{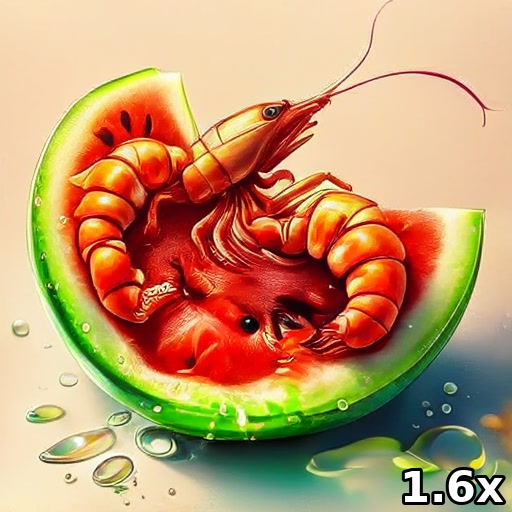} \\

         \includegraphics[trim=2 0 0 2,clip,width=0.3\columnwidth]{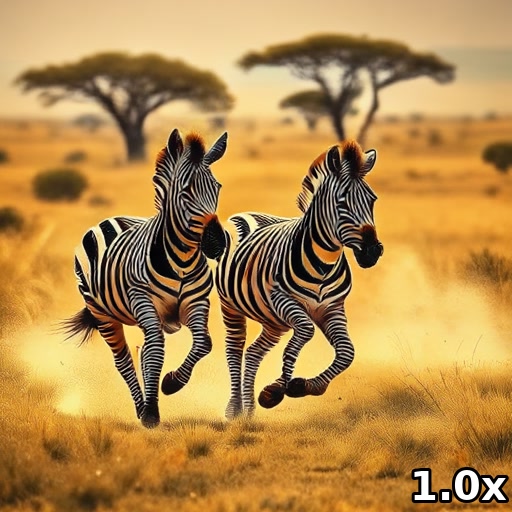} & 
         \includegraphics[trim=2 0 0 2,clip,width=0.3\columnwidth]{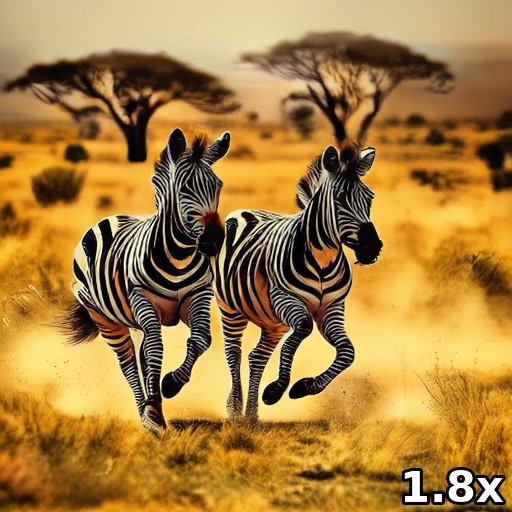} & 
         \includegraphics[trim=2 0 0 2,clip,width=0.3\columnwidth]{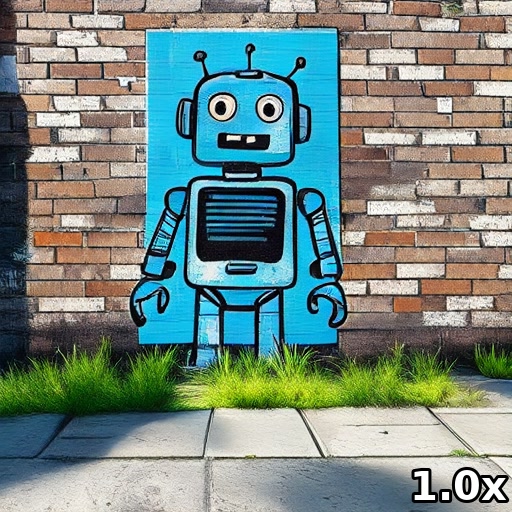} & 
         \includegraphics[trim=2 0 0 2,clip,width=0.3\columnwidth]{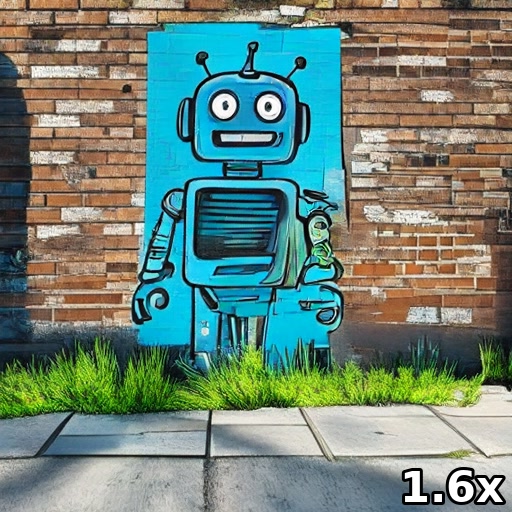} \\

         \includegraphics[trim=2 0 0 2,clip,width=0.3\columnwidth]{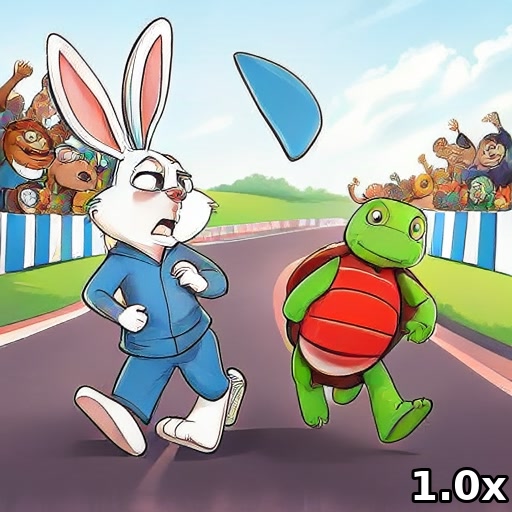} & 
         \includegraphics[trim=2 0 0 2,clip,width=0.3\columnwidth]{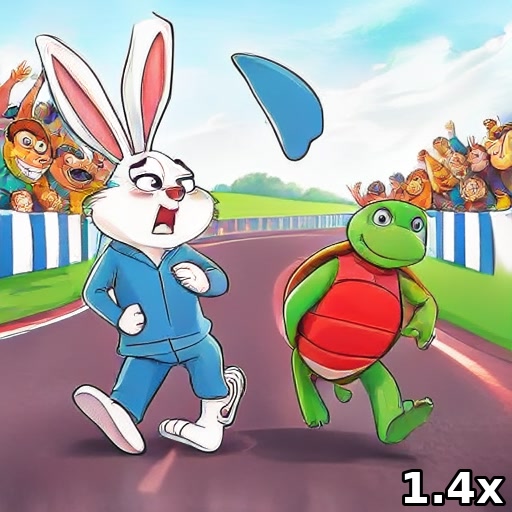} & 
         \includegraphics[trim=2 0 0 2,clip,width=0.3\columnwidth]{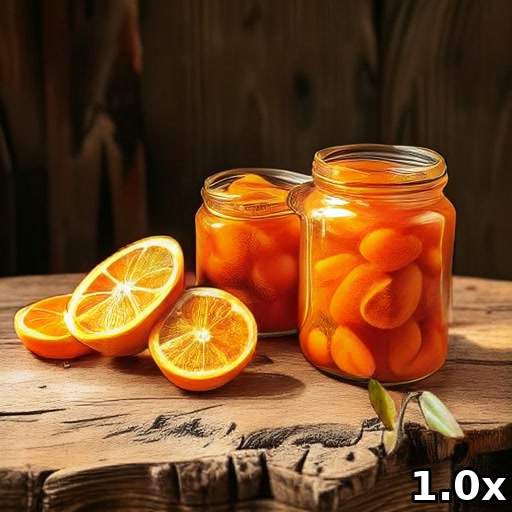} & 
         \includegraphics[trim=2 0 0 2,clip,width=0.3\columnwidth]{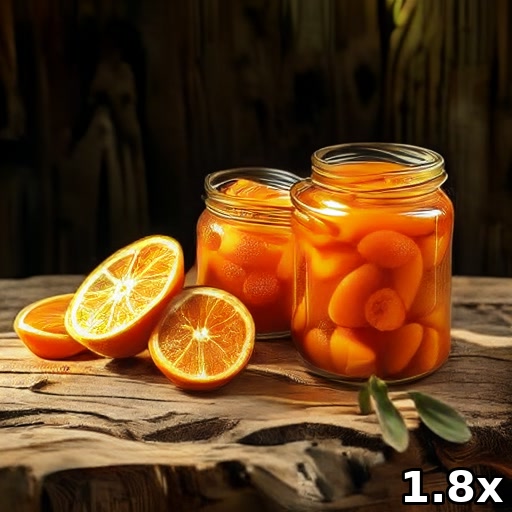} \\

        \includegraphics[trim=2 0 0 2,clip,width=0.3\columnwidth]{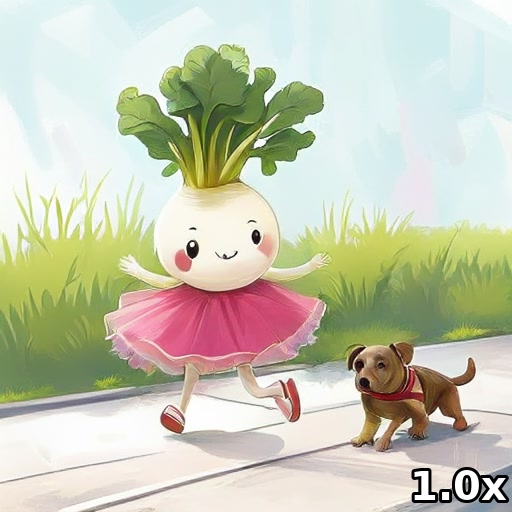} & 
         \includegraphics[trim=2 0 0 2,clip,width=0.3\columnwidth]{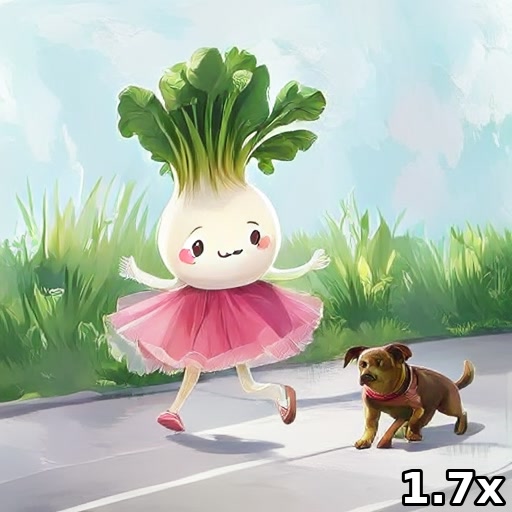} & 
         \includegraphics[trim=2 0 0 2,clip,width=0.3\columnwidth]{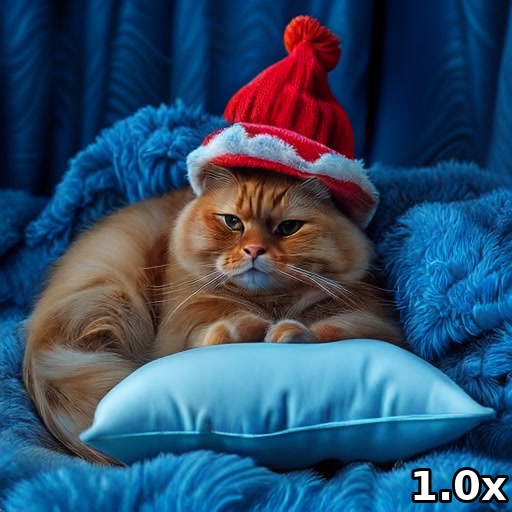} & 
         \includegraphics[trim=2 0 0 2,clip,width=0.3\columnwidth]{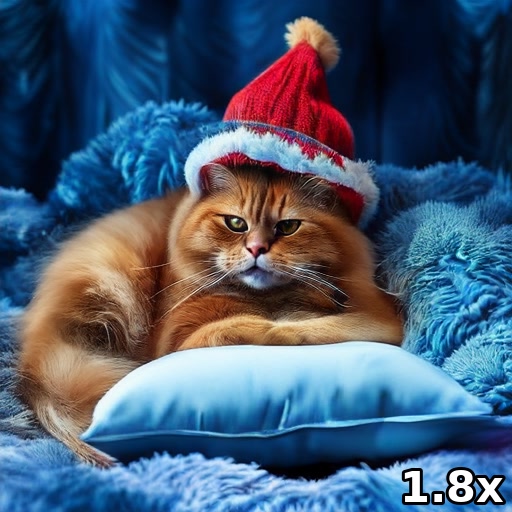} \\

        \includegraphics[trim=2 0 0 2,clip,width=0.3\columnwidth]{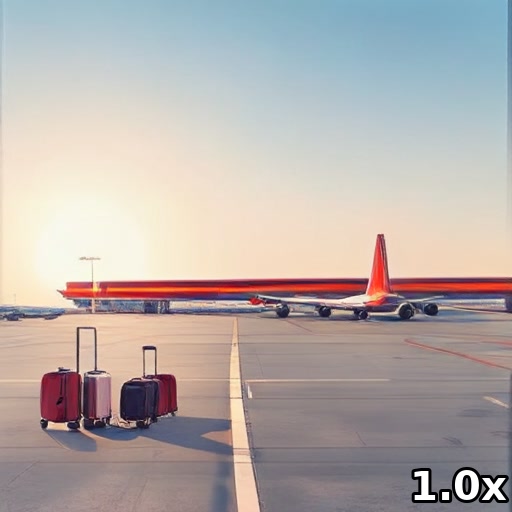} & 
         \includegraphics[trim=2 0 0 2,clip,width=0.3\columnwidth]{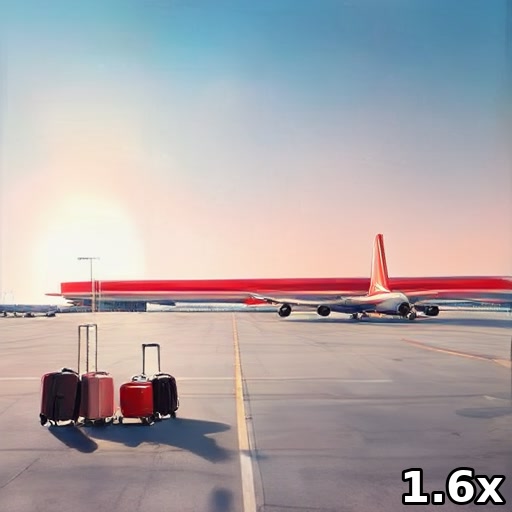} & 
         \includegraphics[trim=2 0 0 2,clip,width=0.3\columnwidth]{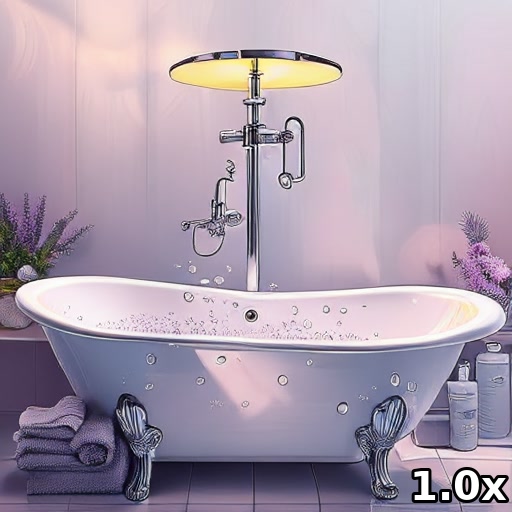} & 
         \includegraphics[trim=2 0 0 2,clip,width=0.3\columnwidth]{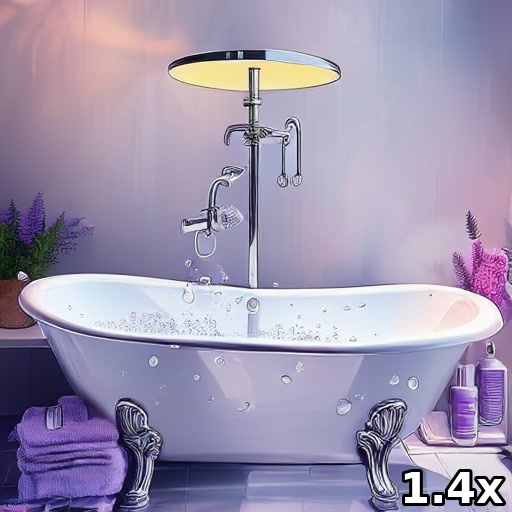} \\

    \end{tabular}
    }
    \captionof{figure}{Visual comparison of 512p resolution, speedup displayed at the bottom-left corner. Prompts from DPG-Bench (top-bottom, left-right): \texttt{partiprompts175.txt, 55.txt,  partiprompts124.txt, partiprompts303.txt, stanford6.txt, 180.txt, partiprompts177.txt, COCOval2014000000231527.txt, stanford36.txt, 189.txt}}
    \label{fig:supp-qualitative-512}
\end{table*}

\begin{table*}[!ht]
    \centering
    \def\arraystretch{1.0}
    \resizebox{\linewidth}{!}{
    \setlength\tabcolsep{0pt}
    \footnotesize
    \renewcommand{\arraystretch}{0.5}
    \begin{tabular}{cc}
         {\tiny Tar-1.5B @ 1024} & {\tiny \methodname{} (4\ttimes)} \\
         \includegraphics[trim=2 0 0 2,clip,width=0.5\columnwidth]{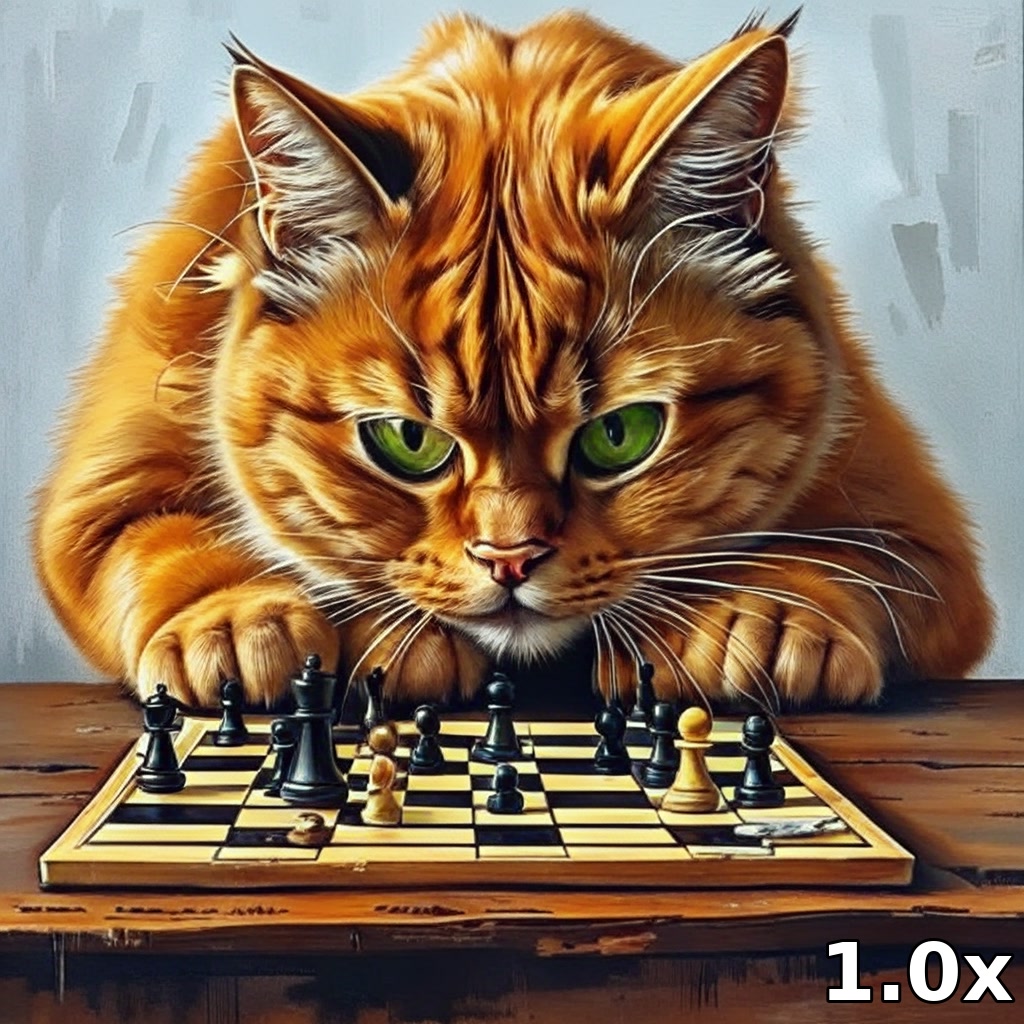} & 
         \includegraphics[trim=2 0 0 2,clip,width=0.5\columnwidth]{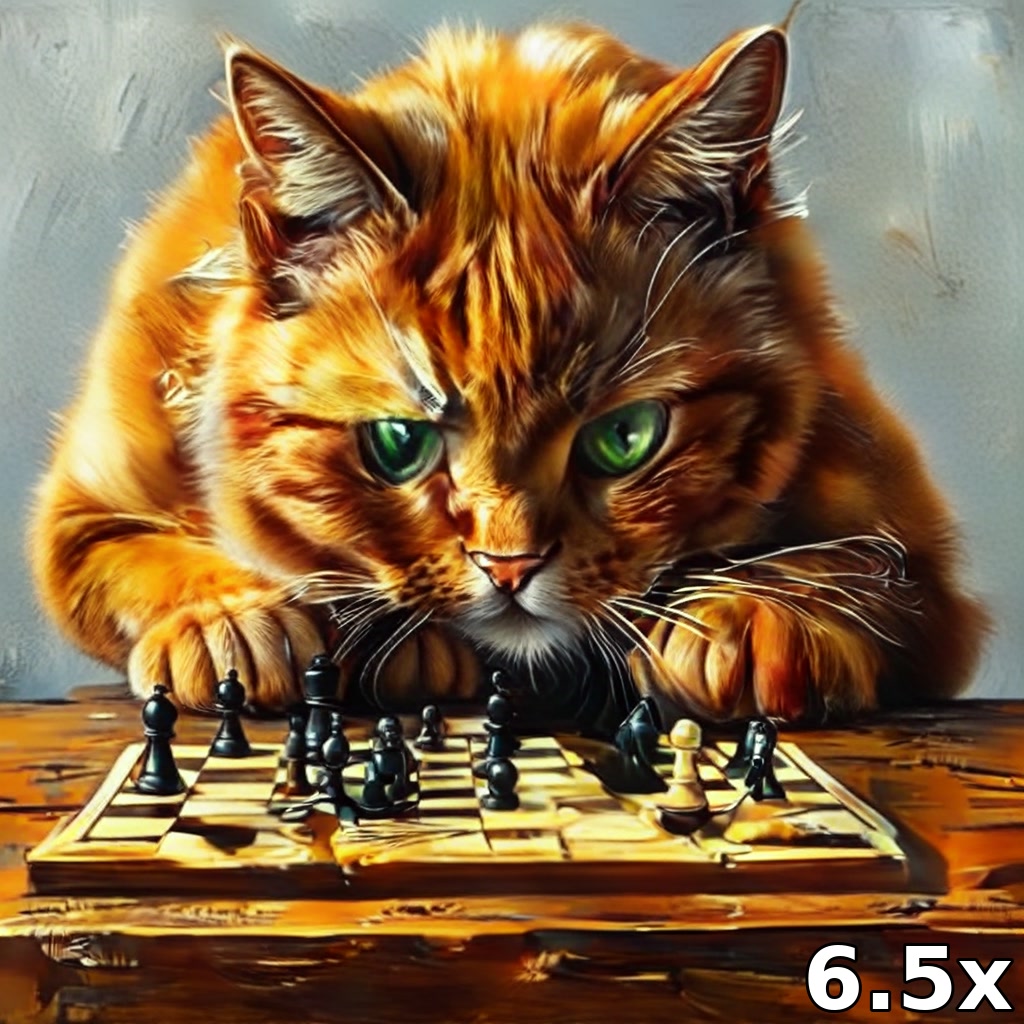} \\

         \includegraphics[trim=2 0 0 2,clip,width=0.5\columnwidth]{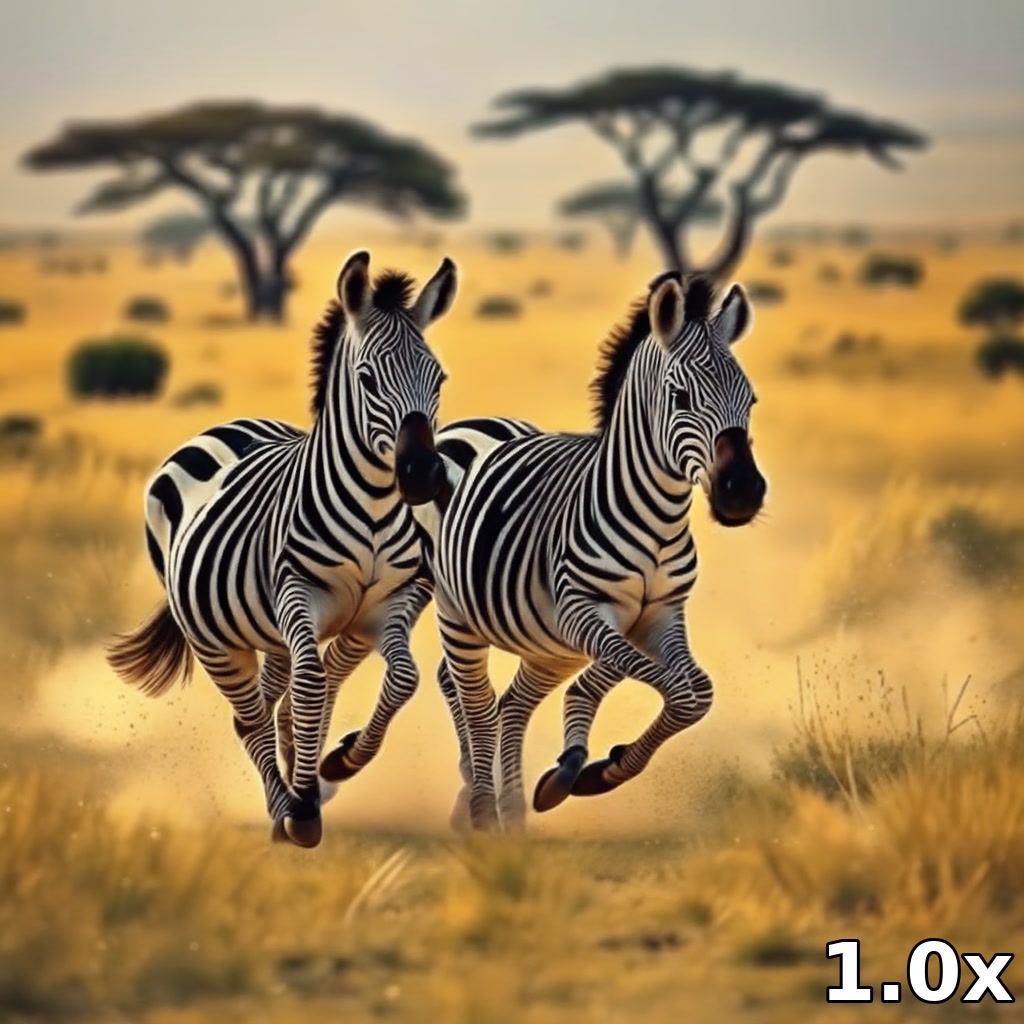} & 
         \includegraphics[trim=2 0 0 2,clip,width=0.5\columnwidth]{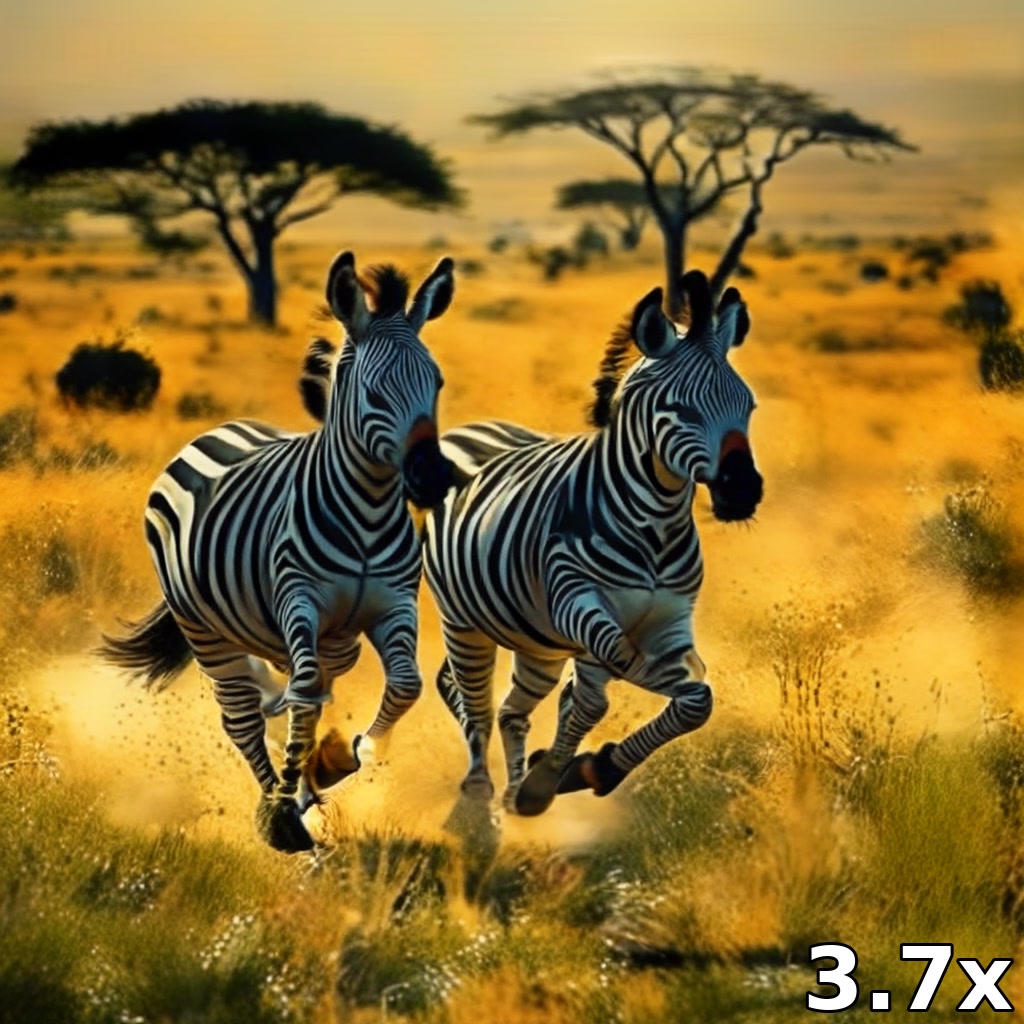} \\

    \end{tabular}
    }
    \captionof{figure}{Visual comparison of 1024p image generations. Prompts from DPG-Bench: \texttt{partiprompts175.txt, 55.txt}.}
    \label{fig:supp-qualitative-1024-1}
\end{table*}
\begin{table*}[!ht]
    \centering
    \def\arraystretch{1.0}
    \resizebox{\linewidth}{!}{
    \setlength\tabcolsep{0pt}
    \footnotesize
    \renewcommand{\arraystretch}{0.5}
    \begin{tabular}{cc}
         {\tiny Tar-1.5B @ 1024} & {\tiny \methodname{} (4\ttimes)} \\
         \includegraphics[trim=2 0 0 2,clip,width=0.5\columnwidth]{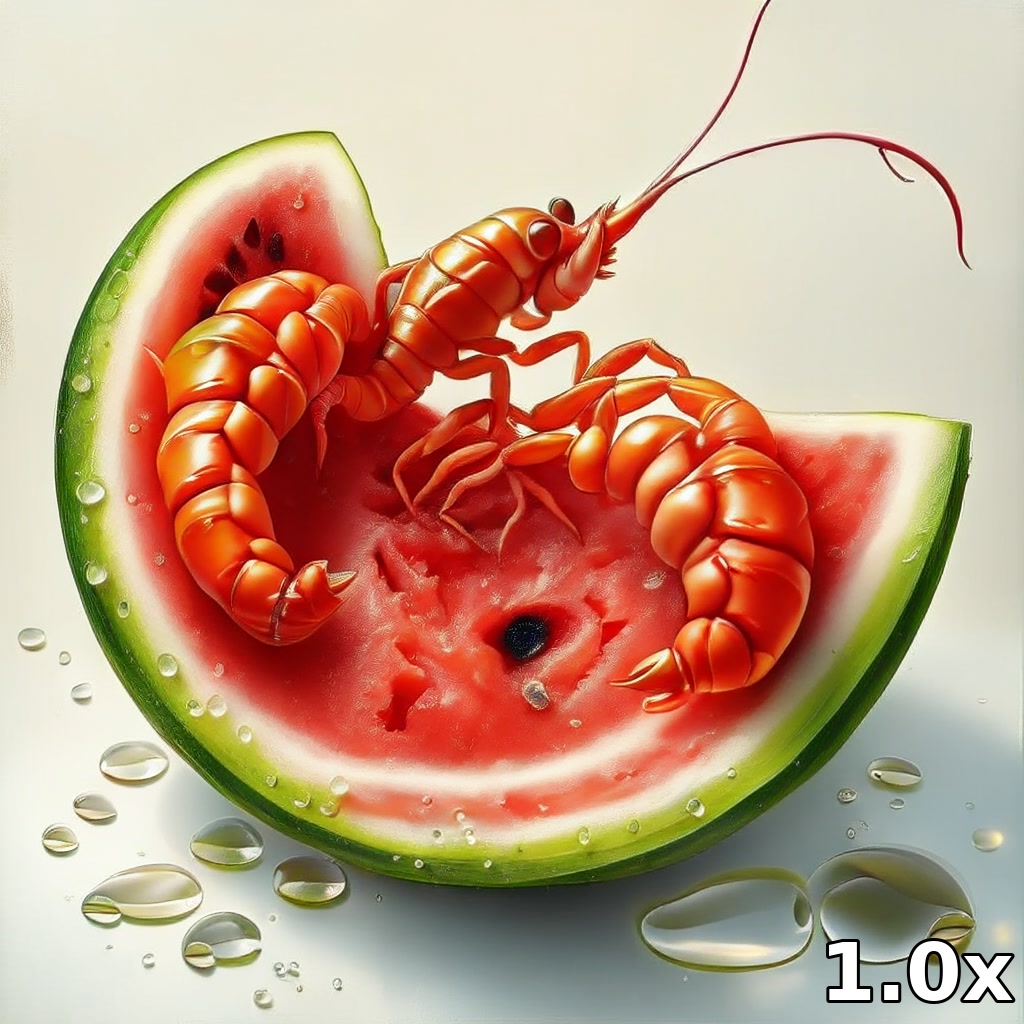} & 
         \includegraphics[trim=2 0 0 2,clip,width=0.5\columnwidth]{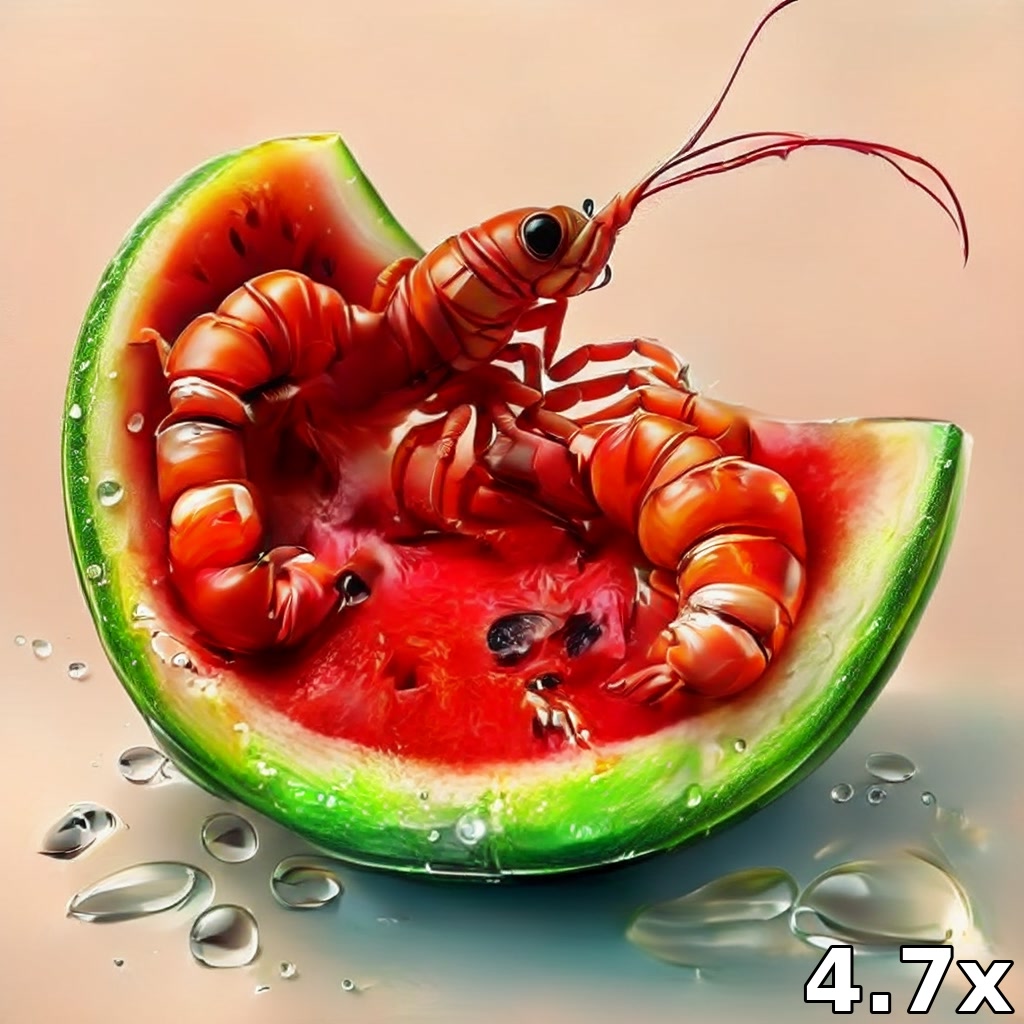} \\

         \includegraphics[trim=2 0 0 2,clip,width=0.5\columnwidth]{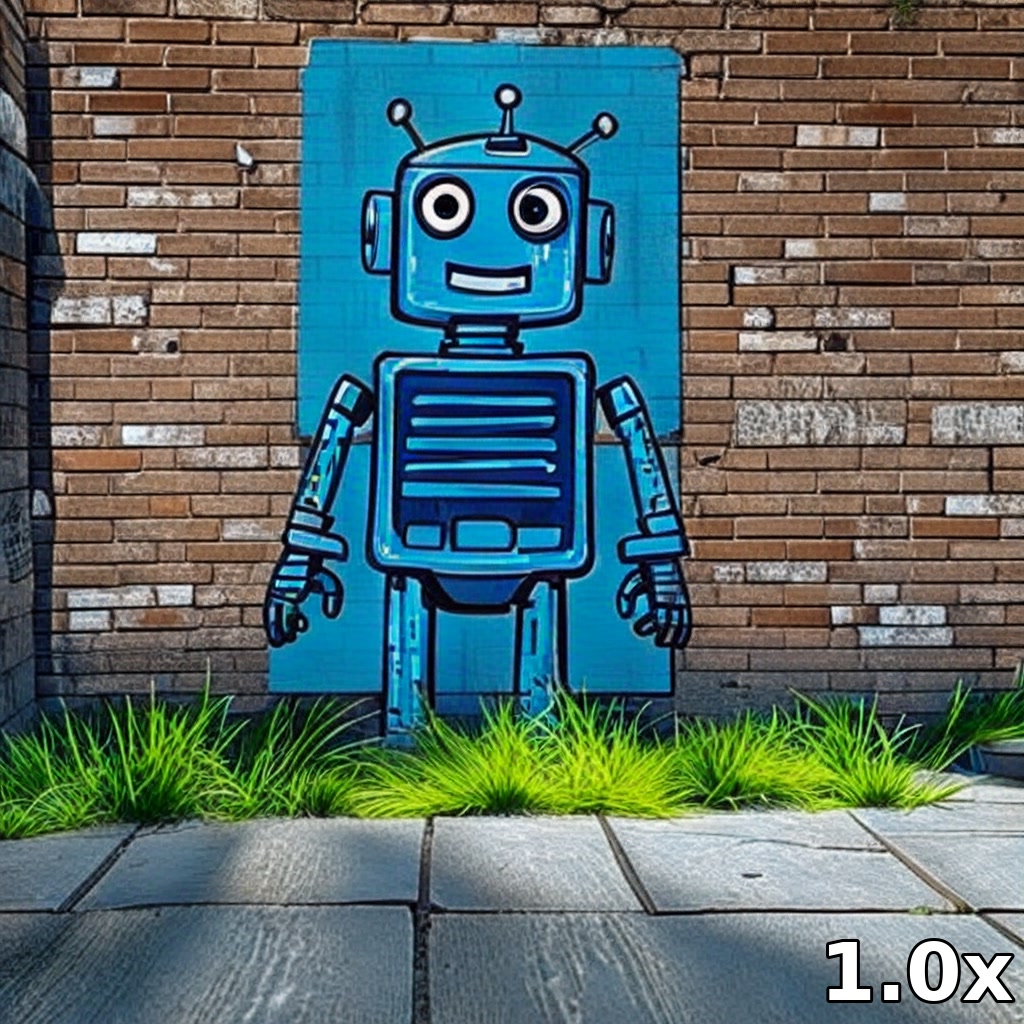} & 
         \includegraphics[trim=2 0 0 2,clip,width=0.5\columnwidth]{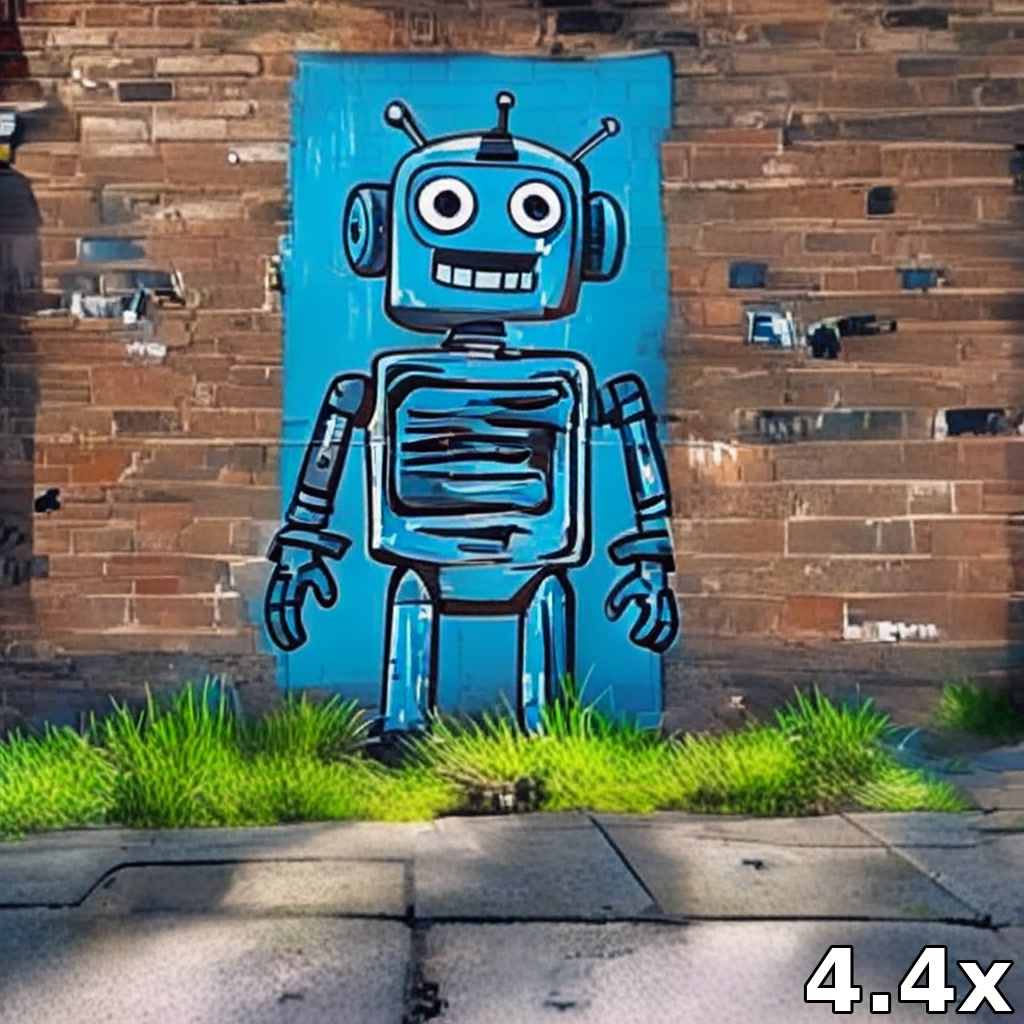} \\
 
    \end{tabular}
    }
    \captionof{figure}{Visual comparison of 1024p image generations. Prompts from DPG-Bench: \texttt{180.txt, partiprompts177.txt}.}
    \label{fig:supp-qualitative-1024-2}
\end{table*}
\begin{table*}[!ht]
    \centering
    \def\arraystretch{1.0}
    \resizebox{\linewidth}{!}{
    \setlength\tabcolsep{0pt}
    \setlength{\fboxrule}{0.75pt} 
    \footnotesize
    \renewcommand{\arraystretch}{0.5}
    \begin{tabular}{c@{\hskip 0.5mm}cccc}
         \tiny Tar-1.5B@1024 & \tiny $\tau\!=\!1\mathrm{e}\!-\!3$ & \tiny $\tau\!=\!1\mathrm{e}\!-\!4$ & \tiny $\tau\!=\!5\mathrm{e}\!-\!5$ & \tiny $\tau\!=\!1\mathrm{e}\!-\!5$ \\
         \includegraphics[trim=2 0 0 2,clip,width=0.2\columnwidth]{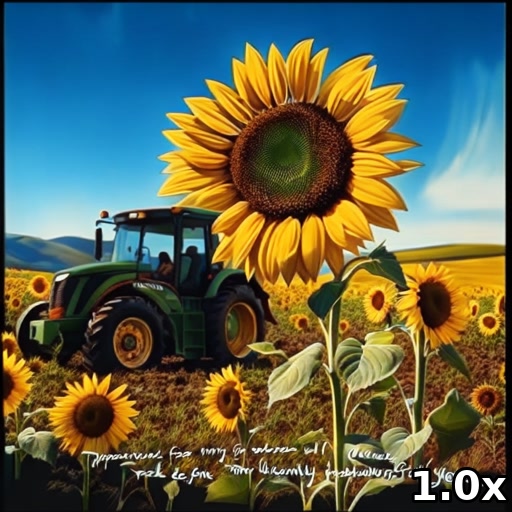} & 
         \includegraphics[trim=2 0 0 2,clip,width=0.2\columnwidth]{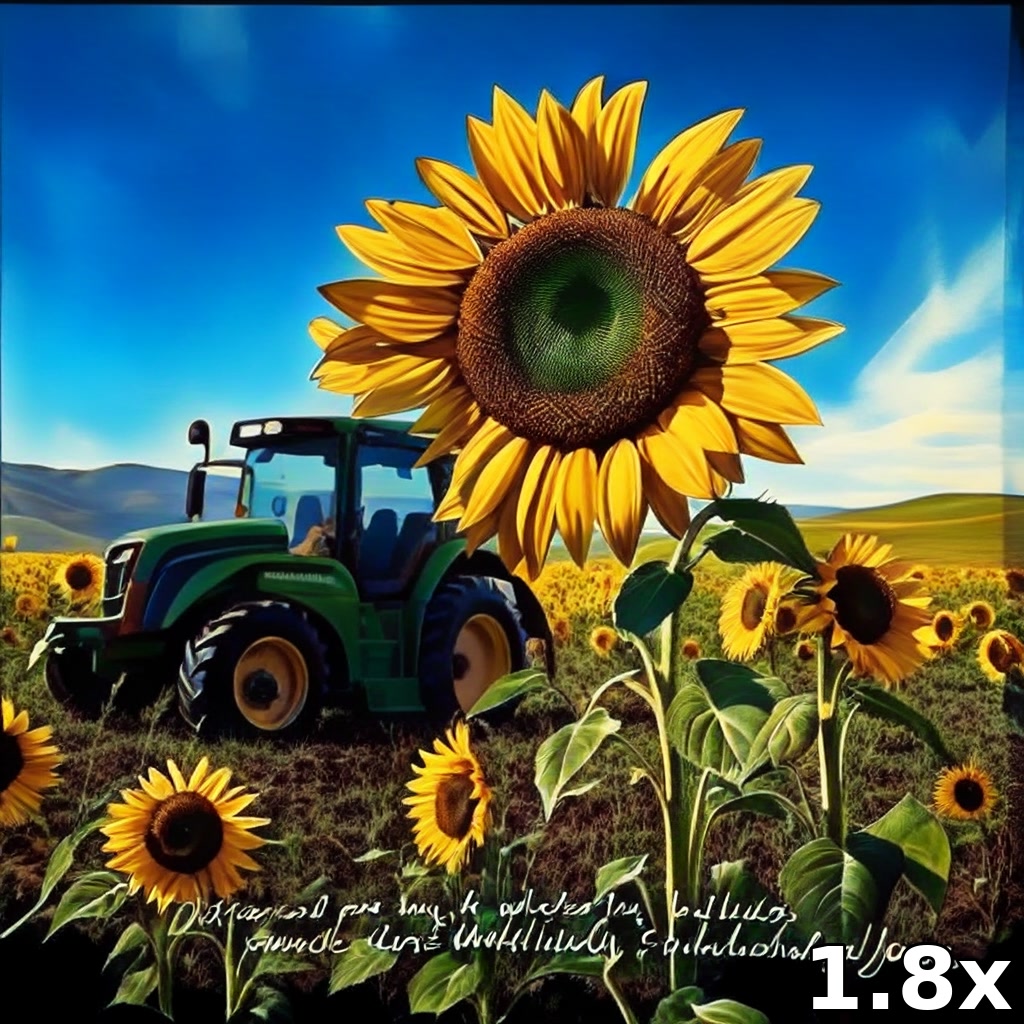} & 
         \includegraphics[trim=2 0 0 2,clip,width=0.2\columnwidth]{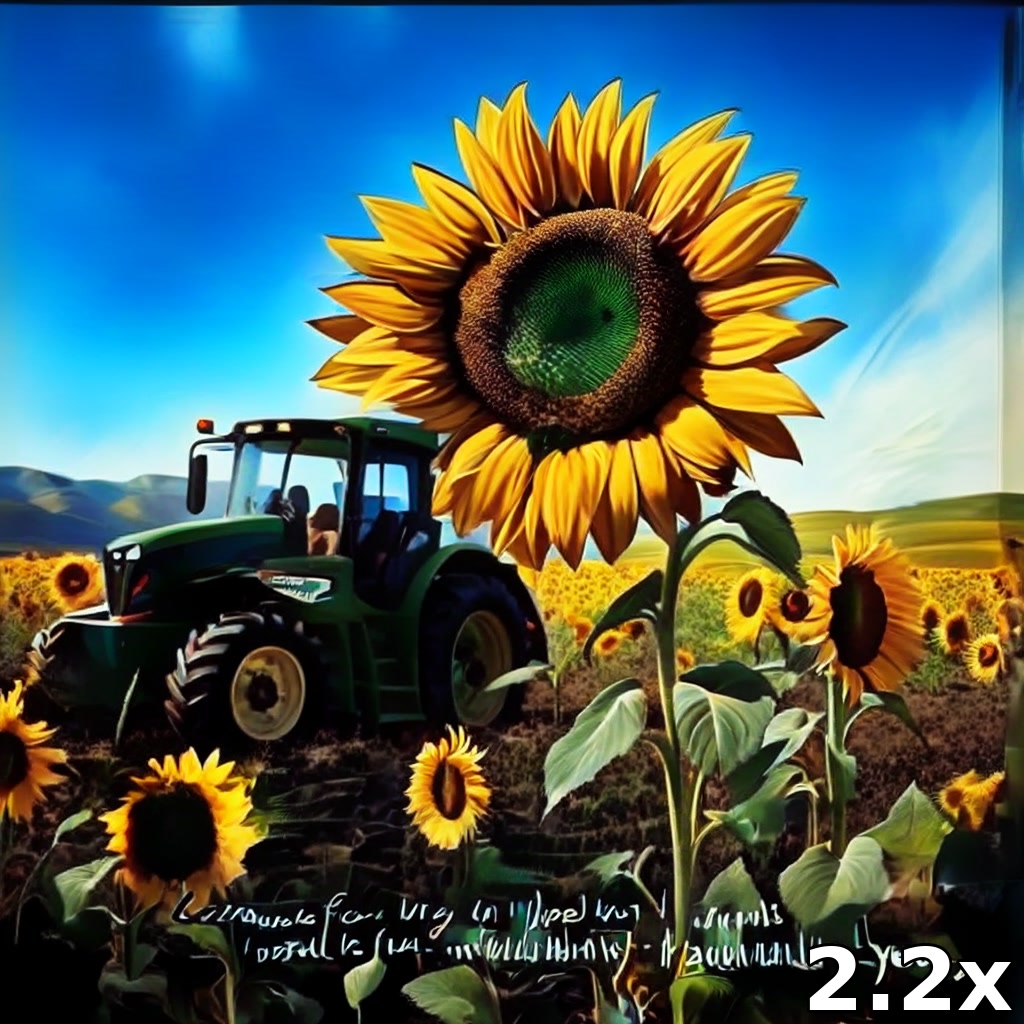} & 
         \includegraphics[trim=2 0 0 2,clip,width=0.2\columnwidth]{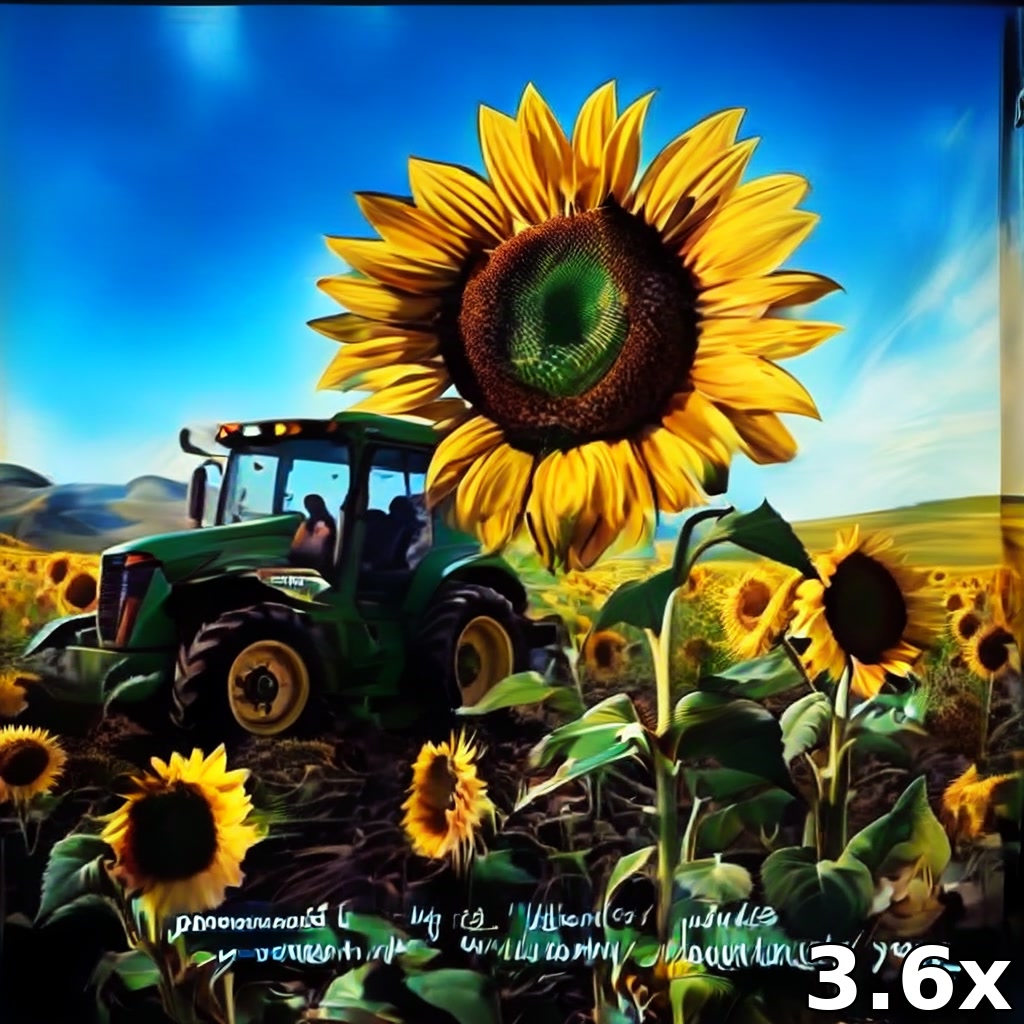} & 
         \includegraphics[trim=2 0 0 2,clip,width=0.2\columnwidth]{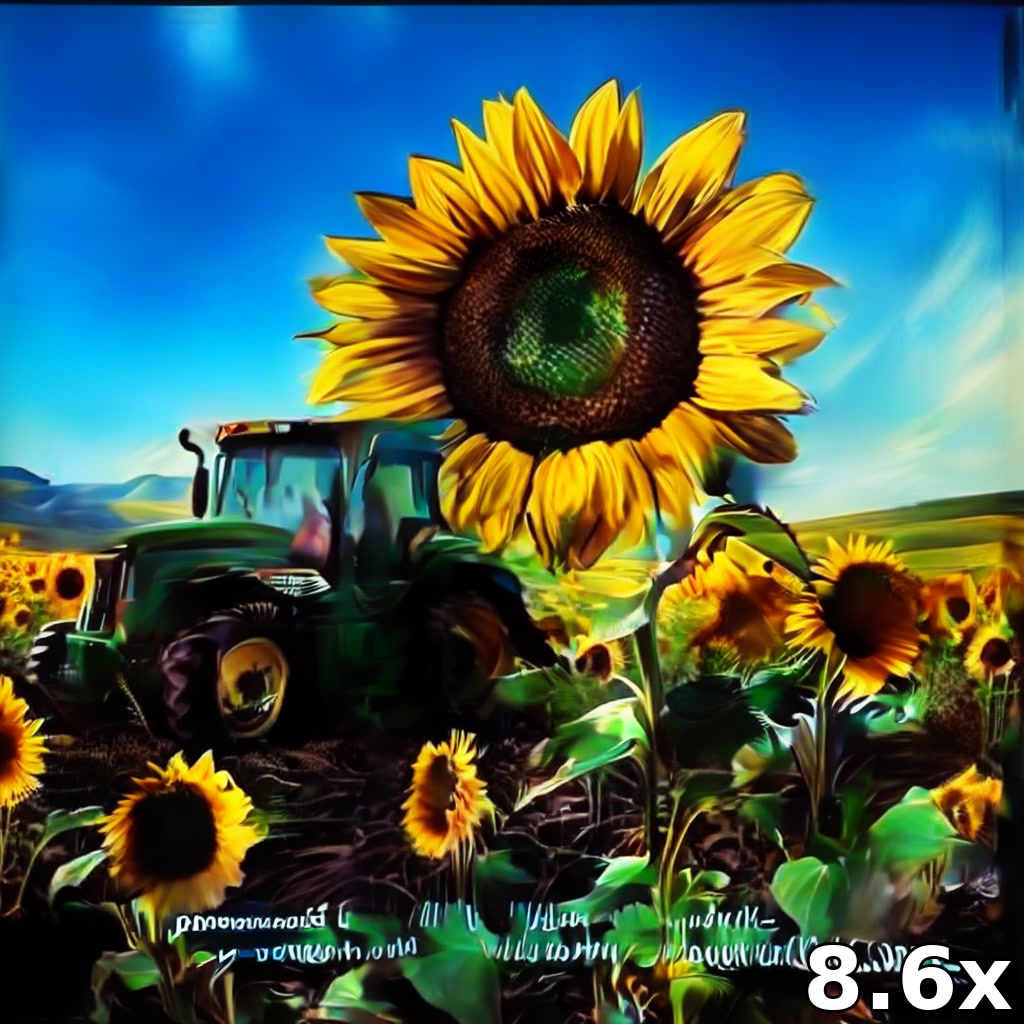}  \\

        \includegraphics[trim=2 0 0 2,clip,width=0.2\columnwidth]{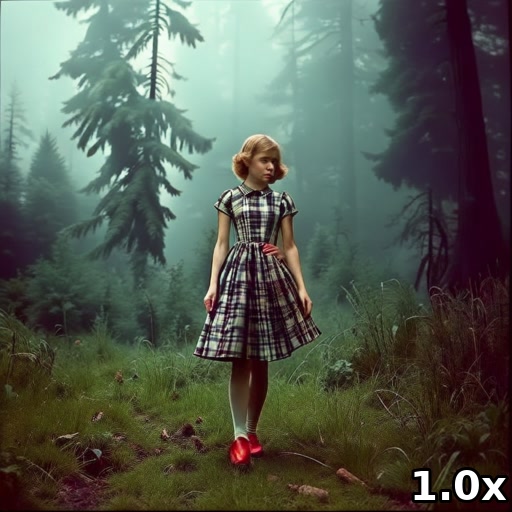} & 
        \includegraphics[trim=2 0 0 2,clip,width=0.2\columnwidth]{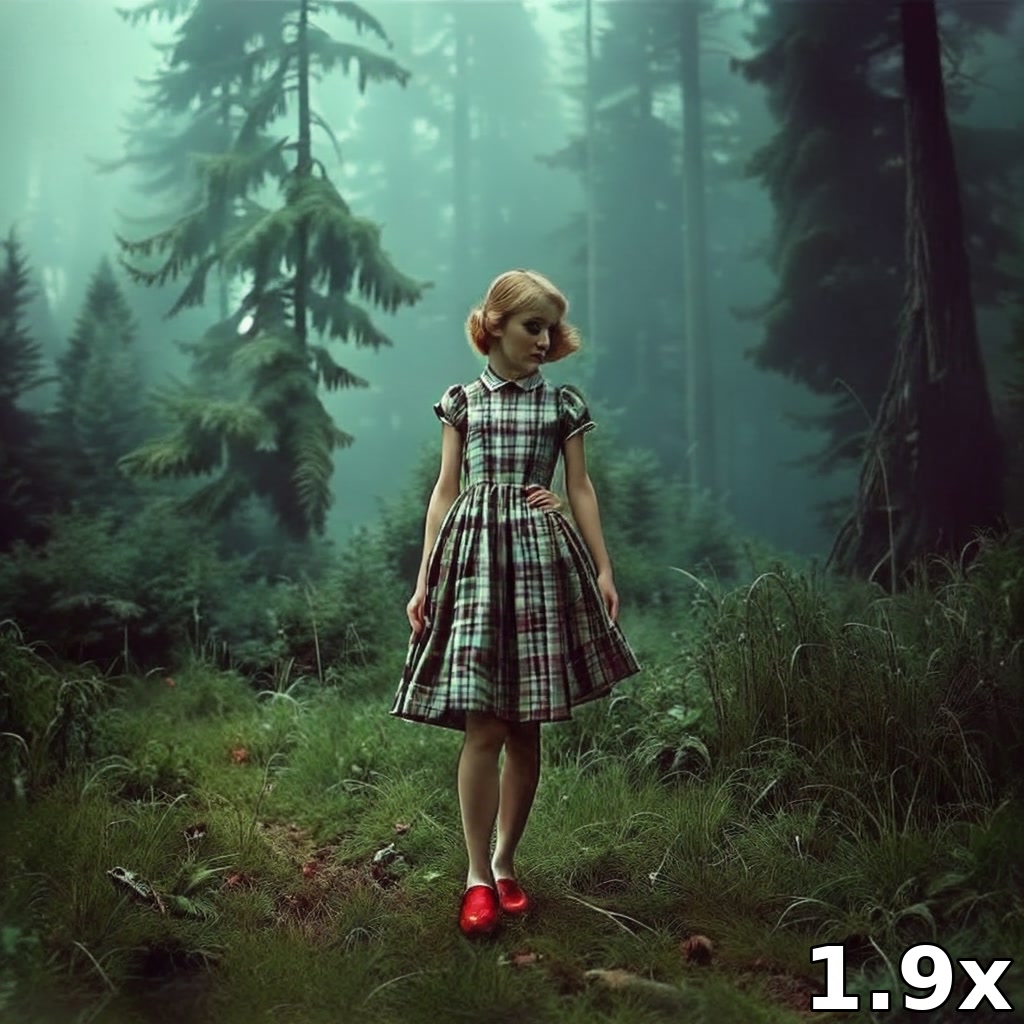} & 
        \includegraphics[trim=2 0 0 2,clip,width=0.2\columnwidth]{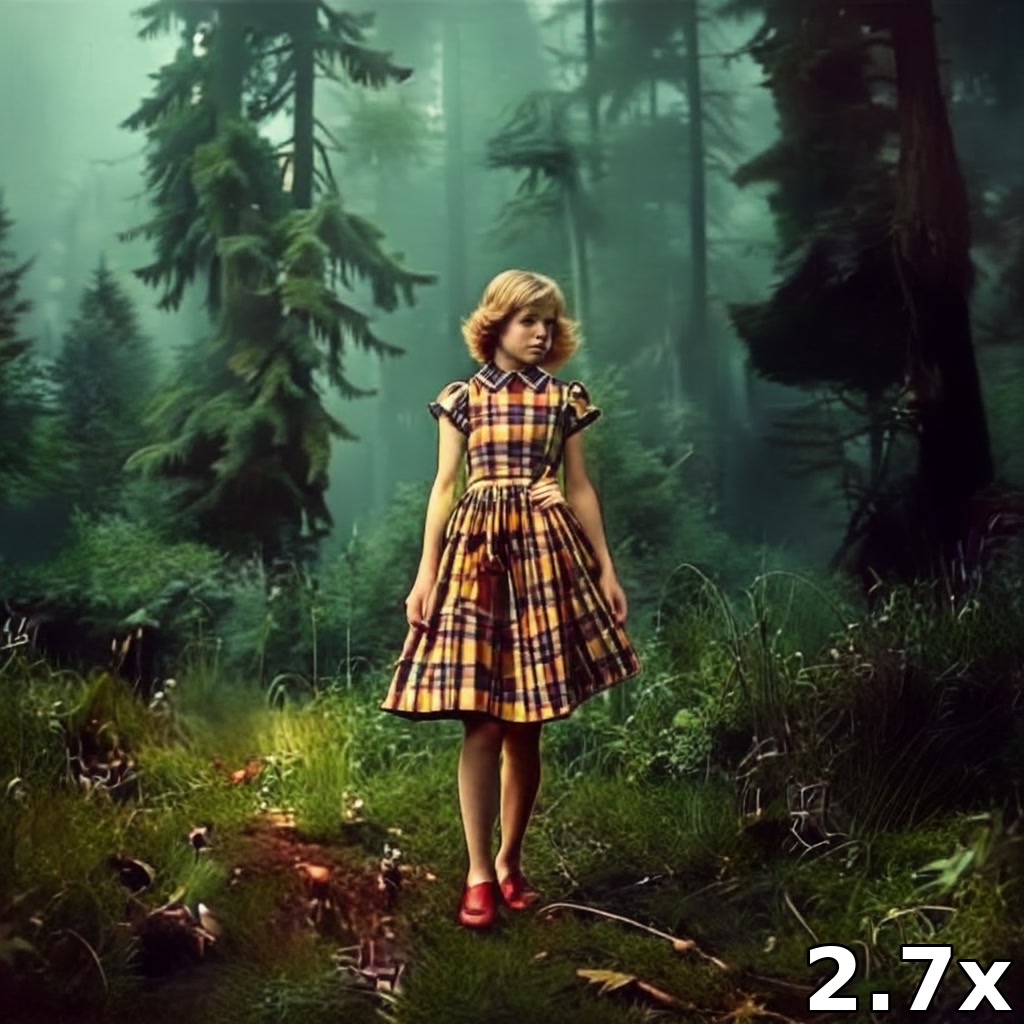} & 
        \includegraphics[trim=2 0 0 2,clip,width=0.2\columnwidth]{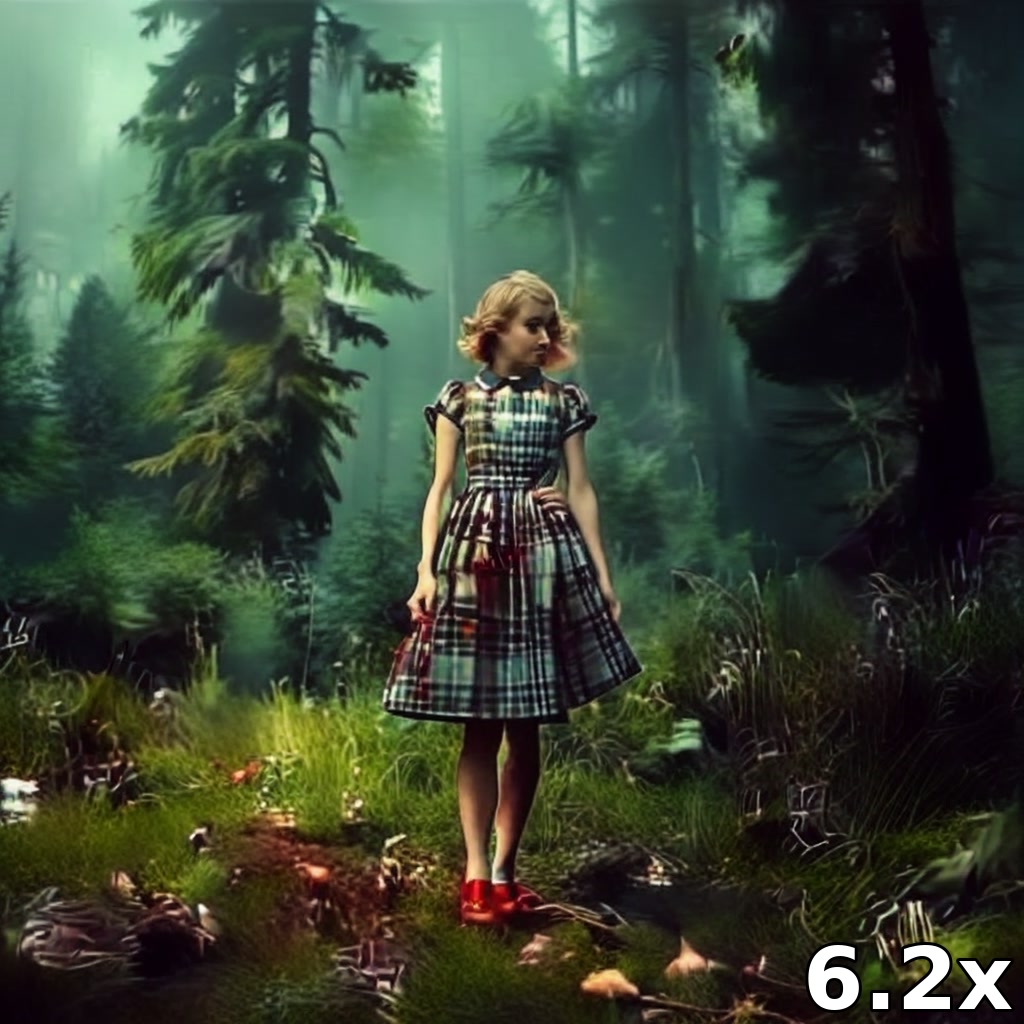} & 
        \includegraphics[trim=2 0 0 2,clip,width=0.2\columnwidth]{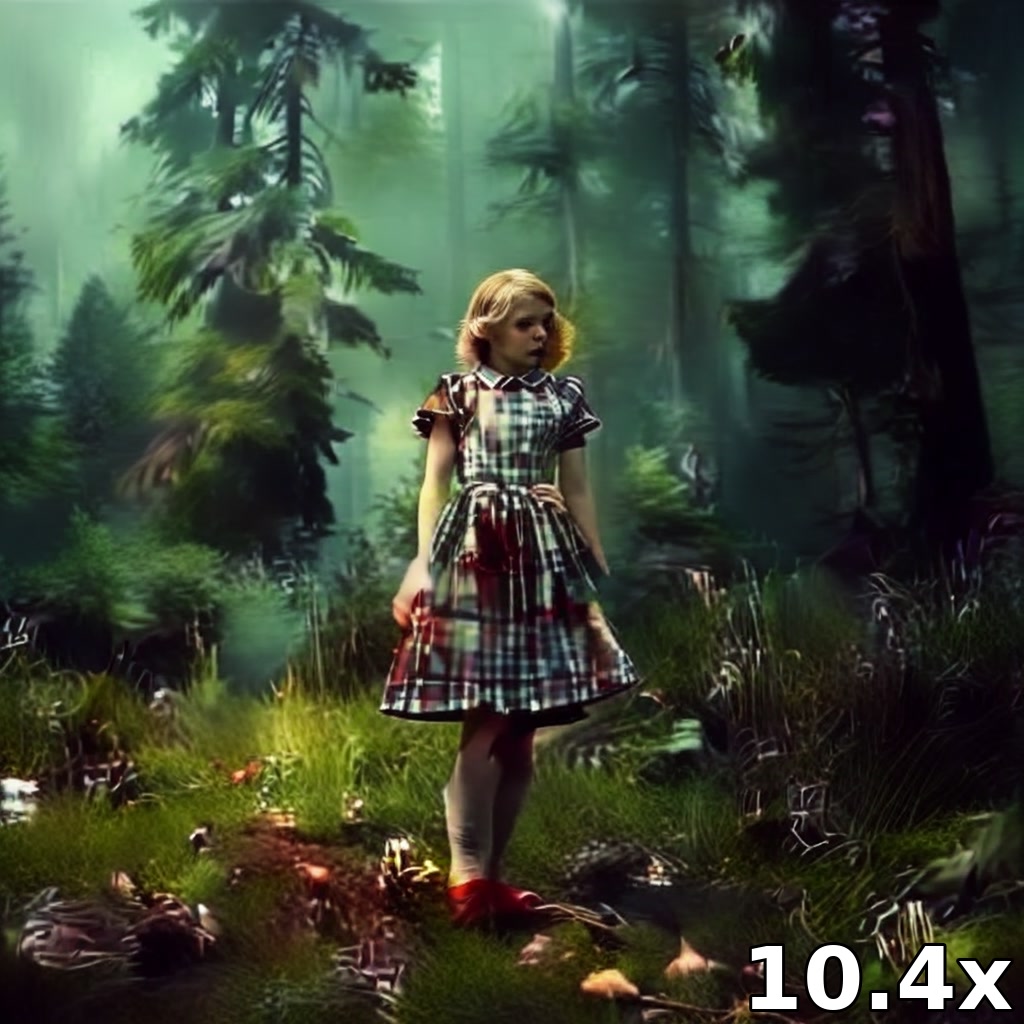}  \\

        \includegraphics[trim=2 0 0 2,clip,width=0.2\columnwidth]{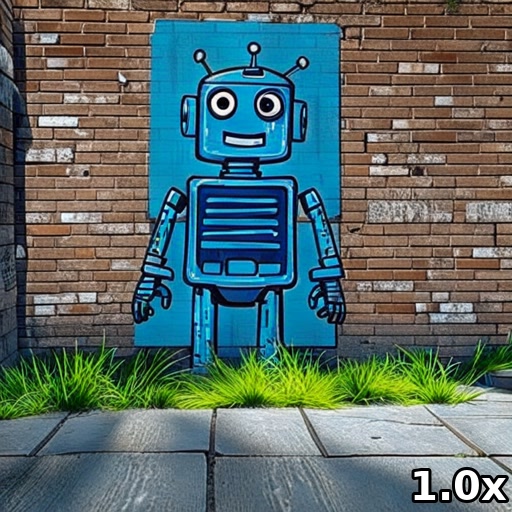} & 
        \includegraphics[trim=2 0 0 2,clip,width=0.2\columnwidth]{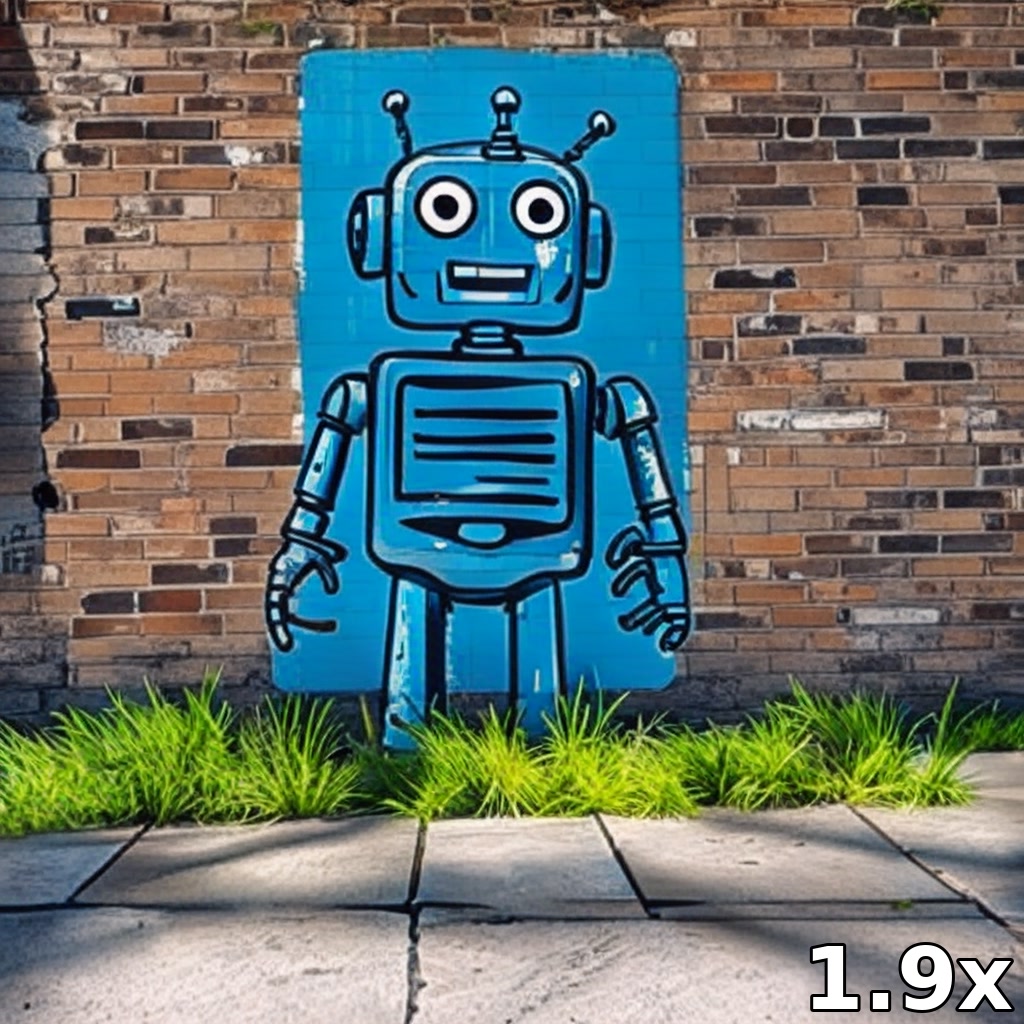} & 
        \includegraphics[trim=2 0 0 2,clip,width=0.2\columnwidth]{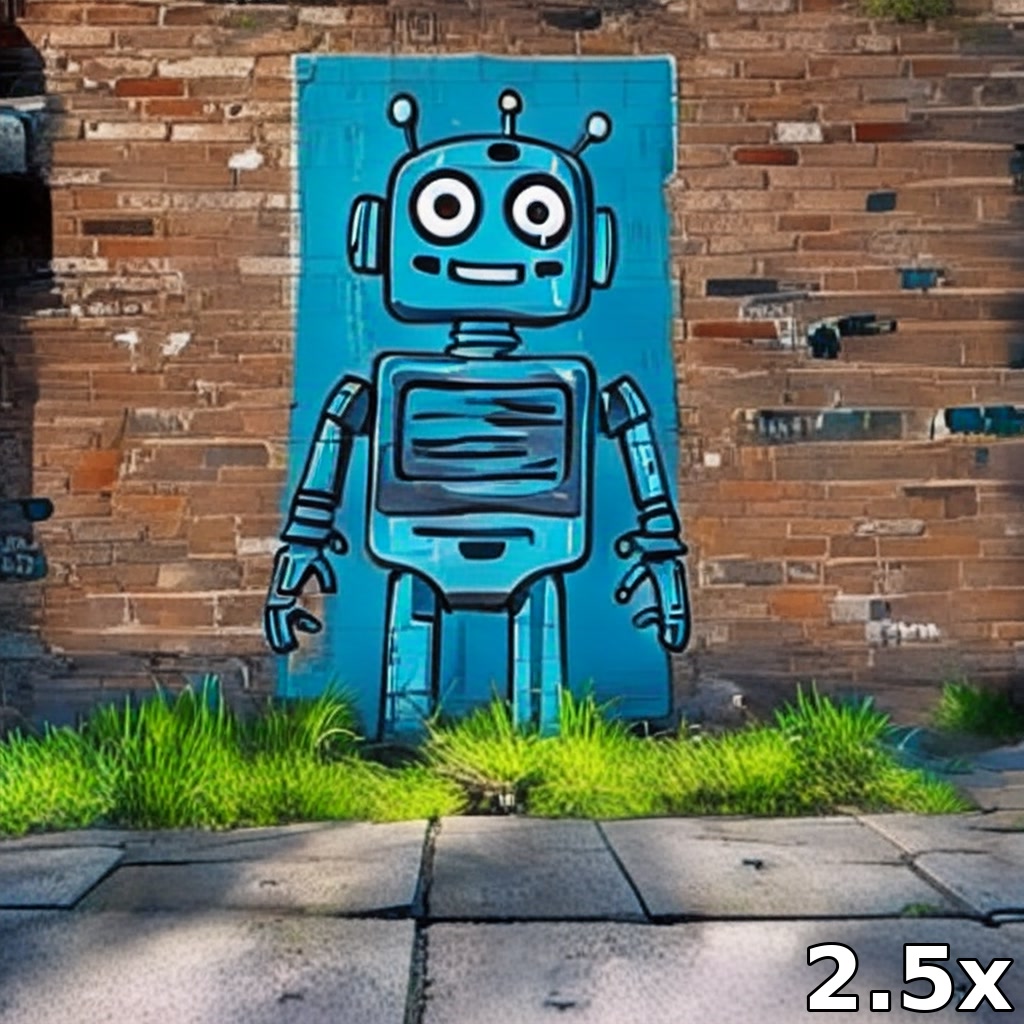} & 
        \includegraphics[trim=2 0 0 2,clip,width=0.2\columnwidth]{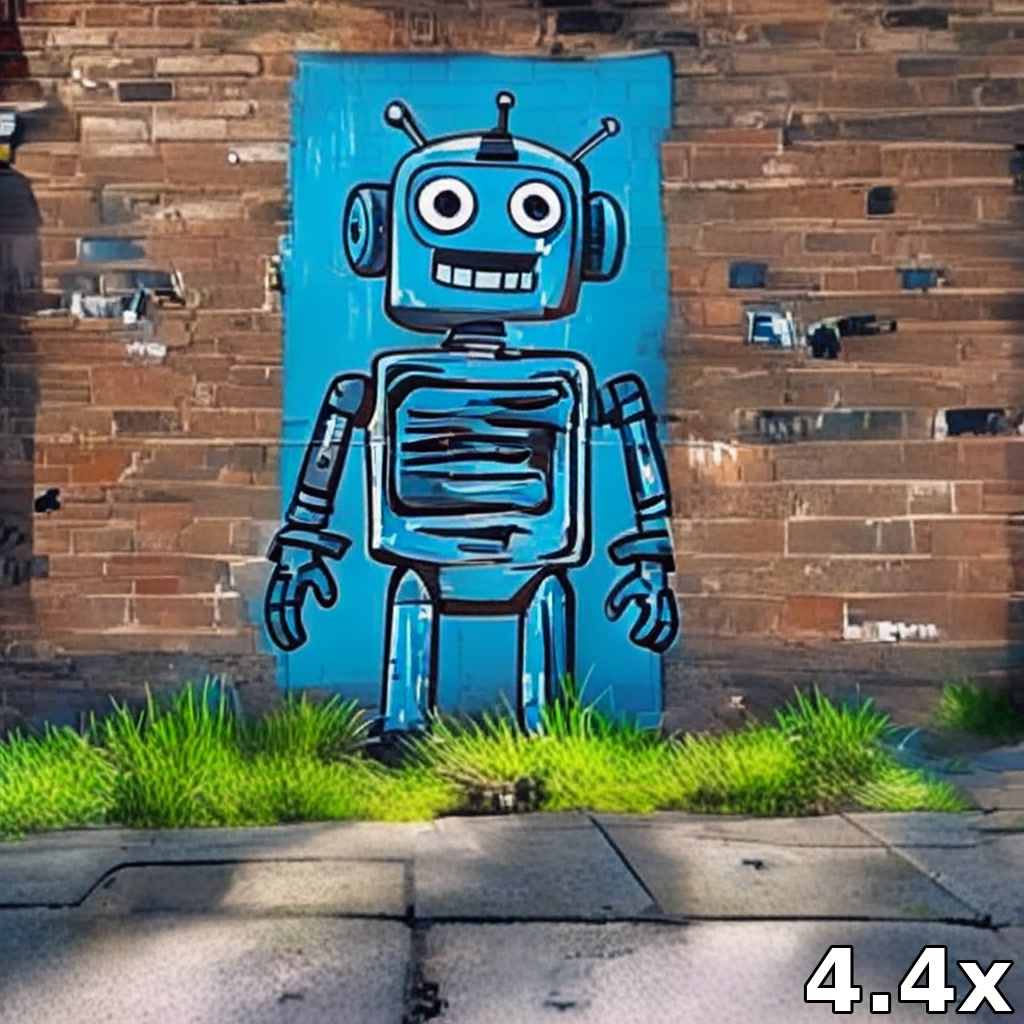} & 
        \includegraphics[trim=2 0 0 2,clip,width=0.2\columnwidth]{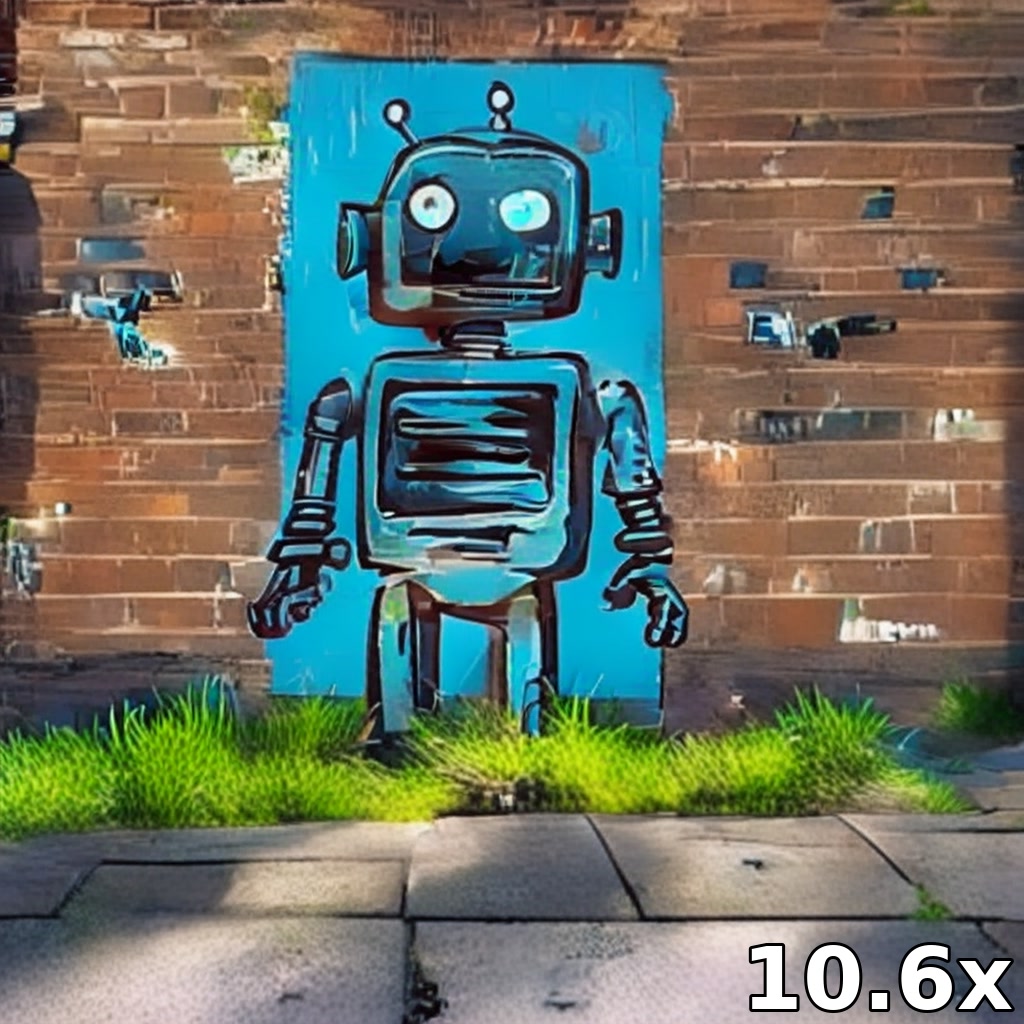}  \\

        \includegraphics[trim=2 0 0 2,clip,width=0.2\columnwidth]{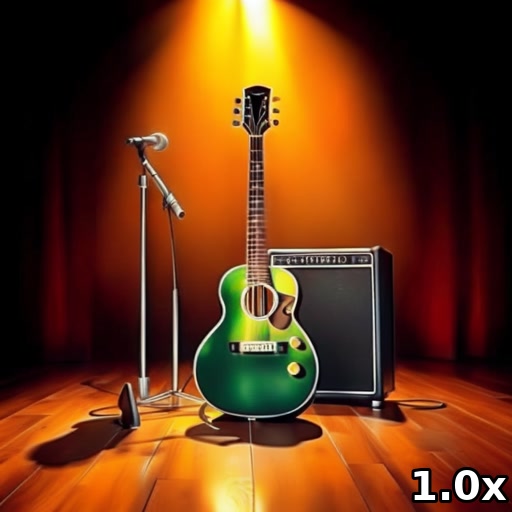} & 
        \includegraphics[trim=2 0 0 2,clip,width=0.2\columnwidth]{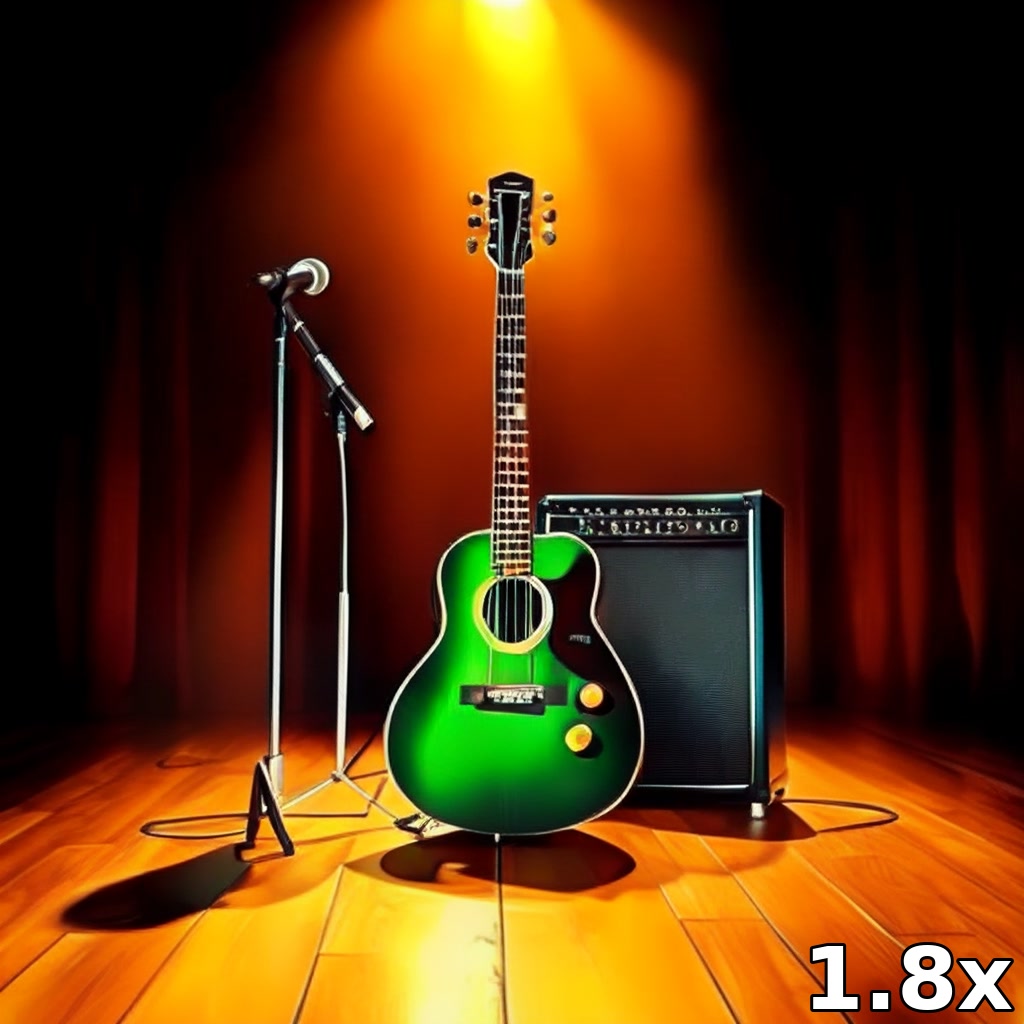} & 
        \includegraphics[trim=2 0 0 2,clip,width=0.2\columnwidth]{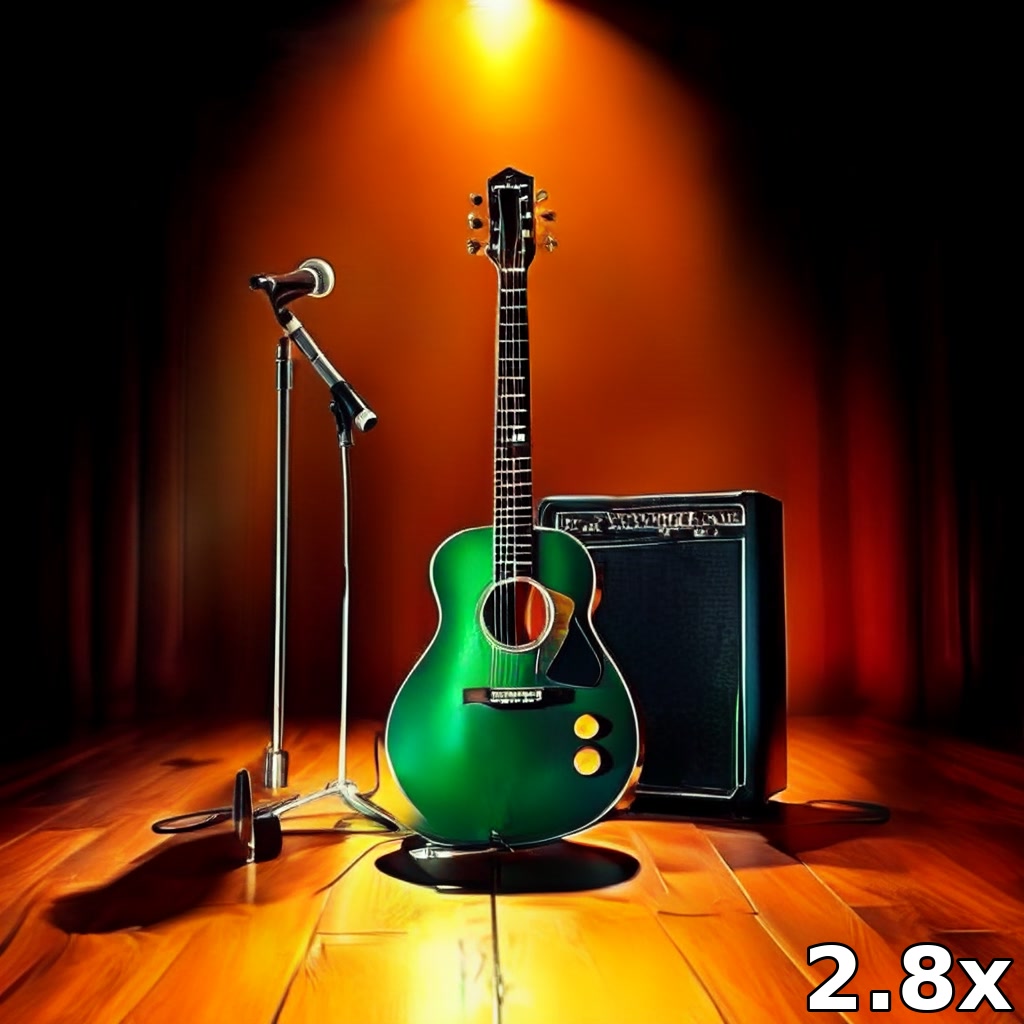} & 
        \includegraphics[trim=2 0 0 2,clip,width=0.2\columnwidth]{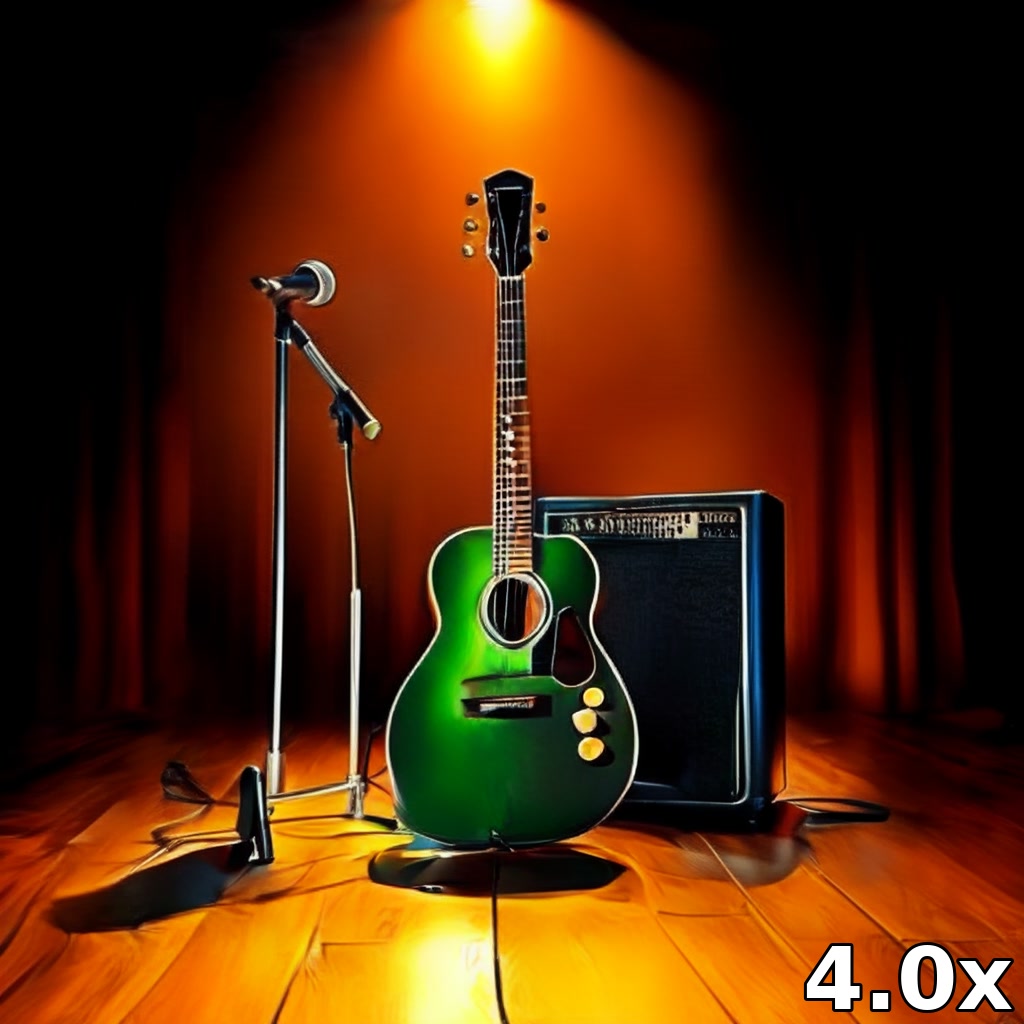} & 
        \includegraphics[trim=2 0 0 2,clip,width=0.2\columnwidth]{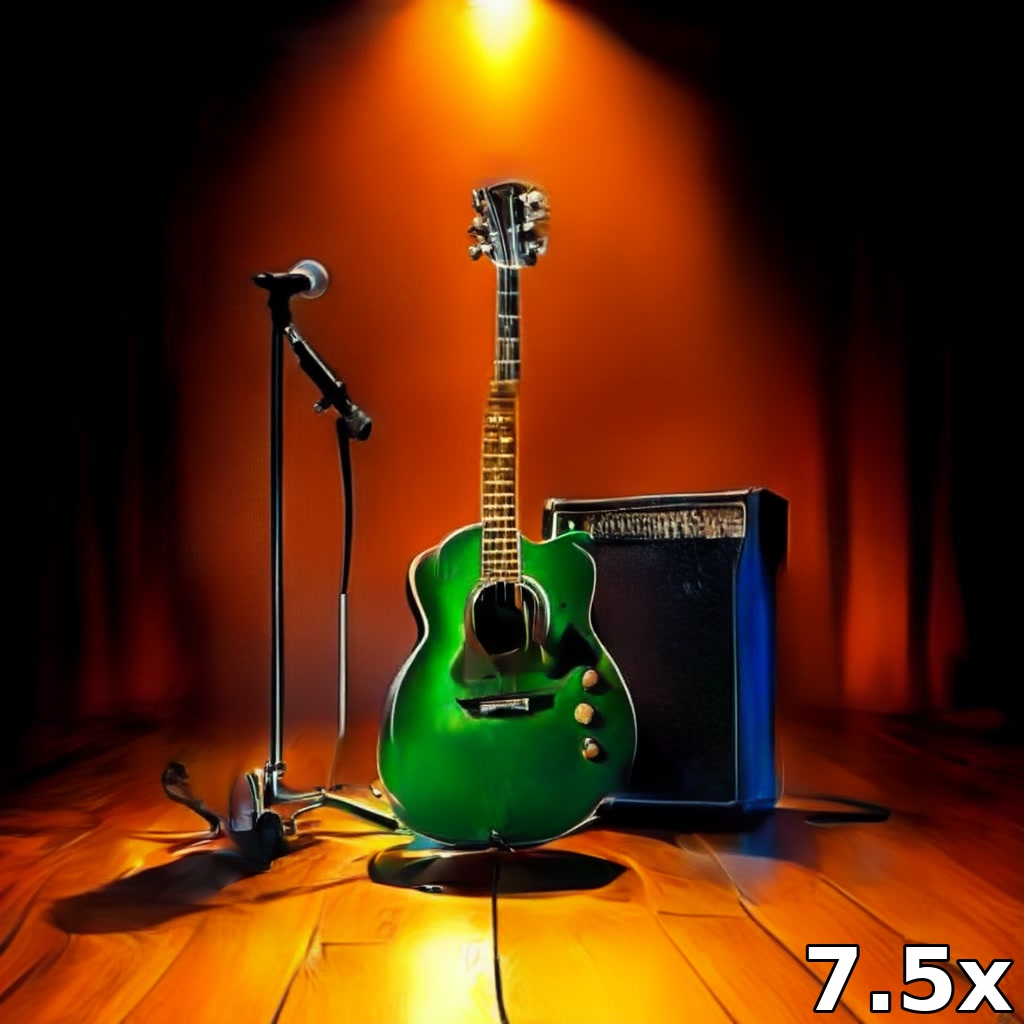}  \\

        \includegraphics[trim=2 0 0 2,clip,width=0.2\columnwidth]{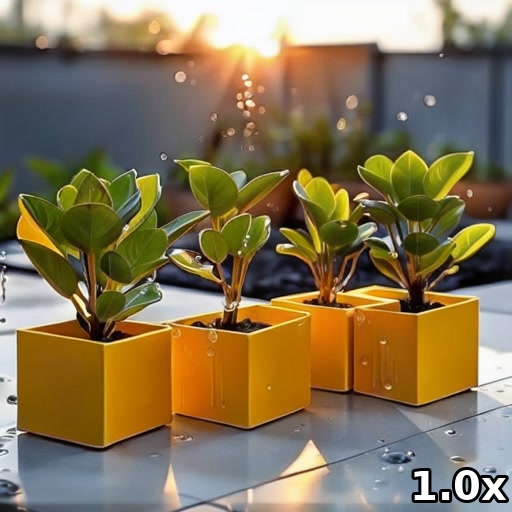} & 
        \includegraphics[trim=2 0 0 2,clip,width=0.2\columnwidth]{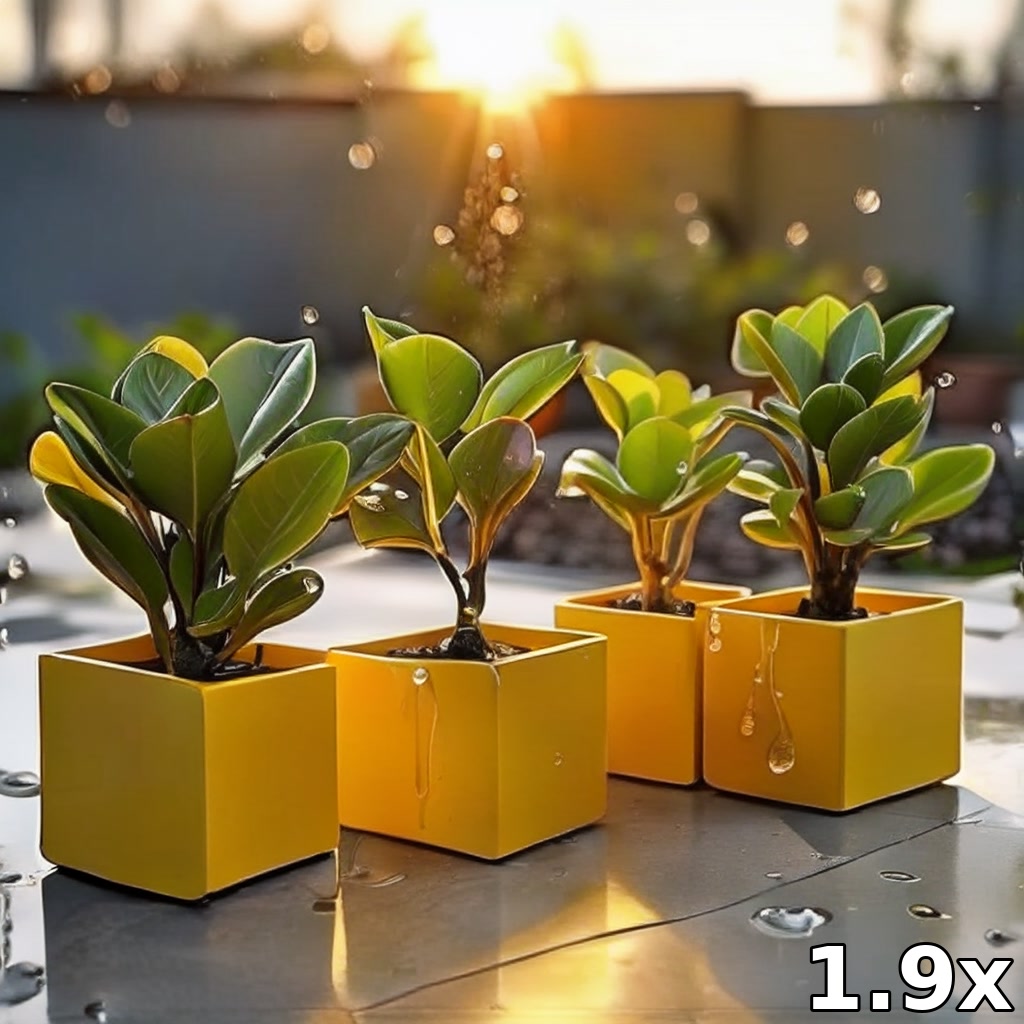} & 
        \includegraphics[trim=2 0 0 2,clip,width=0.2\columnwidth]{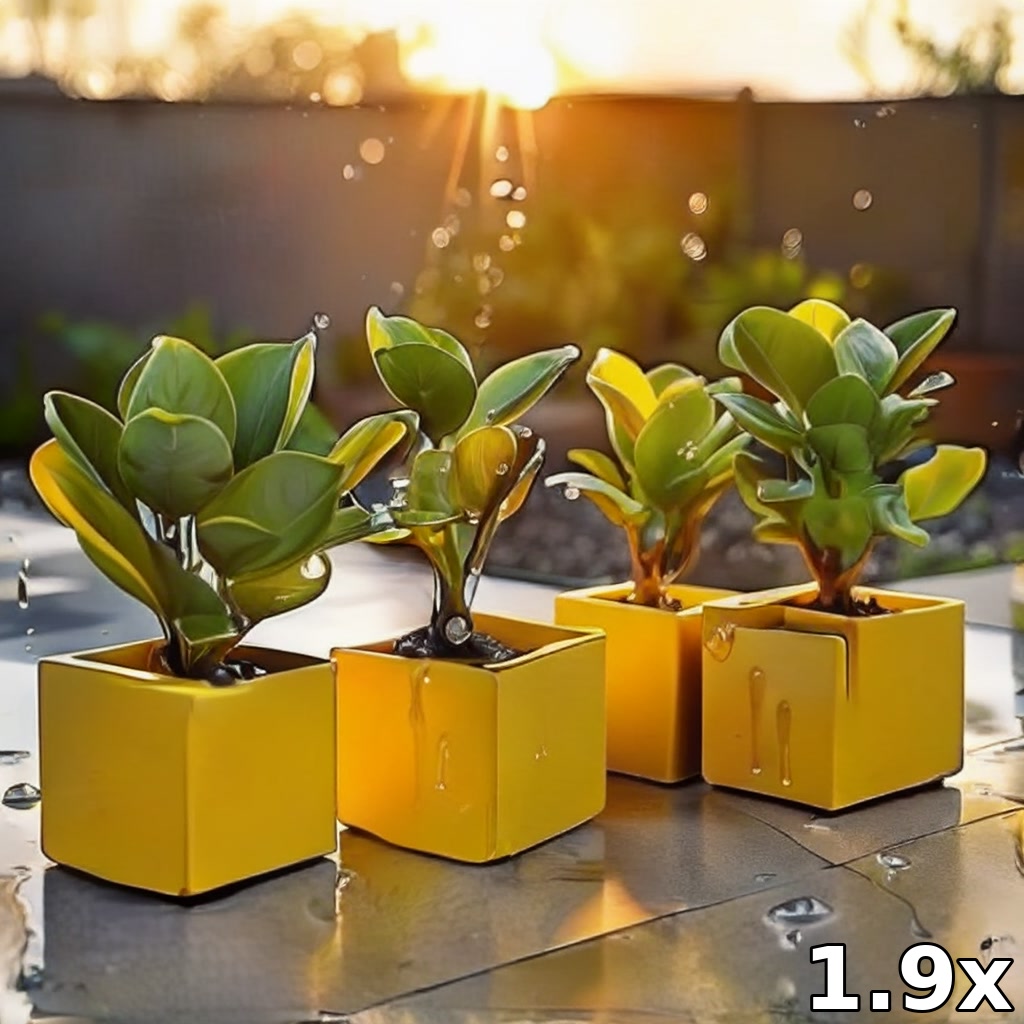} & 
        \includegraphics[trim=2 0 0 2,clip,width=0.2\columnwidth]{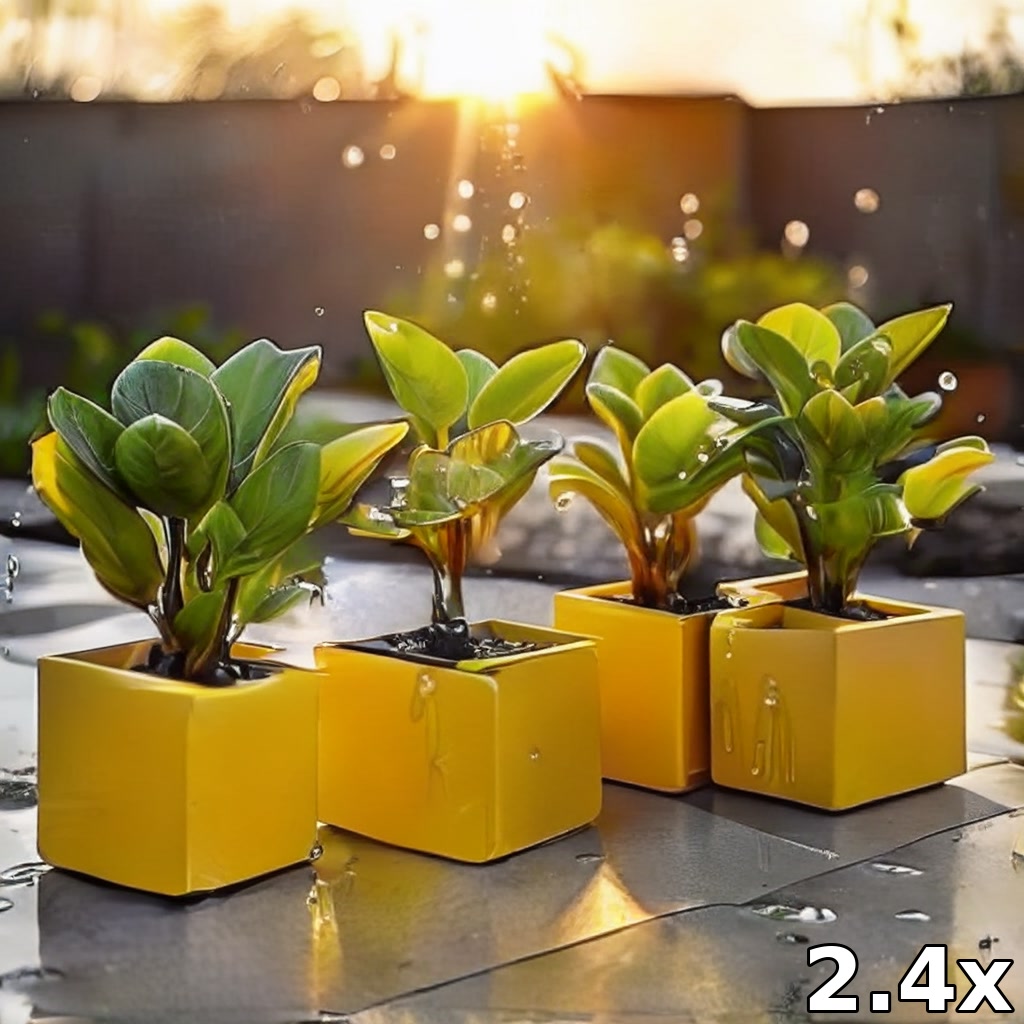} & 
        \includegraphics[trim=2 0 0 2,clip,width=0.2\columnwidth]{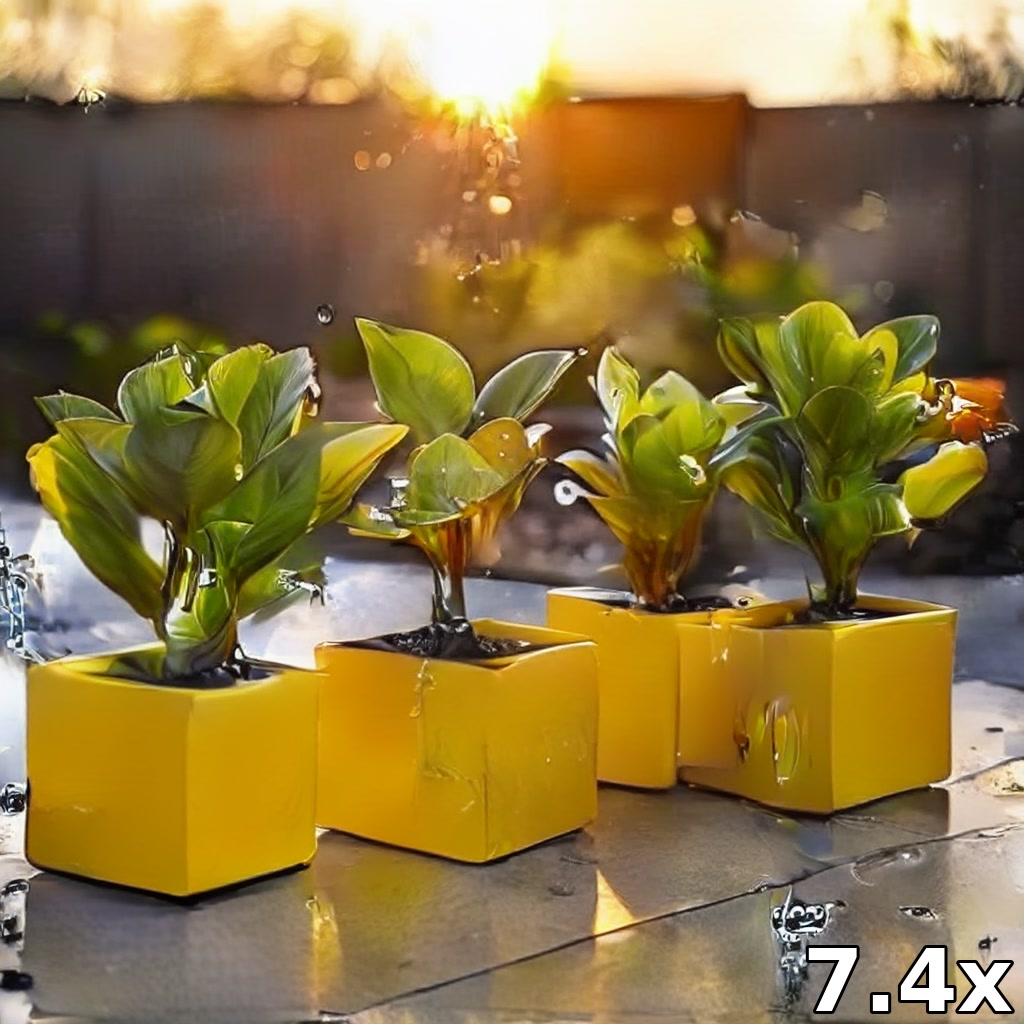}  \\

        \includegraphics[trim=2 0 0 2,clip,width=0.2\columnwidth]{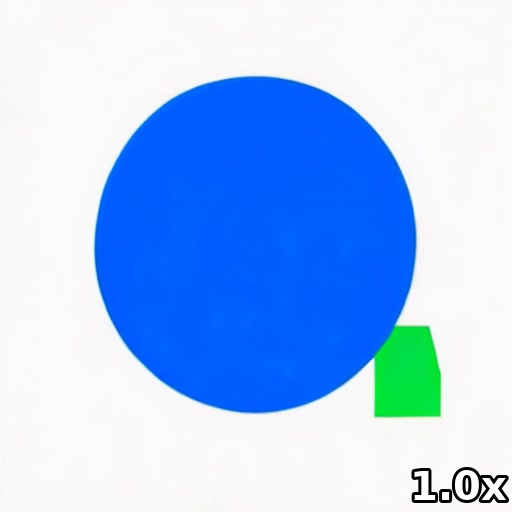} & 
        \includegraphics[trim=2 0 0 2,clip,width=0.2\columnwidth]{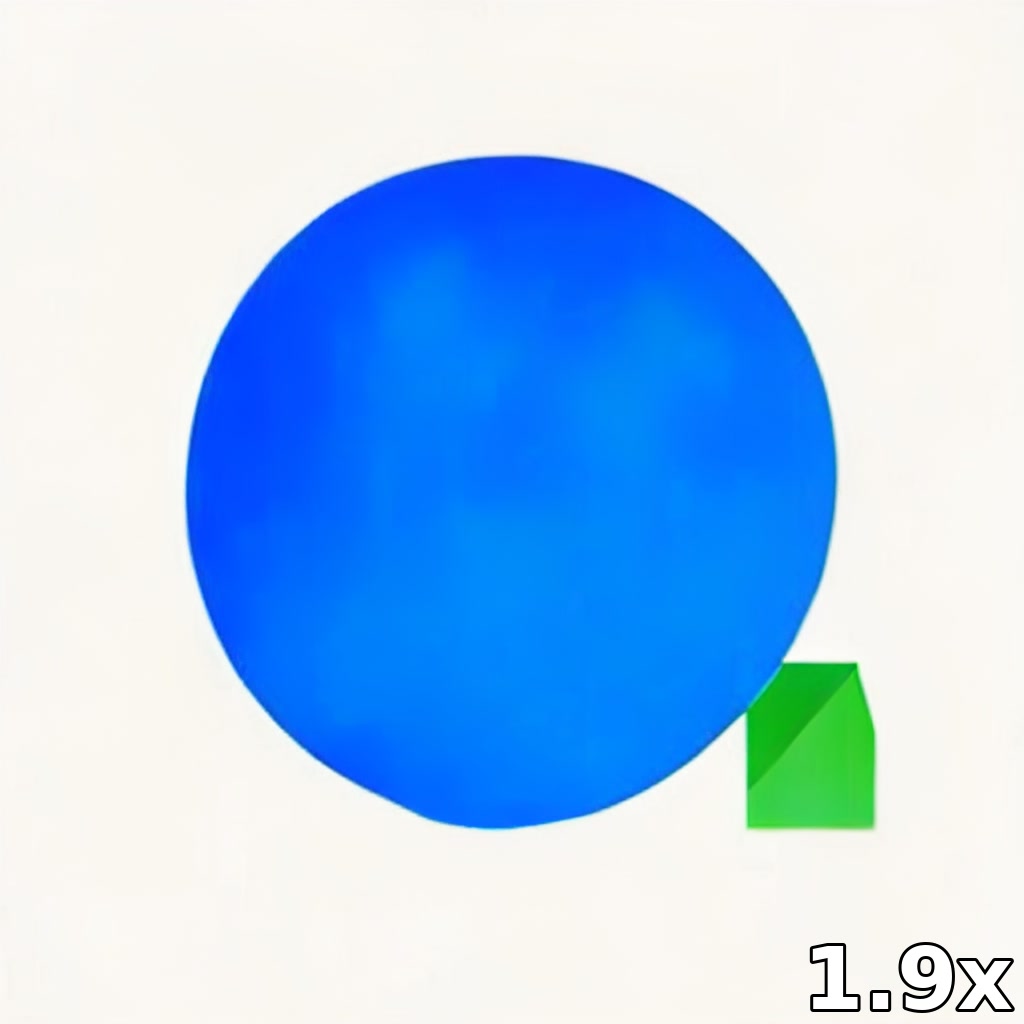} & 
        \includegraphics[trim=2 0 0 2,clip,width=0.2\columnwidth]{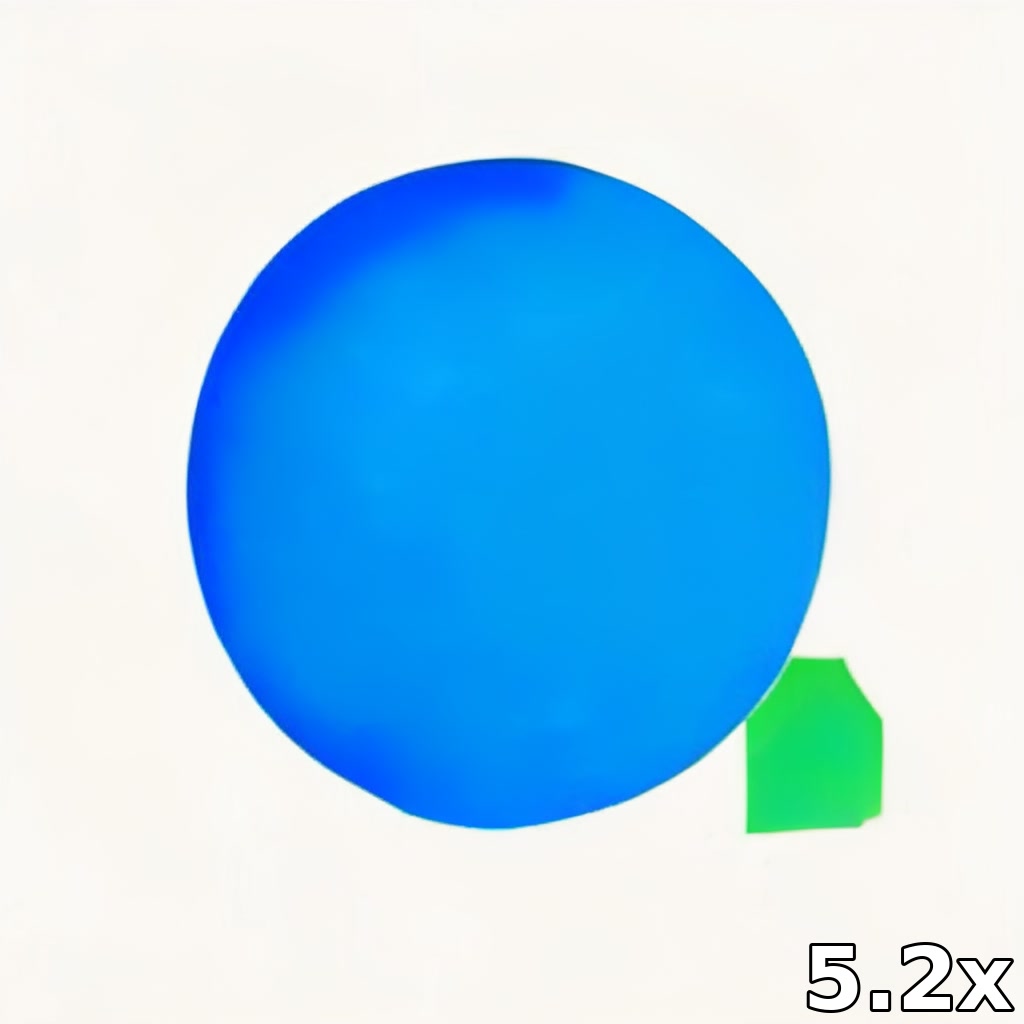} & 
        \includegraphics[trim=2 0 0 2,clip,width=0.2\columnwidth]{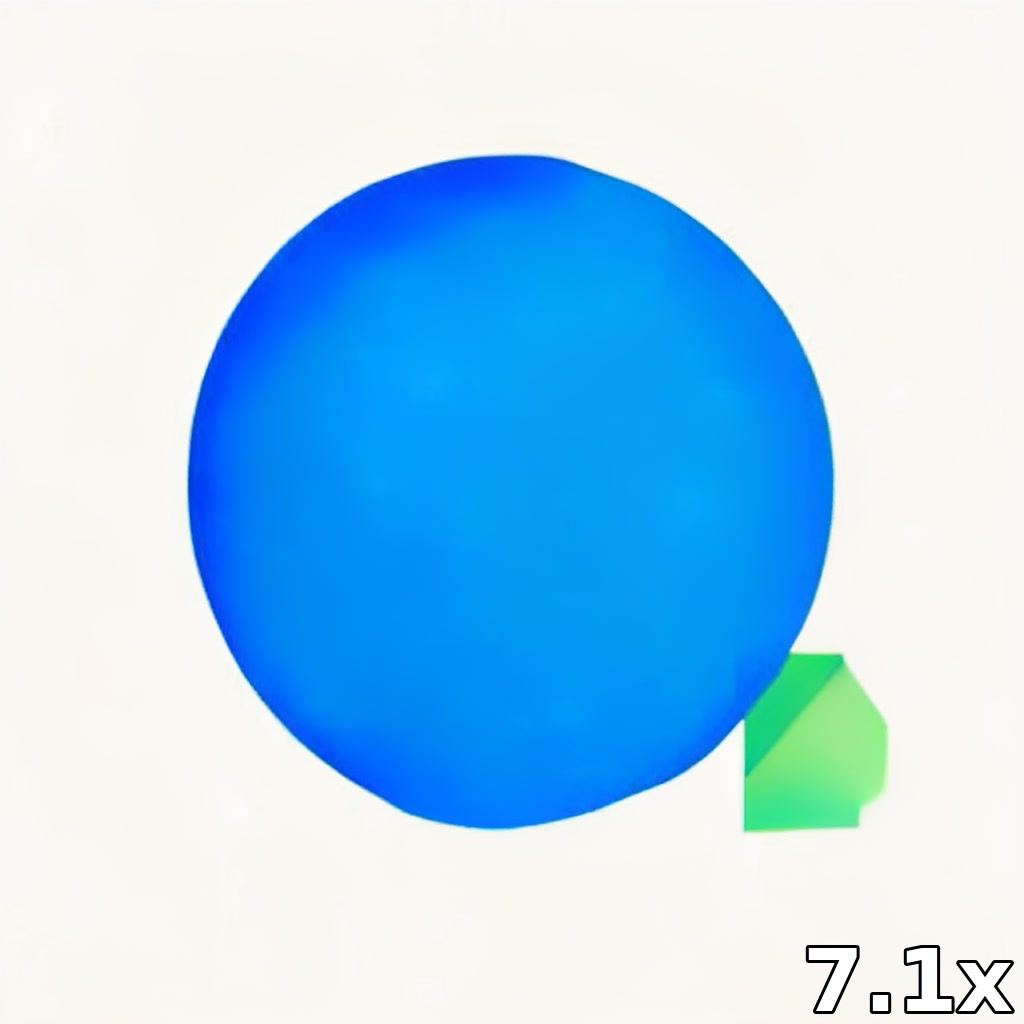} & 
        \includegraphics[trim=2 0 0 2,clip,width=0.2\columnwidth]{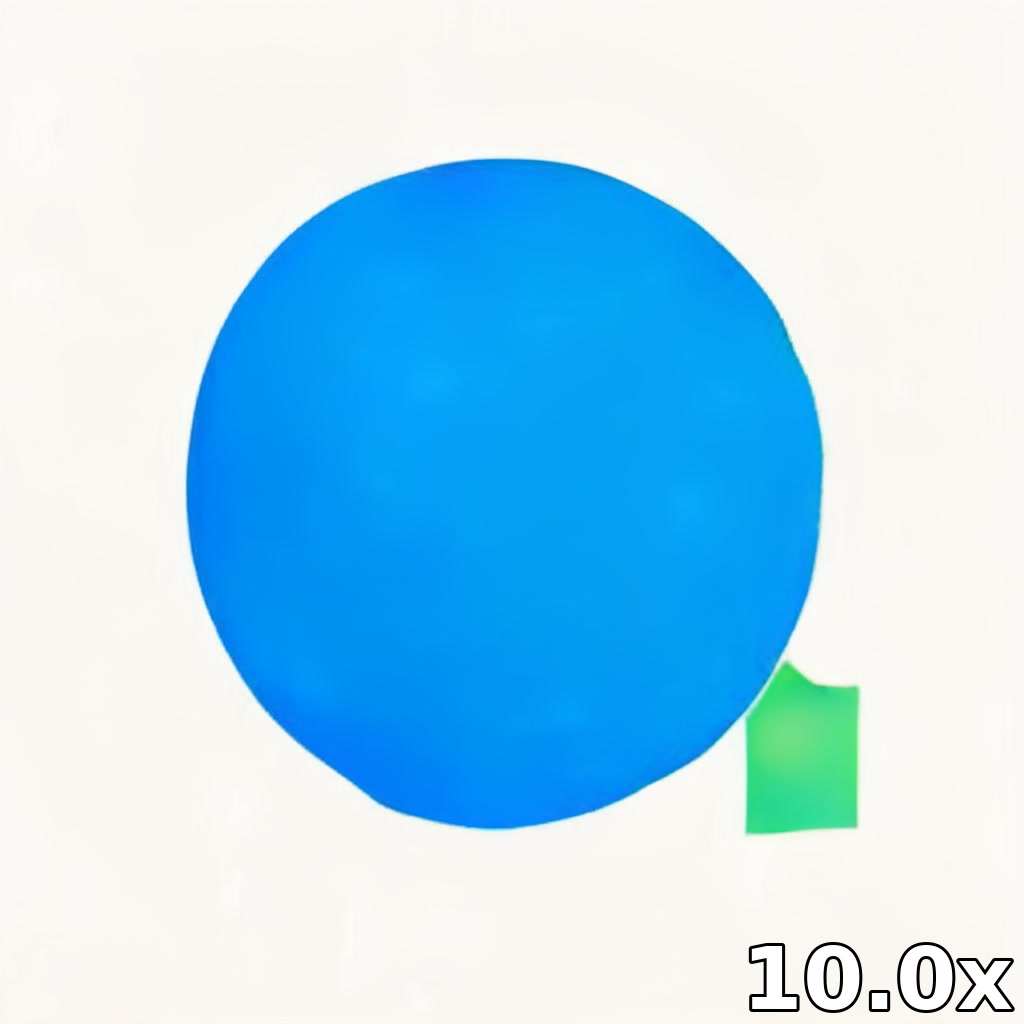}  \\

    \end{tabular}
    }
    \captionof{figure}{Visual comparison of 1024p image generations. We sweep the value of $\tau$, our relaxed acceptance threshold as defined in Equation 5 in the main paper and show the related results. Prompts from DPG-Bench: \texttt{drawtext19.txt, midjourney33.txt, partiprompts177.txt, 73.txt, 74.txt, partiprompts77.txt.}}
    \label{fig:qualitative-tau-sweep-1024}
\end{table*}

\end{document}